%% file: main.tex
\documentclass[10pt,twocolumn,letterpaper]{article}
\usepackage[OT1]{fontenc}
\usepackage[pass]{geometry}
\usepackage[pagenumbers]{cvpr}
\usepackage{cuted}
\usepackage{colortbl,array,tabularx,multirow,longtable,float,fancyhdr}
\definecolor{tablebest}{HTML}{DDE7F5}
\definecolor{tablesecond}{HTML}{EEF3FA}
\definecolor{tablegroup}{HTML}{F0F4FA}
\definecolor{ink}{HTML}{171717}
\definecolor{muted}{HTML}{444444}
\definecolor{rulegray}{HTML}{A9CEED}
\usepackage[pdfversion=1.7,breaklinks,colorlinks,pagebackref,hypertexnames=false,bookmarksdepth=3]{hyperref}
\input{print-links}

\newcommand{\runningtitle}{CaveBench: Benchmarking Abductive Physical Reasoning from Internal Observations}
\input{paper-metadata}
\begin{document}
\input{paper-content}
\newgeometry{onecolumn,left=21mm,right=21mm,top=19mm,bottom=19mm,headheight=13pt,headsep=7pt,footskip=22pt}
\onecolumn
\captionsetup{font=small,labelsep=period}
\captionsetup[table]{position=top,skip=5pt}
\pagestyle{fancy}
\fancyhf{}
\fancyhead[C]{\small\bfseries\runningtitle}
\fancyfoot[C]{\small\thepage}
\renewcommand{\headrulewidth}{.65pt}
\setlength{\parindent}{0pt}
\setlength{\parskip}{3pt plus .5pt}
\setlength{\textfloatsep}{11pt plus 2pt}
\setlength{\intextsep}{6pt plus 1pt}
\renewcommand{\arraystretch}{1.06}
\setcounter{tocdepth}{2}
\setcounter{secnumdepth}{2}
\makeatletter
\renewcommand{\@seccntformat}[1]{\csname the#1\endcsname.\enspace}
\renewcommand{\section}{\@startsection{section}{1}{\z@}{-2.5ex plus -.5ex minus -.2ex}{1.1ex plus .2ex}{\normalfont\large\bfseries}}
\renewcommand{\subsection}{\@startsection{subsection}{2}{\z@}{-1.8ex plus -.4ex minus -.2ex}{.6ex plus .1ex}{\normalfont\fontsize{11}{13}\selectfont\bfseries}}
\makeatother
\setlength{\emergencystretch}{1.2em}
\raggedbottom
\pdfbookmark[0]{Appendix}{appendix-start}
\input{appendix-content}

\end{document}

%% file: print-links.tex
\definecolor{CaveBenchLink}{HTML}{367DBD}
\hypersetup{colorlinks=true,linkcolor=CaveBenchLink,citecolor=CaveBenchLink,
            urlcolor=CaveBenchLink,pdfborder={0 0 0}}

%% file: paper-metadata.tex
\hypersetup{hypertexnames=false,pdftitle={CaveBench: Benchmarking Abductive Physical Reasoning from Internal Observations},pdfauthor={Guanxin Lin, Chen Yang, Shenxiang Zeng, Honghui Zhu, Xin Li, Wenbo Deng, Jiansheng Fan, Chen Wang}}
\title{CaveBench: Benchmarking Abductive\\Physical Reasoning from Internal Observations}
\author{Guanxin Lin\textsuperscript{1,*}\hspace{7.5pt}Chen Yang\textsuperscript{1,*}\hspace{7.5pt}Shenxiang Zeng\textsuperscript{1}\quad Honghui Zhu\textsuperscript{1}\\
Xin Li\textsuperscript{1}\quad Wenbo Deng\textsuperscript{1}\quad Jiansheng Fan\textsuperscript{1}\quad Chen Wang\textsuperscript{1,\textdagger}\\[2.7pt]
\textsuperscript{1}Tsinghua University}

%% file: paper-content.tex
\maketitle
\begingroup
\setlength{\stripsep}{-20pt}
\begin{strip}
\centering
\includegraphics[width=\textwidth]{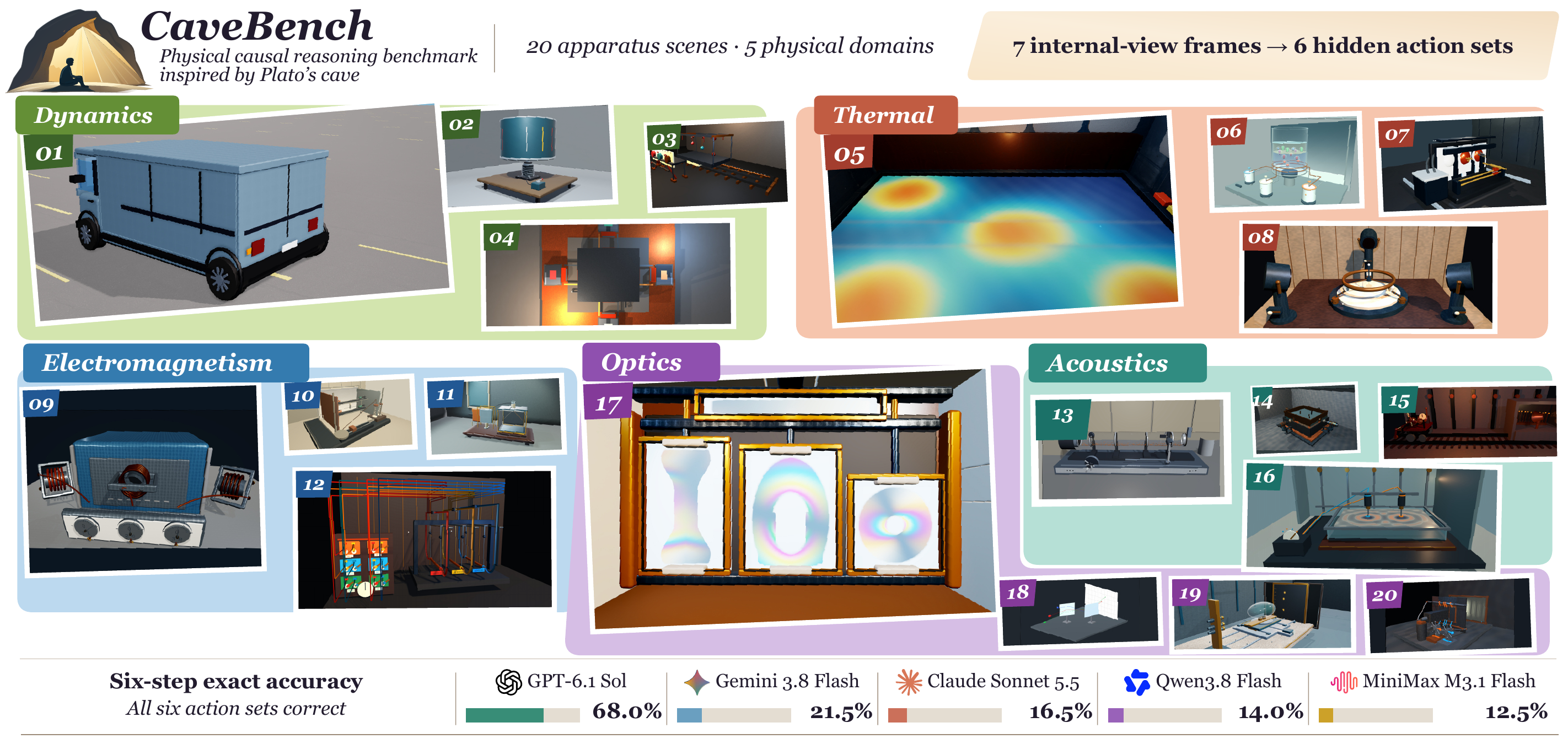}
\captionsetup{hypcap=false}
\captionof{figure}{\textbf{CaveBench: Abductive Physical Reasoning Inspired by Plato's Cave.} Twenty apparatus scenes span five physical domains. Models receive seven internal-view images and identify the hidden external controls applied at each of six steps. Bars show six-event exact-match accuracy on 200 fixed episodes for representative model configurations; the images above illustrate the apparatuses.}
\label{fig:teaser}
\captionsetup{hypcap=true}
\vspace{21.3pt}\end{strip}
\endgroup
\begingroup
\renewcommand{\thefootnote}{*}\footnotetext{These authors contributed equally.}
\renewcommand{\thefootnote}{\textdagger}\footnotetext{Corresponding author.}
\endgroup
\begin{figure*}[t]
\centering
\includegraphics[width=\textwidth]{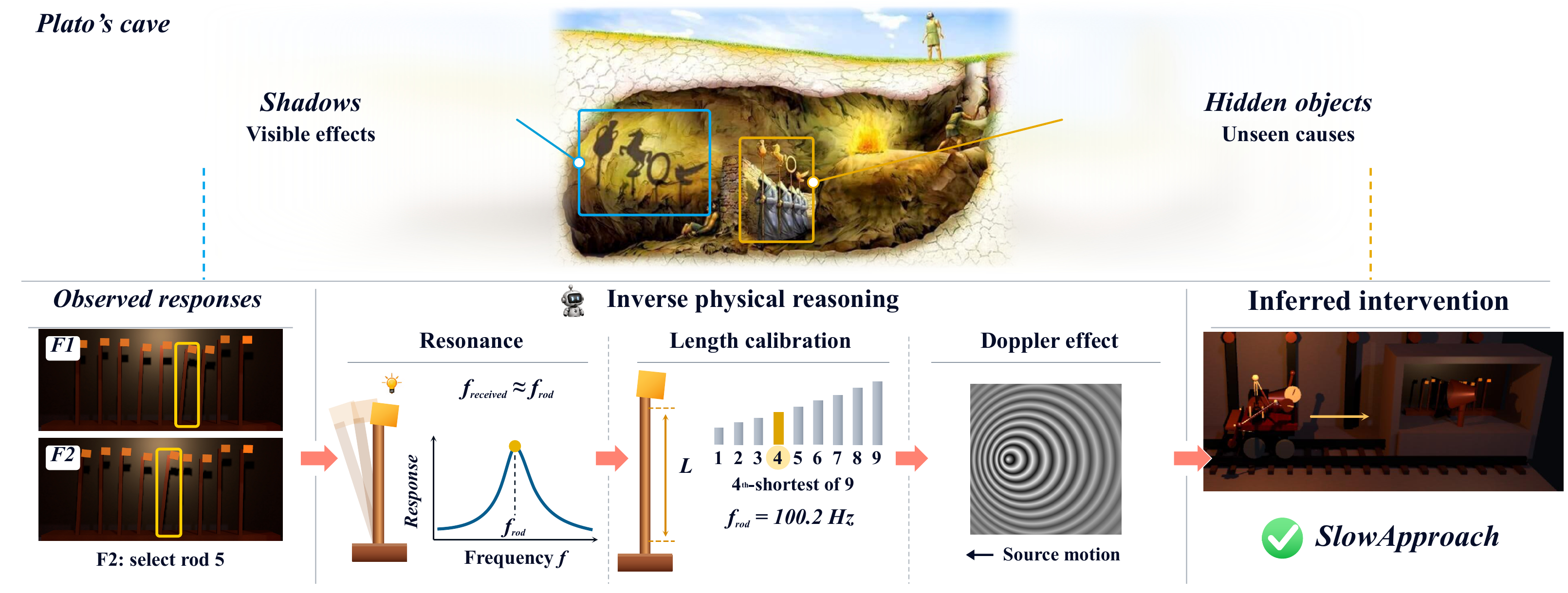}
\caption{\textbf{From Internal Observations to External Causes: Abductive Reasoning with Resonant Rods.} Source frequency: $100\,\mathrm{Hz}$.}
\label{fig:cave}
\end{figure*}
\begin{abstract}
Vision--language models (VLMs) are advancing in physical reasoning, but whether they can understand external physical drivers from internal observations remains an open question. Existing benchmarks have not specifically targeted this setting. To this end, we introduce \textbf{CaveBench}, a benchmark for abductive physical reasoning from internal observations. Models receive apparatus descriptions and internal images while the external controls remain hidden, and must infer the operations responsible for the observed responses. CaveBench spans 20 apparatus scenes in five physical domains and contains 10,200 simulated episodes, split into 9,000 training, 1,000 validation, and 200 test episodes. Questions span one to six events and require the complete control set at every event, testing whether an explanation accounts for the entire process. Synchronized observations and control records provide the inputs and reference answers. Evaluating 27 VLM configurations yields six-event exact-match accuracies of 68.0\% for the strongest closed-source model and 12.5\% for the strongest open-weight model. Seven adaptation baselines show that single-event gains need not improve complete sequences. Error analysis identifies failures in visual identification, physical inference, evidence integration, and action composition, specifying where current models struggle to connect internal phenomena with their external causes. Project page: \url{https://pier-lab.github.io/CaveBench/}.
\end{abstract}
\input{body}
\clearpage
\section*{LLM Usage Statement}
\addcontentsline{toc}{section}{LLM Usage Statement}
GPT was used to polish and revise the manuscript and to assist with figure and table preparation. GPT generated only the CaveBench logo in the upper-left corner of Figure~\ref{fig:teaser} and small decorative icons, such as robots, light bulbs, and clouds; all other images use original source material. GPT was also used to organize data for large-model evaluations and to assist with the initial screening and preliminary categorization of error cases. GPT was also used to assist with references, experimental design, and writing portions of the code.
{\small\input{main.bbl}}

%% file: body.tex
\section{Introduction}

Vision--language models (VLMs) have made rapid progress in understanding physical phenomena and reasoning about physical processes~\cite{qwen25vl,qwen35,spatialvlm}. However, a central question for real-world deployment is whether models reason by understanding and applying physical laws, or merely reproduce memorized mappings between phenomena and answers. Existing evaluations examine physical properties and phenomena~\mbox{\cite{physobjects,physbench,phyx}}, physical plausibility~\cite{intphys,intphys2}, and the prediction of motion and interaction outcomes~\cite{physion,physicsiq}. Related work also considers inferring hidden physical properties from interactions~\cite{physics101,physionpp,comphy} and explaining unseen events through visual evidence~\mbox{\cite{sherlock,videoabc,blackswan}}. These studies provide valuable measures of physical understanding and causal reasoning, but the ability to reconstruct external driving processes from a system's internal responses remains insufficiently assessed.

\begin{figure*}[t]
\centering
\includegraphics[width=\textwidth]{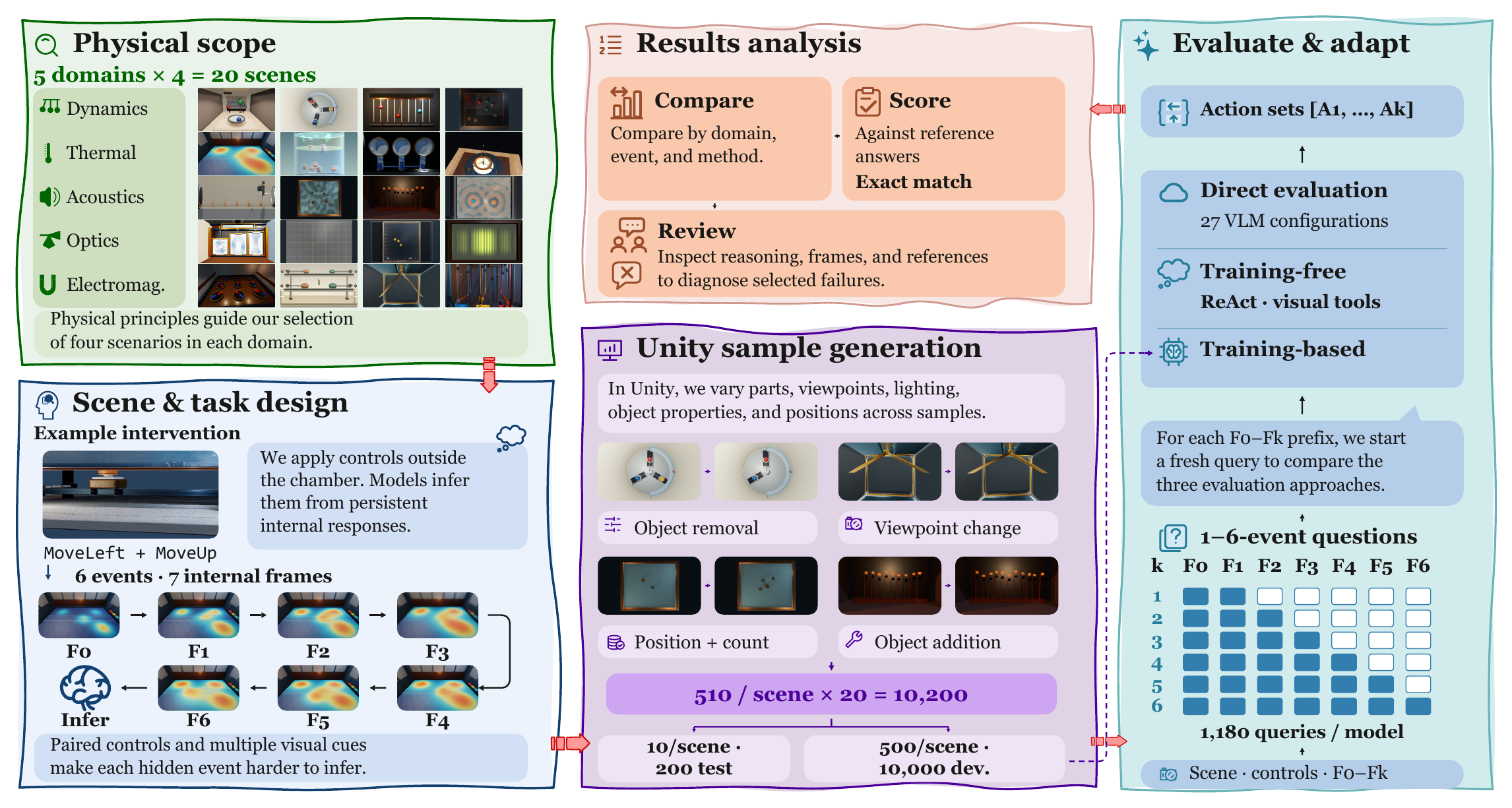}
\caption{\textbf{CaveBench Construction and Evaluation.}}
\label{fig:overview}
\end{figure*}

Existing visual models have grown up ``in the sunlight,'' accustomed to a god's-eye view of physical events. If their viewpoint is confined to the interior of a system, can they still understand the external physical drivers? From this viewpoint, the causes are hidden, and models must infer what happened outside from the internal responses (Figure~\ref{fig:cave}). Because earlier effects can persist and several operations can act together, an explanation must remain consistent with observations across time. To this end, we introduce \textbf{CaveBench}, a benchmark for abductive physical reasoning from internal observations (Figure~\ref{fig:teaser}). Each question provides apparatus principles, available controls, and internal image sequences, asking the model to explain the observations by specifying the complete control set for every event in chronological order. An answer is counted as correct only when the control sets for all events are correct.

CaveBench covers dynamics, thermal physics, electromagnetism, acoustics, and optics, with four apparatus scenes per domain. We generate the experiments in Unity, varying apparatus configurations, physical parameters, viewpoints, and operations across 510 episodes per scene, for 10,200 episodes in total. Each episode preserves the evolving physical state and records an initial image followed by six response images alongside the applied controls, aligning the observations available to the model with the reference answers. We select ten test episodes per scene to cover different configurations and operation combinations, and check image completeness, control validity, and alignment between images and records (Figure~\ref{fig:overview}).

We construct 1,180 questions spanning one to six events from 200 test episodes and evaluate 27 VLM configurations and seven adaptation baselines. The strongest closed-source and open-weight models achieve 68.0\% and 12.5\% six-event exact-match accuracy, respectively, while domain difficulty rankings differ across model groups. Adaptation can improve single-event accuracy without improving complete sequences. Error analysis further identifies four categories of failure: visual identification, physical inference, evidence integration, and action composition.

We make three contributions. First, we introduce CaveBench to evaluate physical abduction through complete control-sequence reconstruction from internal observations. Second, we compare VLMs across physical domains and event lengths, and examine the effects of task adaptation and reasoning effort. Third, we analyze recurring errors to identify concrete targets for improving how models connect observed phenomena with the physical processes that produce them.

\section{Related Work}

\textbf{Physical Understanding Benchmarks.}\ Physical understanding benchmarks cover physical knowledge, object properties, and system behavior. Everyday physical knowledge and object properties are studied in PIQA~\cite{piqa}, PhysObjects~\cite{physobjects}, and PACS~\cite{pacs}. Multimodal evaluations examine physical concepts and reasoning across diverse scenes, including PhysBench~\cite{physbench}, PhyX~\cite{phyx}, and GRASP~\cite{grasp}. Quantitative evaluation ranges from scientific problem solving in PhysReason~\cite{physreason} and PHYBench~\cite{phybench} to estimating physical quantities from video in QuantiPhy~\cite{quantiphy}. Dynamic evaluations assess physical plausibility in IntPhys~\cite{intphys}, IntPhys~2~\cite{intphys2}, InfLevel~\cite{inflevel}, and ADEPT~\cite{adept}, and future behavior in Physion~\cite{physion} and Physics-IQ~\cite{physicsiq}. However, models may arrive at correct answers through familiar visual patterns, linguistic cues, or learned input--answer associations without fully analyzing the physical process at hand~\cite{mvp}. High task accuracy therefore does not necessarily demonstrate the ability to apply physical laws. Final-answer accuracy alone cannot reliably distinguish evidence-based physical reasoning from the reuse of memorized associations between phenomena and answers.

\textbf{Visual Causal Reasoning Benchmarks.}\ Visual causal reasoning benchmarks relate observations to events, hidden physical properties, and possible actions. Controlled physical environments support causal explanations in CLEVRER~\cite{clevrer} and CRAFT~\cite{craft}, counterfactual prediction in CRIPP-VQA~\cite{crippvqa}, ContPhy~\cite{contphy}, and CoPhy~\cite{cophy}, and choosing actions to achieve physical goals in PHYRE~\cite{phyre} and Phy-Q~\cite{phyq}. Another line infers hidden physical properties, such as mass, charge, or friction, from motion and interaction, including Galileo~\cite{galileo}, Physics~101~\cite{physics101}, Physion++~\cite{physionpp}, and ComPhy~\cite{comphy}. Video benchmarks examine explanations and relations among everyday actions, including NExT-QA~\cite{nextqa}, STAR~\cite{star}, and Causal-VidQA~\cite{causalvidqa}. Inference beyond directly visible content includes plausible interpretations and missing events, as examined by Sherlock~\cite{sherlock}, NL-Eye~\cite{nleye}, VideoABC~\cite{videoabc}, and BlackSwanSuite~\cite{blackswan}. However, predictive and counterfactual tasks primarily test the consequences of given causes or interventions, and do not directly establish the ability to reason backward from effects to causes. Existing abductive tasks address scene interpretations and missing events, but tracing hidden physical drivers from internal responses alone remains insufficiently tested. In particular, the ability to distinguish sequential external influences with overlapping effects and explain the entire observed process using physical laws still lacks systematic evaluation.

\begin{figure*}[t]
\centering
\includegraphics[width=\textwidth]{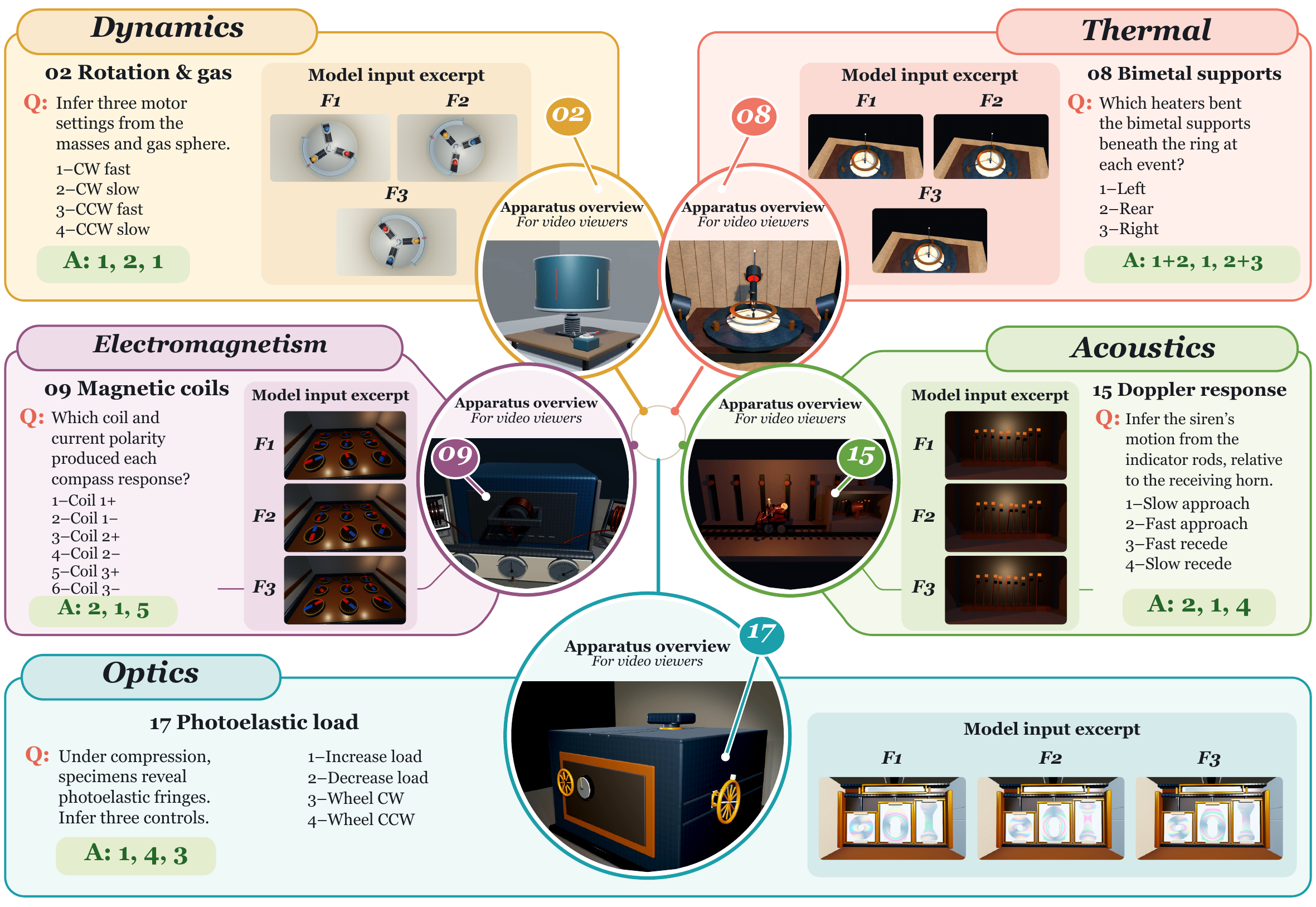}
\caption{\textbf{Representative CaveBench Questions.} Full inputs also include $F_0$ and apparatus descriptions. $+$ joins labels within an event.}
\label{fig:atlas}
\end{figure*}

\section{CaveBench}
CaveBench pairs continuous apparatus experiments with discrete records of the controls applied. We specify the information available to the model and the required answer in Section~\ref{sec:problem_formulation}, describe the evidence offered by different apparatuses in Section~\ref{sec:benchmark_overview}, and explain data generation and verification in Section~\ref{sec:benchmark_construction}.

\subsection{Problem Formulation}
\label{sec:problem_formulation}
CaveBench studies abductive physical reasoning: recovering a sequence of unobserved operations from their effects inside an apparatus. Let $\mathcal C$ denote the available controls and $A_t\subseteq\mathcal C$ the nonempty set specifying the operation at event $t$. Each episode comprises six events, represented by $\mathbf A_{1:6}=(A_1,\ldots,A_6)$, and seven internal images $F_{0:6}=(F_0,F_1,\ldots,F_6)$. The image $F_0$ records the initial state, while $F_t$ records the response after event $t$. A set $A_t$ may contain multiple labels to represent concurrent controls or multiple parameters of an operation.

The apparatus description $d$ specifies its operating principles, control meanings, reference directions, and capture timing. The apparatus retains its physical state between events, so an observation can reflect both the current operation and the effects of preceding operations. A proposed sequence must therefore account for the changes across images in accordance with the apparatus description. The applied controls remain outside the model's view; their recorded values serve as reference answers. The apparatus description provides conventions for interpreting an image, but does not specify which controls were chosen in that episode. Those choices must be inferred by relating the description to the observed sequence. A response can remain visible in subsequent observations, so its continued presence does not by itself establish that the same operation was repeated. Assigning controls to steps requires distinguishing new changes from effects carried forward from preceding events.

Each question provides $d$, $\mathcal C$, and the images $F_{0:k}$ covering the initial state and the first $k\in\{1,\ldots,6\}$ events. The model $f$ predicts the corresponding sequence of control sets:
\begin{equation}
\widehat{\mathbf A}^{(k)}_{1:k}=f(d,\mathcal C,F_{0:k}),\qquad
\mathbf A_{1:k}=(A_1,\ldots,A_k).
\label{eq:inference}
\end{equation}
The answer lists these sets in the order of the corresponding steps. Each set must include all controls applied at that step, while the order of controls within the set does not matter.

\input{table_domains}

\subsection{Overview of CaveBench}
\label{sec:benchmark_overview}
The apparatuses pose different demands on how evidence is read and combined. We construct four Unity scenes in each of five domains: Dynamics, Thermal Physics, Electromagnetism, Acoustics, and Optics (Figures~\ref{fig:teaser} and~\ref{fig:overview}). All follow Eq.~\eqref{eq:inference}, while their observation mechanisms differ.

In the vehicle cabin, balloons, balls, and pendulums provide complementary motion cues that must be interpreted in the apparatus reference frame. In the photoelastic apparatus, both load and polarizer angle affect appearance; a load-independent reference strip helps distinguish these explanations. In the double-mirror apparatus, one downstream pattern reflects the configurations of both mirrors. In the layered-liquid apparatus, an answer must identify both the affected layer and whether it is heated or cooled. These cases exercise component identification, reference-based comparison, and interpretation of joint effects within the same answer format (Figure~\ref{fig:atlas}).

The benchmark contains 10,200 episodes. A test set of 200 episodes, with ten per scene, yields 1,180 questions covering one to six steps. Scene descriptions specify component identities, reference directions, and capture timing. The model must bind these conventions to the observed configuration, then compare responses across the available images.

\subsection{Benchmark Construction Process}
\label{sec:benchmark_construction}
Figure~\ref{fig:overview} outlines how we build questions from internal observations and obtain reference answers from records of the applied external controls.

\textbf{Data Generation.}\ We generate episodes in Unity with the external controls outside the camera's view and their responses visible inside the apparatus. Object arrangements, physical parameters, viewpoints, lighting, and controls vary across episodes. We capture one initial image and one image after each of six control steps, yielding seven internal images. The apparatus continues without resetting between steps. We generate 510 episodes for each of the 20 scenes, giving 10,200 episodes in total.

\textbf{Task Design.}\ Questions cover repeated operations, changes of operation, and combinations of controls. These sequences require distinguishing a new event from a state that persists, and identifying every member of a compound event. Models express their answers using scene-specific control identifiers.

\textbf{Answer Annotation.}\ At each step, the simulator associates the applied controls with the corresponding internal image. We retain both labels when an event involves two controls or two parameters. Reference answers therefore come from the generation process, independently of any interpretation of the rendered images. These labels describe actions executed by the simulator, whereas the images depict the resulting apparatus states. Multiple labels can therefore be associated with one response image without requiring a separate visual region for each label. Questions built from different prefixes of the same episode share their earlier labels: a longer input extends the history to recover, while preserving the recorded operations at preceding steps.

\textbf{Question Generation.}\ We select ten episodes per scene to cover different apparatus settings and action combinations. Each question combines the scene description, initial image, and observations through step $k$, with the corresponding operation records as its reference answer. Each episode normally yields six questions, covering one to six steps. S11 (dielectric response) requires visual references for full insertion, half insertion, and withdrawal to avoid ambiguity. We show them in the first three steps to avoid textual hints that could reveal answers. S11 thus uses only three- to six-step questions, yielding $190\times6+10\times4=1{,}180$ test questions in total.

\textbf{Quality Control.}\ We check that all seven images are present, that each response image corresponds to its recorded control step, and that the recorded controls are valid for the apparatus. We retain videos showing the external controls and internal responses together, allowing the recorded operations to be inspected alongside their physical effects. Only the internal view is included in the images shown to models.

\section{Experiments and Analysis}

\subsection{Evaluation Settings}
\textbf{VLM Evaluation.}\ We evaluate 27 VLM configurations, including 16 proprietary and 11 open-weight models (Tables~\ref{tab:domains} and~\ref{tab:main}). For brevity, we refer to GPT-6.1 Sol, GPT-6.1 Terra, and Gemini 3.8 Flash as Sol, Terra, and Gemini, respectively, throughout the following analysis. All models receive the same scene descriptions, available controls, and ordered images of the apparatus interior, and infer the external controls applied at each step. We score the final answer after any answer-completion call and count unparseable outputs as incorrect.

\textbf{Evaluation Metrics.}\ A step is correct when the predicted controls exactly match those actually applied, regardless of their order within that step. Exact$_k$ is the accuracy on $k$-step questions, requiring every step to be correct. Exact$_6$ thus measures whether a model identifies all external controls across the six steps. Table~\ref{tab:domains} reports Exact$_6$ for each physical domain, with Overall pooling the 200 six-step questions across all five domains. Overall in Tables~\ref{tab:main} and~\ref{tab:methods} pools exact-match accuracy across all 1,180 questions.

\textbf{Adaptation Baselines.}\ For training-based adaptation, we apply supervised fine-tuning (SFT) with LoRA~\cite{lora} to Qwen3.5-9B~\cite{qwen35} on CaveBench, then separately apply DPO~\cite{dpo}, SimPO~\cite{simpo}, and KTO~\cite{kto} to the SFT model (Table~\ref{tab:methods}). For training-free adaptation, we equip Gemini with ReAct~\cite{react}, Visual Sketchpad~\cite{sketchpad}, and ViperGPT~\cite{vipergpt}, using visual tools to crop and compare images and take measurements from them.

\subsection{Main Results}
Tables~\ref{tab:domains} and~\ref{tab:main} compare performance across physical domains and question lengths. We summarize the findings below.

\textbf{The relative difficulty of a physical domain varies across model groups.}\ Models with low overall accuracy generally score poorly across physical domains, but often perform relatively better in electromagnetism (Table~\ref{tab:domains}). MiniMax M3.1 Flash scores 27.5\% in electromagnetism versus 0.0--15.0\% across its other four domains; Qwen3.8 27B scores 25.0\% versus 0.0--17.5\%. For Sol, the other four domains reach 62.5--80.0\%, while electromagnetism is its weakest domain at 57.5\%. Terra likewise scores 52.5--85.0\% in the other domains but only 37.5\% in electromagnetism. These results show that capability gains are uneven across domains: weaknesses of less capable models need not be weaknesses of stronger models. Different models emphasize different capabilities and excel in different domains, while an individual model's understanding of physical mechanisms is itself uneven.

\input{table_main}

\textbf{Multiple events challenge abductive reasoning.}\ All 27 VLM configurations achieve lower exact-match accuracy on six-event questions than on single-event questions (Table~\ref{tab:main}). For example, Sol drops from 83.7\% to 68.0\%, Gemini from 60.5\% to 21.5\%, and MiniMax M3.1 Flash from 38.4\% to 12.5\%. This decline is likely related to the persistence and overlap of earlier effects as the number of events increases. Models must distinguish the response to the current operation from residual effects of previous operations and account for their combined influence. These demands place greater pressure on the ability to apply physical laws and to extract and integrate information across multiple observations. A mistaken attribution at one step can alter the assumed state from which later responses are interpreted. Successive decisions are therefore coupled: the model must preserve a consistent account of which influences remain active and which have changed. An explanation that fits one observation may become inconsistent when applied to another step. Resolving such conflicts requires checking the proposed event history against the sequence as a whole. This offers one possible route for errors to accumulate in longer sequences.

\textbf{GPT-6.1 marks a substantial advance in physical reasoning.}\ Sol and Terra achieve six-event accuracies of 68.0\% and 65.0\%, exceeding GPT-6 Luna, the strongest remaining configuration, by 39.5 and 36.5 percentage points, respectively (Table~\ref{tab:main}). Both scores are more than twice its 28.5\%. Their lead extends to every question length and all five physical domains, where they consistently occupy the top two positions (Table~\ref{tab:domains}). These results indicate that newer generations of large models are steadily improving their understanding and application of physical laws.

\subsection{Impact of Training}
\label{sec:training}
The training set contains 9,000 episodes used for SFT. The validation set contains 1,000 episodes, of which 100 are used to select preference-training checkpoints. The test set contains 200 episodes and yields 1,180 questions for final evaluation. Task-specific learning substantially improves Qwen3.5-9B's sequence accuracy (Table~\ref{tab:methods}). SFT raises Exact$_6$ from 0.5\% to 20.0\%, and SFT + SimPO reaches 31.8\% averaged over two seeds. SFT + SimPO also raises Overall accuracy from the base model's 5.4\% to 57.2\%.

\textbf{Training-free methods yield mixed gains.}\ ReAct raises Gemini's Exact$_6$ from 21.5\% to 26.0\%, while Visual Sketchpad and ViperGPT reduce it to 8.0\% and 11.0\%, respectively (Table~\ref{tab:methods}). Access to additional visual tools therefore does not automatically produce more accurate answers.

\textbf{Single-event and sequence gains can diverge.}\ On single-step questions, SFT + KTO improves accuracy from SFT's 78.9\% to 86.3\%, yet its six-step accuracy falls from 20.0\% to 14.5\% (Table~\ref{tab:methods}). SFT + SimPO achieves a similar single-step accuracy of 86.8\%, but its six-step accuracy of 31.8\% is more than twice that of KTO. Because both methods start from the SFT model, the decrease with KTO shows that an additional optimization stage can weaken a capability already acquired during supervised learning. Its simultaneous improvement on single-event questions is consistent with a possible shift in the balance between recognizing an individual response and maintaining an explanation across events. One possible account is that preferences favor some locally convincing decisions without equally reinforcing their compatibility with the surrounding sequence. Such a shift would help isolated judgments while making longer interpretations less stable.

\input{table_methods}

\begin{figure}[t]
\centering
\includegraphics[width=\linewidth]{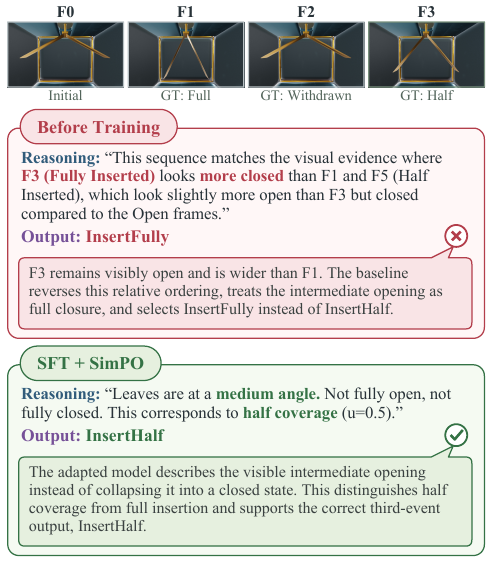}
\caption{\textbf{Training-Case Comparison for Qwen3.5-9B.} Inputs are identical; bubbles show our~analysis.}
\label{fig:training_case}
\end{figure}

\begin{figure*}[t]
\centering
\includegraphics[width=.95\textwidth]{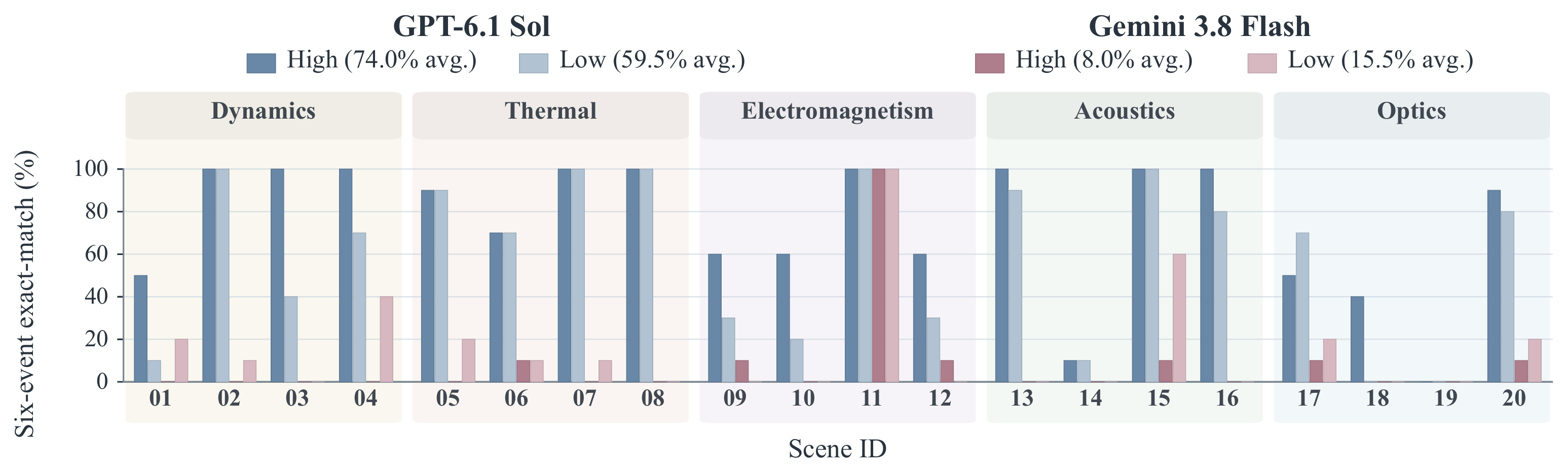}
\caption{\textbf{Effect of Reasoning Effort.} Each condition uses the same ten episodes per scene; scores include conditional answer completion.}
\label{fig:effort}
\end{figure*}

\textbf{Knowing physical laws does not guarantee correct inference.}\ The baseline knows that inserting the dielectric reduces the opening of the metal leaves, but interprets the partly open leaves in $F_3$ as evidence of full insertion (Figure~\ref{fig:training_case}). SFT + SimPO notices that the leaves have not fully closed and selects half insertion. Both explanations invoke the correct physical law, but the latter applies it more accurately to what is visible in the image. The same physical relation can explain responses of different degrees, so recognizing the expected change does not settle which operation occurred. The remaining judgment depends on whether the particular state shown in the image supports that choice. A correct account of the general relationship can therefore coexist with an incorrect interpretation of the observed event.

\subsection{Impact of Reasoning Effort}
We compare Low and High reasoning effort for Sol and Gemini on the same 200 six-step questions, separately from the main model comparison. Prompts, images, and initial output-token limits remain fixed within each model. Figure~\ref{fig:effort} compares scene-level Exact$_6$ at Low and High effort.

\textbf{The best effort level depends on the model.}\ Sol's Exact$_6$ rises from 59.5\% at Low to 74.0\% at High, whereas Gemini's falls from 15.5\% to 8.0\% (Figure~\ref{fig:effort}). These opposing trends show that higher reasoning effort does not consistently improve accuracy. Its benefit depends on how effectively each model uses the additional deliberation to reach correct judgments.

\textbf{Reasoning gains vary across physical mechanisms.}\ For driven pendulums (S03), Sol's accuracy rises from 40\% to 100\%, while for photoelastic load (S17), it falls from 70\% to 50\% (Figure~\ref{fig:effort}). The same increase in effort can therefore improve one apparatus and impair another. One possible explanation is that greater reasoning effort can draw attention to more image details without ensuring that the model interprets them correctly. If these misinterpretations enter subsequent reasoning, they can interfere with the final judgment.

\subsection{Error Analysis}
We compare model explanations and final answers with the input images, apparatus descriptions, and reference controls, and identify four recurring error types.

\textbf{Visual Identification Errors.}\ Models miss visible changes, attribute them to the wrong component, or judge distinct response patterns as unchanged. These failures occur when a component is not consistently tracked across images or its change is not assessed against the available references. The resulting evidence is incomplete or incorrectly assigned, so subsequent reasoning may omit an applied control or attribute the response to a different one.

\textbf{Physical Inference Errors.}\ Even when a visible change is recognized, models may apply an incorrect relation between the response and its cause. They may equate the direction of a response with that of the control, or apply one mapping to components with different physical roles. Such shortcuts disregard how the apparatus transforms external inputs into internal responses, so an accurate description of the observation can still lead to an incorrect physical explanation.

\textbf{Evidence Integration Errors.}\ Models fail to reconcile observations across frames and components with the event history. They may associate a response with the wrong step, introduce unsupported response delays, or treat a persisting effect as evidence of a new control. Earlier effects can persist while several controls contribute to the current response, so assigning a cause requires checking the observations together. A control that appears plausible from one image may otherwise contradict the remaining evidence or be assigned to the wrong event.

\textbf{Action Composition Errors.}\ Models produce incomplete or inconsistent control sets. One pattern is to assume that each event contains only one action even when simultaneous actions are allowed, excluding valid combinations before forming an answer. Another is to omit an action explicitly identified as applied in the explanation, or include an action that contradicts that explanation. The final answer can therefore lose a correctly identified cause, leaving part of the response unexplained or conflicting with the model's own judgments.

\section{Conclusion}

We introduce CaveBench to evaluate abductive physical reasoning from internal observations. The benchmark spans 20 apparatus scenes in five physical domains and contains 10,200 episodes. We evaluate 27 VLM configurations on 1,180 test questions and compare seven adaptation baselines. The strongest proprietary and open-weight models achieve six-event exact-match accuracies of 68.0\% and 12.5\%, respectively. Capability gains differ across physical domains, and improvements on single-event questions do not always extend to complete sequences. Error analysis identifies remaining difficulties in reading visual evidence, applying physical laws, integrating observations across time, and inferring complete combinations of controls (\hyperref[app:error-analysis]{Appendix~\ref*{app:error-analysis}}).\label{page:body_end}

%% file: table_domains.tex
\begin{table}[!t]
\centering
\caption{Six-Event Exact-Match Accuracy across Domains~(\%).}
\label{tab:domains}
\footnotesize
\renewcommand{\arraystretch}{1.23}
\setlength{\tabcolsep}{0pt}
\begin{tabular}{@{}>{\raggedright\arraybackslash}p{80pt}*{6}{>{\centering\arraybackslash}p{\dimexpr(\linewidth-80pt)/6\relax}}@{}}
\toprule
Model & Dyn. & Therm. & EM & Acous. & Opt. & \makebox[\linewidth][r]{Overall} \\
\midrule
\rowcolor{tablegroup}
\multicolumn{7}{l}{\textbf{Proprietary Models}} \\
GPT-6.1 Sol & \underline{70.0} & \underline{80.0} & \cellcolor{tablebest}\textbf{57.5} & \underline{70.0} & \cellcolor{tablebest}\textbf{62.5} & \cellcolor{tablebest}\textbf{68.0} \\
GPT-6.1 Terra & \cellcolor{tablebest}\textbf{75.0} & \cellcolor{tablebest}\textbf{85.0} & \underline{37.5} & \cellcolor{tablebest}\textbf{75.0} & \underline{52.5} & \underline{65.0} \\
GPT-6 Luna & 52.5 & 30.0 & 22.5 & 27.5 & 10.0 & 28.5 \\
Gemini 3.8 Flash & 40.0 & 7.5 & 32.5 & 15.0 & 12.5 & 21.5 \\
Claude Sonnet 5.5 & 17.5 & 5.0 & 30.0 & 17.5 & 12.5 & 16.5 \\
GPT-5.6 Luna & 42.5 & 10.0 & 22.5 & 17.5 & 5.0 & 19.5 \\
Gemini 3.6 Flash & 25.0 & 7.5 & 25.0 & 17.5 & 10.0 & 17.0 \\
Claude Haiku 5.5 & 20.0 & 17.5 & 22.5 & 0.0 & 7.5 & 13.5 \\
Qwen3.8 Flash & 22.5 & 7.5 & 25.0 & 10.0 & 5.0 & 14.0 \\
Qwen3.7 Plus & 10.0 & 5.0 & 27.5 & 5.0 & 10.0 & 11.5 \\
Seed-2.0-Lite & 0.0 & 0.0 & 27.5 & 0.0 & 0.0 & 5.5 \\
Claude Sonnet 5 & 10.0 & 0.0 & 5.0 & 0.0 & 0.0 & 3.0 \\
Seed 2.1 Turbo & 0.0 & 0.0 & 17.5 & 0.0 & 2.5 & 4.0 \\
GLM 5V Turbo & 5.0 & 0.0 & 7.5 & 0.0 & 0.0 & 2.5 \\
Seed-2.0-Mini & 0.0 & 0.0 & 15.0 & 0.0 & 0.0 & 3.0 \\
Gemini 3.5 Flash Lite & 2.5 & 2.5 & 0.0 & 0.0 & 0.0 & 1.0 \\
\midrule
\rowcolor{tablegroup}
\multicolumn{7}{l}{\textbf{Open-Weight Models}} \\
MiniMax M3.1 Flash & 15.0 & 0.0 & 27.5 & 12.5 & 7.5 & 12.5 \\
GLM 5.3 Flash & 15.0 & 0.0 & 17.5 & 5.0 & 7.5 & 9.0 \\
Qwen3.8 27B & 17.5 & 2.5 & 25.0 & 0.0 & 7.5 & 10.5 \\
DeepSeek V4.1 Flash & 5.0 & 5.0 & 20.0 & 0.0 & 2.5 & 6.5 \\
MiniMax M3 & 2.5 & 0.0 & 20.0 & 2.5 & 0.0 & 5.0 \\
Gemma 4 31B & 5.0 & 0.0 & 15.0 & 2.5 & 0.0 & 4.5 \\
Kimi K2.6 & 0.0 & 0.0 & 15.0 & 0.0 & 2.5 & 3.5 \\
Kimi K2.5 & 0.0 & 0.0 & 10.0 & 0.0 & 0.0 & 2.0 \\
Llama 4 Maverick & 2.5 & 0.0 & 0.0 & 0.0 & 0.0 & 0.5 \\
Qwen3.5-9B & 0.0 & 0.0 & 2.5 & 0.0 & 0.0 & 0.5 \\
DeepSeek V4 Flash & 0.0 & 0.0 & 2.5 & 0.0 & 0.0 & 0.5 \\
\bottomrule
\end{tabular}
\end{table}

%% file: table_main.tex
\begin{table}[!t]
\centering
\caption{Exact-Match Accuracy by Question Length (\%). $k$ denotes the number of events.}
\label{tab:main}
\footnotesize
\renewcommand{\arraystretch}{1.23}
\setlength{\tabcolsep}{0pt}
\begin{tabular}{@{}>{\raggedright\arraybackslash}p{80pt}*{7}{>{\centering\arraybackslash}p{\dimexpr(\linewidth-80pt)/7\relax}}@{}}
\toprule
Model & $k{=}1$ & $k{=}2$ & $k{=}3$ & $k{=}4$ & $k{=}5$ & $k{=}6$ & \makebox[\linewidth][r]{Overall} \\
\midrule
\rowcolor{tablegroup}
\multicolumn{8}{l}{\textbf{Proprietary Models}} \\
GPT-6.1 Sol & \cellcolor{tablebest}\textbf{83.7} & \cellcolor{tablebest}\textbf{75.3} & \cellcolor{tablebest}\textbf{76.5} & \cellcolor{tablebest}\textbf{71.0} & \cellcolor{tablebest}\textbf{70.5} & \cellcolor{tablebest}\textbf{68.0} & \cellcolor{tablebest}\textbf{74.1} \\
GPT-6.1 Terra & \underline{81.1} & \underline{74.7} & \underline{71.5} & \underline{67.5} & \underline{66.5} & \underline{65.0} & \underline{70.9} \\
GPT-6 Luna & 57.9 & 43.2 & 36.5 & 31.5 & 27.0 & 28.5 & 37.2 \\
Gemini 3.8 Flash & 60.5 & 38.4 & 36.0 & 28.5 & 25.5 & 21.5 & 34.8 \\
Claude Sonnet 5.5 & 58.9 & 47.9 & 31.0 & 29.0 & 24.5 & 16.5 & 34.3 \\
GPT-5.6 Luna & 56.3 & 40.5 & 33.0 & 22.0 & 19.5 & 19.5 & 31.5 \\
Gemini 3.6 Flash & 53.2 & 33.2 & 24.5 & 22.0 & 20.5 & 17.0 & 28.1 \\
Claude Haiku 5.5 & 51.1 & 29.5 & 29.0 & 20.0 & 13.5 & 13.5 & 25.8 \\
Qwen3.8 Flash & 43.7 & 29.5 & 22.0 & 19.0 & 15.5 & 14.0 & 23.7 \\
Qwen3.7 Plus & 46.8 & 24.2 & 19.0 & 11.5 & 11.0 & 11.5 & 20.4 \\
Seed-2.0-Lite & 42.6 & 22.1 & 17.5 & 10.5 & 7.0 & 5.5 & 17.3 \\
Claude Sonnet 5 & 35.3 & 16.8 & 11.0 & 6.5 & 4.0 & 3.0 & 12.5 \\
Seed 2.1 Turbo & 38.4 & 13.2 & 10.5 & 6.5 & 4.0 & 4.0 & 12.5 \\
GLM 5V Turbo & 31.6 & 13.2 & 9.5 & 6.0 & 5.0 & 2.5 & 11.1 \\
Seed-2.0-Mini & 30.5 & 14.2 & 7.5 & 3.5 & 4.5 & 3.0 & 10.3 \\
Gemini 3.5 Flash Lite & 30.5 & 10.0 & 8.0 & 3.0 & 2.5 & 1.0 & 9.0 \\
\midrule
\rowcolor{tablegroup}
\multicolumn{8}{l}{\textbf{Open-Weight Models}} \\
MiniMax M3.1 Flash & 38.4 & 23.7 & 23.0 & 13.5 & 13.5 & 12.5 & 20.6 \\
GLM 5.3 Flash & 42.1 & 23.7 & 21.0 & 13.0 & 9.5 & 9.0 & 19.5 \\
Qwen3.8 27B & 35.3 & 22.1 & 17.0 & 12.0 & 12.0 & 10.5 & 18.0 \\
DeepSeek V4.1 Flash & 37.4 & 17.9 & 15.0 & 12.0 & 11.5 & 6.5 & 16.5 \\
MiniMax M3 & 30.5 & 14.7 & 8.5 & 6.0 & 2.0 & 5.0 & 10.9 \\
Gemma 4 31B & 27.9 & 10.0 & 7.5 & 7.0 & 6.5 & 4.5 & 10.4 \\
Kimi K2.6 & 34.2 & 5.8 & 6.5 & 4.0 & 4.5 & 3.5 & 9.6 \\
Kimi K2.5 & 25.3 & 6.8 & 5.5 & 3.5 & 3.0 & 2.0 & 7.5 \\
Llama 4 Maverick & 20.0 & 4.7 & 4.5 & 3.0 & 0.5 & 0.5 & 5.4 \\
Qwen3.5-9B & 23.7 & 3.2 & 4.0 & 1.0 & 1.0 & 0.5 & 5.4 \\
DeepSeek V4 Flash & 21.1 & 2.1 & 3.5 & 2.0 & 1.5 & 0.5 & 5.0 \\
\bottomrule
\end{tabular}
\end{table}

%% file: table_methods.tex
\begin{table}[!t]
\centering
\caption{Exact-Match Accuracy of Training-Based and Training-Free Methods (\%). $k$ denotes the number of events.}
\label{tab:methods}
\footnotesize
\setlength{\tabcolsep}{0pt}
\begin{tabular}{@{}>{\raggedright\arraybackslash}p{60pt}*{7}{>{\centering\arraybackslash}p{\dimexpr(\linewidth-60pt)/7\relax}}@{}}
\toprule
Method & $k{=}1$ & $k{=}2$ & $k{=}3$ & $k{=}4$ & $k{=}5$ & $k{=}6$ & Overall \\
\midrule
\rowcolor{tablegroup}
\multicolumn{8}{l}{\textit{Training-Based (Qwen3.5-9B)}} \\
Baseline & 23.7 & 3.2 & 4.0 & 1.0 & 1.0 & 0.5 & 5.4 \\
SFT & 78.9 & 62.6 & 47.5 & 35.0 & 21.0 & 20.0 & 43.7 \\
SFT + DPO & 86.3 & 74.2 & 69.0 & 52.5 & 29.0 & 30.5 & 56.5 \\
SFT + SimPO & 86.8 & 72.6 & 65.8 & 56.0 & 32.5 & 31.8 & 57.2 \\
SFT + KTO & 86.3 & 74.7 & 68.0 & 38.2 & 15.8 & 14.5 & 49.1 \\
\midrule
\rowcolor{tablegroup}
\multicolumn{8}{l}{\textit{Training-Free (Gemini 3.8 Flash)}} \\
Baseline & 60.5 & 38.4 & 36.0 & 28.5 & 25.5 & 21.5 & 34.8 \\
ReAct & 64.2 & 47.9 & 37.5 & 34.0 & 26.5 & 26.0 & 39.1 \\
Visual Sketchpad & 59.5 & 34.2 & 20.0 & 15.5 & 10.0 & 8.0 & 24.2 \\
ViperGPT & 57.4 & 28.4 & 25.0 & 18.0 & 12.0 & 11.0 & 25.0 \\
\bottomrule
\end{tabular}
\end{table}

%% file: appendix-content.tex
\appendix
\numberwithin{equation}{section}
\input{protocol}
\input{detailed_results}
\input{error_analysis}
\section{Representative CaveBench Samples from Each Category}\label{app:samples}
Figures~\ref{promptpage:S01}--\ref{promptpage:S20} give one six-event question for each apparatus, in the order used in the results tables. The leftmost image shows the initial state. Each subsequent image belongs to the next intervention, so neighboring images bracket a single event. Below the images, the illustrated prompt introduces the apparatus and its reference frame, explains which components respond to a control, and specifies the choices available at each event. The answer beneath the prompt groups controls that were applied together.
\input{case-prompts}
\section{Case Studies}\label{app:cases}
Figures~\ref{platepage:S01}--\ref{platepage:S20} follow the same episodes as Section~\ref{app:samples}. Each panel shows the reference sequence beneath the images, followed by the name of the Qwen model, a continuous passage from its recorded response, and the controls it submitted. The count at the foot of the panel gives the number of events that match the reference at the same position. Readers can compare an observation described in the passage with the corresponding image and check the controls chosen for that interval.
\input{cases}

%% file: protocol.tex
\section{Appendix Overview and Organization}\label{app:overview}
The appendix first describes how a physical question is constructed, from the choice of apparatus to the record of the controls applied. It then reports the experiments and presents the questions alongside the models' answers. Sections~\ref{app:comparison}--\ref{app:cases} can be read in sequence, or an apparatus identifier can be used to move from its aggregate scores to one illustrated test episode.

\noindent\textbf{(1) Benchmark Comparison} (Section~\ref{app:comparison}). We compare the evidence supplied to a model and the kind of answer required in physical prediction, causal question answering, and intervention reconstruction.

\noindent\textbf{(2) Interpreting Apparatus Responses} (Section~\ref{app:constraints}). We describe how the observation frame, the apparatus dynamics, and its control mechanism constrain an event history.

\noindent\textbf{(3) Dataset Statistics} (Section~\ref{app:statistics}). We give the distribution of test questions and single- and paired-label events across physical domains.

\noindent\textbf{(4) Data Curation} (Section~\ref{app:curation}). We follow one episode through simulation, image capture, and control annotation. The description then shows how to recover the inputs and reference answer for any question derived from that episode.

\noindent\textbf{(5) Implementation and Results} (Section~\ref{app:results}). We specify what counts as a correct event or sequence and how each model receives the observations. The tables report scores by apparatus at each question length, followed by the individual preference-training runs and the reasoning-effort experiment. Section~\ref{app:error-analysis} examines four types of error in recorded model answers.

\noindent\textbf{(6) Representative Samples} (Section~\ref{app:samples}). For every apparatus, an illustrated question places its internal observations alongside the prompt and reference answer.

\noindent\textbf{(7) Case Studies} (Section~\ref{app:cases}). Each example is paired with the recorded response of a model so that its interpretation and final choices can be read against the same observations.

\section{Comparison with Other Physical Reasoning Benchmarks}\label{app:comparison}
Table~\ref{tab:benchmark-context} compares the evidence available to a model with the answer each benchmark scores. Consider a collision: predicting that two objects will meet, explaining why they met, and choosing an action that would bring them together all require physical reasoning. Yet a correct prediction leaves open which earlier event the model regards as its cause.

Physion~\cite{physion} and IntPhys 2019~\cite{intphys} both study intuitive physics through object motion. Physion shows the beginning of a simulated interaction and asks whether two designated objects will touch later. A falling object might hit a support, for instance, or pass it on the way down. The same contact judgment can therefore test different kinds of motion. IntPhys 2019 instead presents matched videos and asks for a plausibility score for each. In its object-permanence task, disappearance behind a screen can be physically consistent. What matters is whether the objects seen when the screen moves are consistent with those seen before occlusion. QuantiPhy~\cite{quantiphy} tests a third ability. Given a video and a known physical quantity such as a ball's diameter, the model can use that diameter as a scale to estimate speed from displacement over time. Here the numerical error matters even if the direction of motion is clear.

CLEVRER~\cite{clevrer} and ComPhy~\cite{comphy} both use collisions to ask about causes, but they provide different evidence for the answer. In CLEVRER, a question can designate one collision and ask which earlier events contributed to it. A counterfactual question can remove an object and ask which collisions would still occur. The relevant evidence lies in how encounters within that scene depend on one another. ComPhy pairs a target video with four reference videos. Each target object appears in at least one reference video and retains its mass and charge across appearances. The way two objects accelerate toward or away from each other gives evidence about charge. The change in their motion on impact gives evidence about relative mass. A property inferred in one encounter can then help explain or predict another encounter in the target video.

BlackSwanSuite~\cite{blackswan} directly examines reasoning from an outcome back to an unobserved event. In its Detective task, the beginning and end of a real video are visible and the middle is withheld. A model may describe what happened in the missing segment, choose among proposed accounts, or judge whether an earlier hypothesis still fits after seeing the ending. The ending matters because it can rule out a continuation that seemed plausible from the opening alone. Because the missing segment is the unit of explanation, the same video can support several descriptions that account for the outcome.

The role of the intervention itself distinguishes PHYRE~\cite{phyre} from CaveBench. In PHYRE an agent sees an initial configuration and a desired outcome, then places one or more objects in the scene. Any placement that achieves the goal when the simulator runs can solve the puzzle. CaveBench asks which interventions were actually applied in a recorded experiment. The apparatus description gives the available controls and how they act on the system. A camera inside the apparatus records an initial state and the responses that follow. The controls themselves remain outside its view. An effect can persist after the control that produced it, and two controls applied together can leave different traces in the same image. The model must decide what changed in each interval, using the preceding observations to interpret the current one. Its answer is an ordered sequence of control sets. The simulator record distinguishes an omitted control from a correctly identified control assigned to the wrong interval.

\begin{table}[H]\centering\small
\caption{Visual Evidence and Inference Targets in Representative Benchmarks.}\label{tab:benchmark-context}
\begin{tabularx}{\linewidth}{@{}l>{\raggedright\arraybackslash}X>{\raggedright\arraybackslash}X@{}}\toprule
\textbf{Benchmark} & \textbf{Evidence supplied} & \textbf{Inference target}\\\midrule
Physion~\cite{physion} & Initial segment of a simulated interaction & Whether two designated objects will touch later \\
IntPhys 2019~\cite{intphys} & Matched clips with possible or impossible events & Physical plausibility of each clip \\
QuantiPhy~\cite{quantiphy} & Video and a physical quantity supplied as context & A requested measurement, such as speed or size \\
CLEVRER~\cite{clevrer} & Video of moving and colliding objects & Descriptive, explanatory, predictive, or counterfactual event questions \\
ComPhy~\cite{comphy} & Target video and four reference clips with recurring objects & Hidden properties such as mass and charge, and their effects on events \\
BlackSwanSuite~\cite{blackswan} & Video before and after an unseen event (Detective task) & The unexpected event connecting the visible segments \\
PHYRE~\cite{phyre} & Initial arrangement and a specified goal in a simulator & A placement of objects that achieves the goal \\
\textbf{CaveBench} & Apparatus description and ordered internal observations & All applied external controls, grouped by interval \\
\bottomrule\end{tabularx}\end{table}

\section{Reference Frames, State Persistence, and Control Combinations}\label{app:constraints}
\textbf{Observation Frames.} To interpret apparent motion, first establish how the camera moves relative to the apparatus. The camera in \hyperref[promptpage:S04]{S04} is attached to the tray. When a hammer strikes one edge, the tray and camera recoil together while the loose objects lag behind. The rim stays fixed in the image, so displacement of a ball or drum relative to that rim helps identify the struck edge. \hyperref[promptpage:S17]{S17} requires a different change of viewpoint. The direction of a polarizer turn is named from outside the right wall, facing inward. A clockwise turn from that position increases the optical angle measured counterclockwise from image-right in the front view.

\textbf{Persistence of the Physical State.} The response to one command can still be visible after the next command takes effect. On the heated plate in \hyperref[promptpage:S05]{S05}, the coating cools more slowly than it warms, and the bands farther from the camera respond more slowly. A patch of colour can linger where the mobile heater has already passed. To trace the heater's movement, follow where fresh warming appears relative to the seams and the stationary heaters. In \hyperref[promptpage:S18]{S18}, persistence has a mechanical origin. Each mirror holds its orientation until the next turn of that axis. The preceding pattern of coloured spots on the receiver therefore gives the starting positions for measuring the effects of the latest turn.

\textbf{Composition of Controls.} The controls available on an apparatus determine which labels can belong to the same event. In \hyperref[promptpage:S06]{S06} a valve connects a temperature supply to a jacket surrounding either liquid layer. Its setting requires both the selected layer and whether the supply heats or cools it. In \hyperref[promptpage:S20]{S20} two mechanisms can move independently. A change in slit spacing shifts the fine interference fringes closer together or farther apart, while slit width controls the broader diffraction envelope. Changes at both scales can support two simultaneous adjustments. In \hyperref[promptpage:S17]{S17} each event changes either the load or the polarizer angle. A reference strip outside the load path responds to rotation alone and helps distinguish the two.

\section{Dataset Statistics}\label{app:statistics}
Each recorded episode begins with an initial image, followed by six images of the same apparatus after successive interventions. Each of the 20 apparatuses contributes 450 training, 50 validation, and 10 test episodes. This yields 9,000 training episodes, 1,000 validation episodes, and 200 test episodes. The test episodes supply 1,400 recorded frames. Because longer questions reuse the opening frames of the same experiment, the question records contain 5,350 image entries in total.

\begin{table}[H]\centering\small
\caption{Test Questions and Recorded Events by Physical Domain.}\label{tab:dataset-statistics}
\begin{tabular*}{\linewidth}{@{\extracolsep{\fill}}llrrr@{}}\toprule
Domain & Apparatuses & Questions & One-label events & Two-label events\\\midrule
Dynamics & \hyperref[promptpage:S01]{S01}--\hyperref[promptpage:S04]{S04} & 240 & 140 & 100\\
Thermal physics & \hyperref[promptpage:S05]{S05}--\hyperref[promptpage:S08]{S08} & 240 & 68 & 172\\
Electromagnetism & \hyperref[promptpage:S09]{S09}--\hyperref[promptpage:S12]{S12} & 220 & 203 & 37\\
Acoustics & \hyperref[promptpage:S13]{S13}--\hyperref[promptpage:S16]{S16} & 240 & 108 & 132\\
Optics & \hyperref[promptpage:S17]{S17}--\hyperref[promptpage:S20]{S20} & 240 & 179 & 61\\\midrule
Total & & 1,180 & 698 & 502\\\bottomrule
\end{tabular*}\end{table}

Table~\ref{tab:dataset-statistics} distinguishes the number of questions from the number of recorded interventions. Every apparatus contributes 60 test interventions, giving the same denominator of 240 per domain. Of the 1,200 interventions, 502 carry two labels. Their frequency varies with the controls available. Two labels occur in 71.7\% of thermal events and 15.4\% of electromagnetic events. A comparison between domains at the same question length thus holds the number of event positions fixed while allowing the number of labels to recover at those positions to differ.

For 19 apparatuses each episode supplies a question at every length from one to six events. \hyperref[promptpage:S11]{S11} uses the responses to full insertion, half insertion, and withdrawal of a dielectric as visual references for one another. Beginning its questions at three events keeps all three reference states in view without naming their order in the prompt. There are consequently 190 questions at each of the first two lengths and 200 at each remaining length, for a total of 1,180.

\section{Details of the Data Curation Process}\label{app:curation}
\subsection{Dataset Sources}
Each episode is a continuous run of one Unity apparatus. The generator selects a six-event schedule from the controls available on that apparatus and varies the component layout, physical parameters, viewpoint, and lighting across runs. The controls act outside the interior camera's field of view. A synchronized recording lasting roughly 30--40 seconds places the external controls on the left and the interior view on the right.

The initial image is captured before any control acts. The next six images are sampled at points in the response appropriate to the apparatus. In the vehicle cabin, the capture window preserves the transient displacement of the suspended and freely moving objects. In the dielectric apparatus, the image is taken as the leaves settle, so that their opening can be compared across settings. Both procedures associate one interior observation with each external event while allowing the apparatus to evolve continuously between captures.

We select the ten test episodes for each apparatus from its completed runs. The selection favors six-event sequences that are least represented among those already chosen. Among eligible runs it compares the initial layouts, the positions at which their controls differ, and features such as paired controls and switches between successive steps. After the first selection, these three criteria carry weights of 45\%, 35\%, and 20\%, respectively. Ties are settled by episode identifier. We fix these episode identities before deriving questions of different lengths, so comparisons across lengths refer back to the same set of experiments wherever both lengths apply.

\subsection{Data Annotation and Quality Control}
The simulator records which controls it executes and associates them with the corresponding capture. For an event that changes two settings, the two labels share one event index in the record. This gives a direct correspondence between an entry in the control log and a pair of neighboring interior images, even when one control changes several visible components. In the synchronized video, the motion of the external mechanism and the internal response can be inspected side by side at the same point in the experiment.

Three checks connect the simulation record to the released evaluation inputs. For each question, we verify that the image list starts with the initial state, contains one further image for each requested event, and agrees with the episode prefix. For each reference event, we check the complete group of labels against the combinations allowed by that apparatus. Finally, file hashes connect the fixed test set to the original recordings. All 1,400 interior test images match their source files, and the original control records for all 200 test episodes match the records selected for evaluation.

\subsection{Data Format and Question Generation}
The episode is the unit of storage and the question is the unit of evaluation. An episode record holds the ordered image paths, the dual-view video, and the full control history. A length-$k$ question appends \texttt{/k} and the length to the episode identifier. For example, the three-event version uses the initial state and the first three response images. Its reference contains the first three control sets from the same log. The six-event version reuses those images and adds the later responses. Each question is presented separately with its own prompt and image list.

The question file stores the evaluated prompt in \texttt{prompt}, the interior views in \texttt{images}, the apparatus vocabulary in \texttt{allowed\_labels}, and its permitted event sets in \texttt{candidates}. A separate answer file uses the same question ID and stores the ordered list of label sets under \texttt{operations}. The field \texttt{paper\_scene} gives the apparatus number used in the figures and results tables. The field \texttt{scene} retains its simulator number. Together these identifiers allow a question, its images, the recorded controls, and the model prediction to be traced to the same run.

\section{Additional Implementation Details and Experimental Results}\label{app:results}
\subsection{Evaluation Metrics}
An event prediction is an unordered set of control labels. We compare it with the recorded set at the same event position. A missing label, an extra label, or a control assigned to a neighbouring interval makes that event incorrect. The parser checks that the response supplies exactly the requested number of event sets and that each set is permitted by the apparatus. An unparseable final response receives zero credit at every position.

Sequence exact match counts a question as correct only when every event matches. Event accuracy counts matching event sets across the same questions. For instance, recovering five of six event sets earns credit at five positions but contributes no correct sequence. At a fixed question length, the denominator for sequence accuracy is the number of questions evaluated. The Overall column in the training comparison pools the questions at all lengths before computing accuracy.

\subsection{Benchmark Models}
The main comparison covers 16 proprietary and 11 open-weight model configurations. Each configuration answers all 1,180 questions, giving 31,860 scored responses. Tables~\ref{tab:matrix-k1}--\ref{tab:matrix-k6} use the same row order throughout, with entries from the same model family listed separately by configuration. This permits a comparison at a fixed apparatus and length even when two variants use different reasoning settings.

\subsection{Implementation Details for Model Evaluation}
\textbf{Inference.} Each request supplies the archived evaluation prompt and the internal images in chronological order, from the initial state through the requested event. Each question length is evaluated in its own request. The model receives the permitted controls and the apparatus description. The external camera recording and reference controls are used to construct and check the questions.

Fourteen configurations use Medium effort, including the GPT, Gemini and Claude configurations, Qwen3.8 27B, MiniMax M3.1 Flash, and the Seed Lite and Mini variants. GLM 5.3 Flash and the two DeepSeek Flash configurations use High effort. Nine configurations enable thinking with model-selected effort, and Llama 4 Maverick uses the provider default. The controlled effort experiment fixes the initial output limit at 8,192 tokens for Sol and 6,144 for Gemini across its three settings. If a completed generation has no usable answer, it allows one Low-effort completion request of up to 2,048 tokens.

\textbf{Adaptation.} For Qwen3.5-9B~\cite{qwen35}, supervised fine-tuning first uses full six-event sequences and then continues with prefixes balanced across lengths. Each stage uses 9,000 examples. We train a rank-16 LoRA~\cite{lora} adapter with scaling parameter 32, dropout 0.05, learning rate $10^{-4}$ and effective batch size 16. The vision encoder and visual projector remain fixed. Both stages supervise the answer tokens.

DPO~\cite{dpo}, SimPO~\cite{simpo} and KTO~\cite{kto} each start from the same supervised adapter. DPO and SimPO use 1,076 preference pairs. KTO uses 5,475 individually labeled responses. Each preference run uses two epochs, learning rate $5\times10^{-6}$ and effective batch size 16. Checkpoint selection uses 100 validation episodes. The training-free methods use Gemini 3.8 Flash. ReAct~\cite{react} and Visual Sketchpad~\cite{sketchpad} can request successive views derived from the input images, while ViperGPT~\cite{vipergpt} executes a single program that combines visual operations and measurements.

%% file: detailed_results.tex
\subsection{Complete Results by Apparatus and Question Length}\label{sec:apparatus-scores}
Tables~\ref{tab:matrix-k1}--\ref{tab:matrix-k6} give the number of complete sequences recovered as a percentage of ten test episodes per apparatus. One correct sequence therefore changes a cell by ten percentage points. \hyperref[promptpage:S11]{S11} begins at three events. Bold and underlining identify the highest and second-highest distinct scores in each column, including ties.

The same apparatus can change rank as later observations become available. Terra solves one of ten one-event questions on two-source ripples (\hyperref[promptpage:S16]{S16}) and all ten six-event questions on those episodes. For wave scattering (\hyperref[promptpage:S14]{S14}) it solves all ten one-event questions and one six-event question. Each prefix is posed separately, so the first event recovered in a six-event answer is inferred with later observations available. It need not match the model's earlier answer to the shorter question.
\par\noindent\begin{minipage}{\linewidth}\centering
\pdfbookmark[2]{Apparatus Scores at k = 1}{matrix-k1}
\captionof{table}{Sequence Exact Match (\%) by Apparatus for $k=1$ ($N=190$).}\label{tab:matrix-k1}
{\fontsize{8.5}{9.3}\selectfont\setlength{\tabcolsep}{0pt}\renewcommand{\arraystretch}{1.0}
\begin{tabular}{@{}l@{\hspace{3pt}}*{20}{>{\centering\arraybackslash}p{\dimexpr(\linewidth-95pt)/20\relax}}@{}}\toprule
 & \multicolumn{4}{c}{Dynamics} & \multicolumn{4}{c}{Thermal} & \multicolumn{4}{c}{Electromagnetism} & \multicolumn{4}{c}{Acoustics} & \multicolumn{4}{c}{Optics} \\
\cmidrule(lr){2-5}\cmidrule(lr){6-9}\cmidrule(lr){10-13}\cmidrule(lr){14-17}\cmidrule(lr){18-21}
Model & \hyperref[promptpage:S01]{S01} & \hyperref[promptpage:S02]{S02} & \hyperref[promptpage:S03]{S03} & \hyperref[promptpage:S04]{S04} & \hyperref[promptpage:S05]{S05} & \hyperref[promptpage:S06]{S06} & \hyperref[promptpage:S07]{S07} & \hyperref[promptpage:S08]{S08} & \hyperref[promptpage:S09]{S09} & \hyperref[promptpage:S10]{S10} & \hyperref[promptpage:S11]{S11} & \hyperref[promptpage:S12]{S12} & \hyperref[promptpage:S13]{S13} & \hyperref[promptpage:S14]{S14} & \hyperref[promptpage:S15]{S15} & \hyperref[promptpage:S16]{S16} & \hyperref[promptpage:S17]{S17} & \hyperref[promptpage:S18]{S18} & \hyperref[promptpage:S19]{S19} & \hyperref[promptpage:S20]{S20} \\\midrule
\multicolumn{21}{@{}l}{\textbf{Proprietary Models}} \\
GPT-6.1 Sol & \textbf{100} & \textbf{100} & \textbf{100} & \underline{90} & \textbf{100} & \textbf{70} & \textbf{100} & \textbf{90} & \textbf{90} & \textbf{60} & --- & \textbf{100} & \textbf{100} & \textbf{100} & \textbf{100} & \textbf{40} & \underline{70} & \textbf{80} & 0 & \textbf{100} \\
GPT-6.1 Terra & \textbf{100} & \textbf{100} & \textbf{100} & \textbf{100} & \textbf{100} & \textbf{70} & \textbf{100} & \textbf{90} & \underline{80} & 10 & --- & \textbf{100} & \textbf{100} & \textbf{100} & \textbf{100} & 10 & \textbf{100} & \textbf{80} & 0 & \textbf{100} \\
GPT-6 Luna & 50 & \textbf{100} & 80 & \textbf{100} & \underline{80} & \underline{60} & \textbf{100} & 0 & 40 & 10 & --- & 60 & 70 & 80 & \underline{90} & \textbf{40} & \underline{70} & \underline{30} & 10 & 30 \\
Gemini 3.8 Flash & 60 & \textbf{100} & \underline{90} & \underline{90} & \underline{80} & \textbf{70} & 60 & 0 & \underline{80} & 10 & --- & \underline{70} & \underline{80} & \textbf{100} & 80 & 20 & 50 & 20 & 20 & 70 \\
Claude Sonnet 5.5 & \underline{90} & \underline{90} & 50 & \textbf{100} & \underline{80} & \textbf{70} & 60 & 0 & \underline{80} & 20 & --- & \textbf{100} & 60 & 80 & \underline{90} & 0 & 40 & 20 & 0 & \underline{90} \\
GPT-5.6 Luna & 60 & \underline{90} & \underline{90} & \textbf{100} & 70 & 50 & \underline{80} & 0 & 70 & 0 & --- & 60 & 60 & 80 & \underline{90} & 10 & 60 & \underline{30} & \underline{30} & 40 \\
Gemini 3.6 Flash & 20 & \textbf{100} & \textbf{100} & \textbf{100} & 30 & 50 & \textbf{100} & 10 & \underline{80} & 20 & --- & 30 & 60 & 80 & 80 & 10 & 60 & 20 & 0 & 60 \\
Claude Haiku 5.5 & 80 & \underline{90} & 70 & \textbf{100} & 70 & \underline{60} & \textbf{100} & 0 & \underline{80} & 0 & --- & 50 & 30 & 40 & 60 & 0 & 40 & 10 & \underline{30} & 60 \\
Qwen3.8 Flash & 40 & \textbf{100} & 30 & \textbf{100} & 40 & \underline{60} & 50 & 10 & 60 & 10 & --- & 30 & 60 & 60 & 70 & 10 & 50 & 0 & 10 & 40 \\
Qwen3.7 Plus & 80 & \textbf{100} & 70 & \underline{90} & 60 & \underline{60} & 60 & 0 & 70 & 10 & --- & 20 & 50 & \underline{90} & 60 & 0 & 40 & 0 & 10 & 20 \\
Seed-2.0-Lite & 50 & \underline{90} & 20 & \textbf{100} & 50 & 50 & 70 & 10 & 70 & 20 & --- & 50 & 40 & 50 & 20 & 0 & 50 & 20 & 20 & 30 \\
Claude Sonnet 5 & 40 & \underline{90} & 60 & 50 & 30 & 40 & 20 & 0 & 40 & 10 & --- & 40 & 30 & 40 & 60 & 10 & 40 & 0 & 10 & 60 \\
Seed 2.1 Turbo & 50 & 70 & 40 & 70 & 30 & 30 & 40 & 0 & 30 & 20 & --- & 20 & 60 & 60 & 40 & 20 & 50 & 10 & 10 & 80 \\
GLM 5V Turbo & 0 & 60 & 30 & 80 & 40 & 30 & \underline{80} & 0 & 30 & 10 & --- & 40 & 0 & 50 & 30 & 10 & 50 & 10 & \textbf{40} & 10 \\
Seed-2.0-Mini & 30 & 60 & 30 & 50 & 20 & 50 & 60 & 0 & 60 & 20 & --- & 40 & 30 & 30 & 10 & 10 & 50 & 0 & \underline{30} & 0 \\
Gemini 3.5 Flash Lite & 0 & 60 & 60 & 70 & 10 & \textbf{70} & 40 & 0 & 40 & 10 & --- & 10 & 10 & 70 & 30 & 20 & 30 & 0 & 10 & 40 \\
\midrule
\multicolumn{21}{@{}l}{\textbf{Open-Weight Models}} \\
MiniMax M3.1 Flash & 40 & \textbf{100} & 20 & 70 & 40 & 20 & 30 & 0 & \underline{80} & 30 & --- & 30 & 40 & 70 & 60 & 0 & 50 & \underline{30} & 10 & 10 \\
GLM 5.3 Flash & 60 & \textbf{100} & 20 & \textbf{100} & 60 & \textbf{70} & 50 & 0 & 50 & \underline{40} & --- & 20 & 20 & 50 & 30 & 0 & 60 & 10 & 10 & 50 \\
Qwen3.8 27B & 10 & \textbf{100} & 40 & \textbf{100} & 30 & 40 & 40 & 0 & 70 & 20 & --- & 10 & 10 & 50 & 40 & 0 & 40 & 20 & 20 & 30 \\
DeepSeek V4.1 Flash & 50 & \underline{90} & 20 & 70 & 30 & \underline{60} & 40 & 10 & 60 & 20 & --- & 20 & 50 & 40 & 50 & 0 & 50 & 20 & 20 & 10 \\
MiniMax M3 & 20 & \textbf{100} & 50 & 50 & 10 & 40 & 10 & 30 & 40 & 0 & --- & 10 & 20 & 60 & 60 & 0 & 40 & 0 & 20 & 20 \\
Gemma 4 31B & 20 & \underline{90} & 30 & 60 & 20 & 50 & 10 & 20 & 30 & 10 & --- & 10 & 10 & 40 & 30 & 0 & 40 & 20 & \underline{30} & 10 \\
Kimi K2.6 & 20 & \underline{90} & 50 & 50 & 30 & 30 & 60 & 20 & 30 & 10 & --- & 30 & 30 & 60 & 30 & 0 & 50 & 20 & 20 & 20 \\
Kimi K2.5 & 0 & 50 & 40 & 20 & 10 & 10 & 40 & 10 & 30 & 10 & --- & 50 & 10 & 50 & 50 & \underline{30} & 30 & 0 & 0 & 40 \\
Llama 4 Maverick & 20 & 30 & 20 & 20 & 10 & 20 & 40 & \underline{40} & 10 & 20 & --- & 30 & 20 & 20 & 40 & 20 & 10 & 0 & 10 & 0 \\
Qwen3.5-9B & 0 & 70 & 10 & 20 & 10 & 50 & 10 & 30 & 20 & 0 & --- & 0 & 20 & 40 & 30 & 10 & 60 & 10 & \underline{30} & 30 \\
DeepSeek V4 Flash & 30 & 50 & 70 & 30 & 0 & 20 & 20 & 10 & 20 & 0 & --- & 20 & 10 & 10 & 40 & 10 & 20 & 0 & 20 & 20 \\
\bottomrule\end{tabular}}
\end{minipage}\par
\par\noindent\begin{minipage}{\linewidth}\centering
\pdfbookmark[2]{Apparatus Scores at k = 2}{matrix-k2}
\captionof{table}{Sequence Exact Match (\%) by Apparatus for $k=2$ ($N=190$).}\label{tab:matrix-k2}
{\fontsize{8.5}{9.3}\selectfont\setlength{\tabcolsep}{0pt}\renewcommand{\arraystretch}{1.0}
\begin{tabular}{@{}l@{\hspace{3pt}}*{20}{>{\centering\arraybackslash}p{\dimexpr(\linewidth-95pt)/20\relax}}@{}}\toprule
 & \multicolumn{4}{c}{Dynamics} & \multicolumn{4}{c}{Thermal} & \multicolumn{4}{c}{Electromagnetism} & \multicolumn{4}{c}{Acoustics} & \multicolumn{4}{c}{Optics} \\
\cmidrule(lr){2-5}\cmidrule(lr){6-9}\cmidrule(lr){10-13}\cmidrule(lr){14-17}\cmidrule(lr){18-21}
Model & \hyperref[promptpage:S01]{S01} & \hyperref[promptpage:S02]{S02} & \hyperref[promptpage:S03]{S03} & \hyperref[promptpage:S04]{S04} & \hyperref[promptpage:S05]{S05} & \hyperref[promptpage:S06]{S06} & \hyperref[promptpage:S07]{S07} & \hyperref[promptpage:S08]{S08} & \hyperref[promptpage:S09]{S09} & \hyperref[promptpage:S10]{S10} & \hyperref[promptpage:S11]{S11} & \hyperref[promptpage:S12]{S12} & \hyperref[promptpage:S13]{S13} & \hyperref[promptpage:S14]{S14} & \hyperref[promptpage:S15]{S15} & \hyperref[promptpage:S16]{S16} & \hyperref[promptpage:S17]{S17} & \hyperref[promptpage:S18]{S18} & \hyperref[promptpage:S19]{S19} & \hyperref[promptpage:S20]{S20} \\\midrule
\multicolumn{21}{@{}l}{\textbf{Proprietary Models}} \\
GPT-6.1 Sol & \underline{60} & \textbf{100} & \textbf{70} & 70 & \underline{90} & \underline{70} & \textbf{100} & \textbf{100} & \textbf{70} & \textbf{40} & --- & \textbf{80} & \textbf{100} & \underline{60} & \underline{90} & \textbf{50} & \textbf{80} & \textbf{90} & \textbf{30} & \underline{80} \\
GPT-6.1 Terra & \textbf{80} & \textbf{100} & 40 & \underline{90} & \textbf{100} & \textbf{100} & \textbf{100} & \underline{90} & \textbf{70} & \underline{30} & --- & \textbf{80} & \textbf{100} & \textbf{70} & \textbf{100} & \underline{40} & \underline{70} & \underline{70} & 0 & \textbf{90} \\
GPT-6 Luna & 40 & \underline{90} & 50 & 60 & 60 & 30 & \underline{90} & 10 & 40 & 10 & --- & 20 & 60 & 10 & \underline{90} & 10 & 50 & 40 & 10 & 50 \\
Gemini 3.8 Flash & 40 & \underline{90} & 50 & \textbf{100} & 50 & 50 & 20 & 0 & \underline{60} & 20 & --- & 10 & 50 & 20 & 80 & 10 & 50 & 10 & 0 & 20 \\
Claude Sonnet 5.5 & 30 & \textbf{100} & 40 & \textbf{100} & 70 & \underline{70} & 60 & 0 & \underline{60} & 0 & --- & \underline{60} & 60 & 30 & \textbf{100} & 0 & \underline{70} & 10 & 0 & 50 \\
GPT-5.6 Luna & 20 & \underline{90} & \underline{60} & \underline{90} & 40 & 40 & 60 & 10 & 40 & 10 & --- & 30 & \underline{70} & 20 & 80 & 10 & 40 & 10 & 10 & 40 \\
Gemini 3.6 Flash & 20 & 80 & \textbf{70} & 80 & 20 & 50 & 60 & 10 & 40 & 0 & --- & 10 & 40 & 20 & 60 & 10 & 20 & 0 & 10 & 30 \\
Claude Haiku 5.5 & 30 & \textbf{100} & 30 & \underline{90} & 50 & 20 & 80 & 0 & 40 & 0 & --- & 10 & 10 & 10 & 0 & 0 & 20 & 0 & 0 & 70 \\
Qwen3.8 Flash & 30 & \underline{90} & 50 & 80 & 40 & 30 & 30 & 0 & 20 & 10 & --- & 10 & 40 & 10 & 40 & 0 & 30 & 0 & 10 & 40 \\
Qwen3.7 Plus & 30 & \textbf{100} & 50 & 0 & 20 & 40 & 10 & 0 & \underline{60} & 0 & --- & 20 & 30 & 20 & 20 & 0 & 40 & 0 & 0 & 20 \\
Seed-2.0-Lite & 10 & 80 & 20 & 30 & 0 & 20 & 10 & 0 & 50 & 10 & --- & 20 & 20 & 20 & 30 & 20 & 30 & 0 & 0 & 50 \\
Claude Sonnet 5 & 10 & \underline{90} & 40 & 30 & 0 & 10 & 0 & 0 & 10 & 10 & --- & 0 & 20 & 30 & 40 & 0 & 10 & 0 & 0 & 20 \\
Seed 2.1 Turbo & 0 & 30 & 0 & 10 & 0 & 40 & 30 & 0 & 20 & 0 & --- & 0 & 0 & 10 & 30 & 0 & 30 & 0 & 10 & 40 \\
GLM 5V Turbo & 10 & 60 & 50 & 10 & 10 & 0 & 20 & 0 & 10 & 0 & --- & 10 & 0 & 10 & 30 & 0 & 20 & 0 & 0 & 10 \\
Seed-2.0-Mini & 10 & 50 & 40 & 0 & 10 & 30 & 20 & 0 & 40 & 20 & --- & 0 & 0 & 0 & 10 & 10 & 10 & 0 & 10 & 10 \\
Gemini 3.5 Flash Lite & 0 & 10 & 20 & 0 & 10 & 30 & 0 & 0 & 10 & 10 & --- & 0 & 0 & 20 & 50 & 10 & 10 & 0 & 0 & 10 \\
\midrule
\multicolumn{21}{@{}l}{\textbf{Open-Weight Models}} \\
MiniMax M3.1 Flash & 0 & \underline{90} & 20 & 50 & 40 & 30 & 40 & 0 & 40 & 10 & --- & 10 & 30 & 0 & 20 & 0 & 40 & 0 & 10 & 20 \\
GLM 5.3 Flash & 30 & \underline{90} & 10 & \underline{90} & 20 & 30 & 0 & 0 & 40 & 0 & --- & 30 & 10 & 0 & 10 & 0 & 50 & 0 & 10 & 30 \\
Qwen3.8 27B & 10 & \textbf{100} & 30 & \underline{90} & 0 & 10 & 20 & 0 & 30 & 0 & --- & 0 & 20 & 10 & 20 & 10 & 30 & 0 & \underline{20} & 20 \\
DeepSeek V4.1 Flash & 30 & 60 & 20 & 40 & 0 & 50 & 0 & 0 & 20 & 10 & --- & 0 & 20 & 20 & 10 & 10 & 0 & 0 & \underline{20} & 30 \\
MiniMax M3 & 10 & 70 & 30 & 30 & 20 & 0 & 0 & 0 & 30 & 20 & --- & 0 & 20 & 10 & 20 & 0 & 0 & 0 & 10 & 10 \\
Gemma 4 31B & 10 & 40 & 10 & 20 & 0 & 10 & 0 & 0 & 20 & 0 & --- & 0 & 0 & 10 & 40 & 0 & 20 & 0 & 0 & 10 \\
Kimi K2.6 & 0 & 0 & 20 & 0 & 10 & 20 & 0 & 0 & 10 & 0 & --- & 0 & 0 & 0 & 20 & 0 & 10 & 0 & 0 & 20 \\
Kimi K2.5 & 10 & 0 & 20 & 0 & 10 & 0 & 0 & 0 & 20 & 10 & --- & 10 & 0 & 0 & 20 & 0 & 20 & 0 & 0 & 10 \\
Llama 4 Maverick & 0 & 0 & 10 & 10 & 0 & 20 & 0 & 0 & 0 & 0 & --- & 0 & 0 & 0 & 20 & 0 & 20 & 0 & 0 & 10 \\
Qwen3.5-9B & 0 & 10 & 0 & 0 & 0 & 0 & 10 & 0 & 0 & 0 & --- & 0 & 0 & 0 & 10 & 0 & 10 & 0 & 0 & 20 \\
DeepSeek V4 Flash & 0 & 10 & 0 & 0 & 0 & 10 & 0 & 0 & 0 & 0 & --- & 0 & 0 & 0 & 0 & 0 & 20 & 0 & 0 & 0 \\
\bottomrule\end{tabular}}
\end{minipage}\par
\par\noindent\begin{minipage}{\linewidth}\centering
\pdfbookmark[2]{Apparatus Scores at k = 3}{matrix-k3}
\captionof{table}{Sequence Exact Match (\%) by Apparatus for $k=3$ ($N=200$).}\label{tab:matrix-k3}
{\fontsize{8.5}{9.3}\selectfont\setlength{\tabcolsep}{0pt}\renewcommand{\arraystretch}{1.0}
\begin{tabular}{@{}l@{\hspace{3pt}}*{20}{>{\centering\arraybackslash}p{\dimexpr(\linewidth-95pt)/20\relax}}@{}}\toprule
 & \multicolumn{4}{c}{Dynamics} & \multicolumn{4}{c}{Thermal} & \multicolumn{4}{c}{Electromagnetism} & \multicolumn{4}{c}{Acoustics} & \multicolumn{4}{c}{Optics} \\
\cmidrule(lr){2-5}\cmidrule(lr){6-9}\cmidrule(lr){10-13}\cmidrule(lr){14-17}\cmidrule(lr){18-21}
Model & \hyperref[promptpage:S01]{S01} & \hyperref[promptpage:S02]{S02} & \hyperref[promptpage:S03]{S03} & \hyperref[promptpage:S04]{S04} & \hyperref[promptpage:S05]{S05} & \hyperref[promptpage:S06]{S06} & \hyperref[promptpage:S07]{S07} & \hyperref[promptpage:S08]{S08} & \hyperref[promptpage:S09]{S09} & \hyperref[promptpage:S10]{S10} & \hyperref[promptpage:S11]{S11} & \hyperref[promptpage:S12]{S12} & \hyperref[promptpage:S13]{S13} & \hyperref[promptpage:S14]{S14} & \hyperref[promptpage:S15]{S15} & \hyperref[promptpage:S16]{S16} & \hyperref[promptpage:S17]{S17} & \hyperref[promptpage:S18]{S18} & \hyperref[promptpage:S19]{S19} & \hyperref[promptpage:S20]{S20} \\\midrule
\multicolumn{21}{@{}l}{\textbf{Proprietary Models}} \\
GPT-6.1 Sol & \textbf{80} & \textbf{100} & \textbf{90} & \underline{80} & \underline{90} & \underline{70} & \textbf{100} & \textbf{100} & \textbf{80} & \textbf{30} & \textbf{100} & \textbf{90} & \textbf{100} & \textbf{30} & \underline{90} & \textbf{60} & \underline{80} & \textbf{70} & \textbf{20} & \underline{70} \\
GPT-6.1 Terra & \underline{50} & \textbf{100} & \underline{50} & \textbf{100} & \textbf{100} & \textbf{80} & \textbf{100} & \textbf{100} & 50 & \underline{10} & \textbf{100} & \underline{80} & \textbf{100} & \textbf{30} & \textbf{100} & \underline{50} & \textbf{100} & \underline{40} & 0 & \textbf{90} \\
GPT-6 Luna & 40 & \textbf{100} & 40 & \underline{80} & 20 & 30 & \underline{80} & 0 & 20 & 0 & \underline{90} & 20 & 20 & \underline{0} & 80 & 0 & 40 & 20 & 0 & 50 \\
Gemini 3.8 Flash & 40 & \underline{90} & 40 & \underline{80} & 60 & 30 & 0 & 0 & \underline{60} & \underline{10} & \textbf{100} & 10 & \underline{40} & \underline{0} & \underline{90} & 20 & 30 & 0 & 0 & 20 \\
Claude Sonnet 5.5 & 30 & 70 & 30 & \textbf{100} & 40 & 20 & 0 & 0 & 20 & 0 & \textbf{100} & 20 & 10 & \underline{0} & 80 & 0 & 40 & 0 & \underline{10} & 50 \\
GPT-5.6 Luna & 20 & \underline{90} & 30 & 70 & 40 & 20 & 40 & 0 & 30 & 0 & \textbf{100} & 10 & 30 & \underline{0} & 70 & 0 & 50 & 10 & 0 & 50 \\
Gemini 3.6 Flash & 10 & \underline{90} & 40 & 30 & 20 & 40 & 40 & 0 & 10 & 0 & \textbf{100} & 0 & 0 & \underline{0} & 60 & 0 & 20 & 0 & 0 & 30 \\
Claude Haiku 5.5 & 30 & \textbf{100} & 0 & 60 & 50 & 20 & \underline{80} & 0 & 10 & 0 & \underline{90} & 0 & 10 & \underline{0} & 30 & 0 & 40 & 0 & 0 & 60 \\
Qwen3.8 Flash & 0 & 50 & 0 & 70 & 50 & 30 & 10 & 0 & 10 & 0 & \textbf{100} & 0 & 20 & \underline{0} & 40 & 0 & 20 & 10 & 0 & 30 \\
Qwen3.7 Plus & 20 & 50 & 20 & 0 & 20 & 20 & 20 & 0 & 30 & 0 & \textbf{100} & 0 & 0 & \underline{0} & 50 & 0 & 30 & 0 & 0 & 20 \\
Seed-2.0-Lite & 0 & 50 & 10 & 0 & 20 & 10 & 20 & 0 & 40 & 0 & \textbf{100} & 20 & 20 & \underline{0} & 10 & 20 & 20 & 0 & 0 & 10 \\
Claude Sonnet 5 & 0 & \underline{90} & 20 & 10 & 0 & 0 & 0 & 0 & 20 & 0 & 60 & 0 & 0 & \underline{0} & 10 & 0 & 0 & 0 & 0 & 10 \\
Seed 2.1 Turbo & 0 & 20 & 10 & 0 & 10 & 10 & 0 & 0 & 10 & 0 & \textbf{100} & 0 & 0 & \underline{0} & 20 & 0 & 20 & 0 & 0 & 10 \\
GLM 5V Turbo & 0 & 10 & 20 & 0 & 10 & 0 & 10 & 0 & 10 & \underline{10} & 80 & 10 & 0 & \underline{0} & 20 & 0 & 0 & 0 & 0 & 10 \\
Seed-2.0-Mini & 0 & 20 & 10 & 0 & 0 & 0 & 10 & 0 & 10 & 0 & 60 & 10 & 0 & \underline{0} & 10 & 10 & 10 & 0 & 0 & 0 \\
Gemini 3.5 Flash Lite & 0 & 40 & 0 & 0 & 0 & 30 & 0 & 0 & 10 & 0 & 60 & 0 & 0 & \underline{0} & 20 & 0 & 0 & 0 & 0 & 0 \\
\midrule
\multicolumn{21}{@{}l}{\textbf{Open-Weight Models}} \\
MiniMax M3.1 Flash & 20 & \textbf{100} & 0 & 40 & 30 & 20 & 0 & 0 & 10 & 0 & \textbf{100} & 0 & 30 & \underline{0} & 50 & 0 & 20 & 0 & 0 & 40 \\
GLM 5.3 Flash & 10 & \textbf{100} & 0 & 50 & 10 & 20 & 0 & 0 & 10 & 0 & \textbf{100} & 10 & 10 & \underline{0} & 40 & 0 & 10 & 0 & 0 & 50 \\
Qwen3.8 27B & 0 & \textbf{100} & 0 & 40 & 10 & 0 & 0 & 0 & 20 & 0 & \textbf{100} & 10 & 20 & \underline{0} & 0 & 0 & 20 & 0 & 0 & 20 \\
DeepSeek V4.1 Flash & 0 & 60 & 0 & 40 & 20 & 10 & 0 & \underline{10} & 10 & 0 & \textbf{100} & 0 & 10 & \underline{0} & 30 & 0 & 10 & 0 & 0 & 0 \\
MiniMax M3 & 0 & 40 & 0 & 0 & 0 & 10 & 0 & 0 & 0 & 0 & 70 & 0 & 0 & \underline{0} & 10 & 0 & 20 & 0 & 0 & 20 \\
Gemma 4 31B & 0 & 30 & 0 & 0 & 0 & 10 & 0 & 0 & 0 & 0 & \underline{90} & 0 & 0 & \underline{0} & 10 & 0 & 0 & 0 & 0 & 10 \\
Kimi K2.6 & 0 & 0 & 20 & 0 & 0 & 0 & 0 & 0 & 0 & 0 & \underline{90} & 0 & 0 & \underline{0} & 0 & 0 & 10 & 0 & 0 & 10 \\
Kimi K2.5 & 0 & 0 & 0 & 0 & 0 & 0 & 0 & 0 & 0 & 0 & 70 & 10 & 10 & \underline{0} & 0 & 0 & 20 & 0 & 0 & 0 \\
Llama 4 Maverick & 0 & 0 & 20 & 0 & 0 & 10 & 0 & 0 & 0 & 0 & 40 & 0 & 0 & \underline{0} & 0 & 0 & 20 & 0 & 0 & 0 \\
Qwen3.5-9B & 0 & 10 & 0 & 0 & 0 & 10 & 0 & 0 & 0 & 0 & 40 & 0 & 0 & \underline{0} & 10 & 0 & 10 & 0 & 0 & 0 \\
DeepSeek V4 Flash & 0 & 10 & 0 & 0 & 0 & 10 & 0 & 0 & 0 & 0 & 20 & 0 & 0 & \underline{0} & 10 & 0 & 20 & 0 & 0 & 0 \\
\bottomrule\end{tabular}}
\end{minipage}\par
\par\noindent\begin{minipage}{\linewidth}\centering
\pdfbookmark[2]{Apparatus Scores at k = 4}{matrix-k4}
\captionof{table}{Sequence Exact Match (\%) by Apparatus for $k=4$ ($N=200$).}\label{tab:matrix-k4}
{\fontsize{8.5}{9.3}\selectfont\setlength{\tabcolsep}{0pt}\renewcommand{\arraystretch}{1.0}
\begin{tabular}{@{}l@{\hspace{3pt}}*{20}{>{\centering\arraybackslash}p{\dimexpr(\linewidth-95pt)/20\relax}}@{}}\toprule
 & \multicolumn{4}{c}{Dynamics} & \multicolumn{4}{c}{Thermal} & \multicolumn{4}{c}{Electromagnetism} & \multicolumn{4}{c}{Acoustics} & \multicolumn{4}{c}{Optics} \\
\cmidrule(lr){2-5}\cmidrule(lr){6-9}\cmidrule(lr){10-13}\cmidrule(lr){14-17}\cmidrule(lr){18-21}
Model & \hyperref[promptpage:S01]{S01} & \hyperref[promptpage:S02]{S02} & \hyperref[promptpage:S03]{S03} & \hyperref[promptpage:S04]{S04} & \hyperref[promptpage:S05]{S05} & \hyperref[promptpage:S06]{S06} & \hyperref[promptpage:S07]{S07} & \hyperref[promptpage:S08]{S08} & \hyperref[promptpage:S09]{S09} & \hyperref[promptpage:S10]{S10} & \hyperref[promptpage:S11]{S11} & \hyperref[promptpage:S12]{S12} & \hyperref[promptpage:S13]{S13} & \hyperref[promptpage:S14]{S14} & \hyperref[promptpage:S15]{S15} & \hyperref[promptpage:S16]{S16} & \hyperref[promptpage:S17]{S17} & \hyperref[promptpage:S18]{S18} & \hyperref[promptpage:S19]{S19} & \hyperref[promptpage:S20]{S20} \\\midrule
\multicolumn{21}{@{}l}{\textbf{Proprietary Models}} \\
GPT-6.1 Sol & \textbf{60} & \textbf{100} & \textbf{80} & \underline{90} & \textbf{90} & \textbf{70} & \textbf{100} & \textbf{100} & \textbf{60} & \textbf{40} & \textbf{100} & \textbf{40} & \textbf{100} & \textbf{10} & \textbf{100} & \textbf{60} & \textbf{80} & \textbf{60} & \textbf{10} & \underline{70} \\
GPT-6.1 Terra & \underline{40} & \textbf{100} & \textbf{80} & \textbf{100} & \textbf{90} & \textbf{70} & \underline{90} & \underline{90} & \underline{50} & \underline{0} & \textbf{100} & \textbf{40} & \textbf{100} & \textbf{10} & \textbf{100} & \textbf{60} & \textbf{80} & \underline{50} & \underline{0} & \textbf{100} \\
GPT-6 Luna & 10 & \textbf{100} & 20 & \underline{90} & \underline{30} & \underline{50} & 80 & 0 & 20 & \underline{0} & \underline{90} & 0 & 10 & \underline{0} & 60 & \underline{0} & 40 & 0 & \underline{0} & 30 \\
Gemini 3.8 Flash & 10 & 80 & 20 & 60 & \underline{30} & 10 & 40 & 0 & \underline{50} & \underline{0} & \textbf{100} & 0 & 20 & \underline{0} & \underline{80} & \underline{0} & 40 & 0 & \underline{0} & 30 \\
Claude Sonnet 5.5 & 10 & 80 & 10 & \textbf{100} & 10 & 30 & 10 & 0 & 30 & \underline{0} & \textbf{100} & 0 & 20 & \underline{0} & \underline{80} & \underline{0} & \underline{50} & 0 & \underline{0} & 50 \\
GPT-5.6 Luna & 10 & \underline{90} & \underline{40} & 50 & \underline{30} & 10 & 20 & 20 & 10 & \underline{0} & 70 & 0 & 10 & \underline{0} & 60 & \underline{0} & 0 & 0 & \underline{0} & 20 \\
Gemini 3.6 Flash & 0 & 80 & 20 & 10 & 20 & 40 & 40 & 10 & 0 & \underline{0} & \textbf{100} & 0 & \underline{30} & \underline{0} & 40 & \underline{0} & 30 & 0 & \underline{0} & 20 \\
Claude Haiku 5.5 & 0 & 80 & 0 & 60 & \underline{30} & 10 & 60 & 0 & 0 & \underline{0} & 80 & 0 & 10 & \underline{0} & 0 & \underline{0} & 20 & 0 & \underline{0} & 50 \\
Qwen3.8 Flash & 0 & 70 & 0 & 40 & \underline{30} & 20 & 20 & 0 & 0 & \underline{0} & \textbf{100} & 0 & 0 & \underline{0} & 30 & \underline{0} & 10 & 0 & \underline{0} & 60 \\
Qwen3.7 Plus & 0 & 40 & 10 & 0 & 10 & 30 & 10 & 0 & 0 & \underline{0} & \textbf{100} & 0 & 10 & \underline{0} & 10 & \underline{0} & 0 & 0 & \underline{0} & 10 \\
Seed-2.0-Lite & 0 & 10 & 20 & 0 & 0 & 20 & 0 & 0 & 40 & \underline{0} & \textbf{100} & 0 & 0 & \underline{0} & 0 & \underline{0} & 10 & 0 & \underline{0} & 10 \\
Claude Sonnet 5 & 0 & \underline{90} & 10 & 10 & 0 & 0 & 0 & 0 & 0 & \underline{0} & 20 & 0 & 0 & \underline{0} & 0 & \underline{0} & 0 & 0 & \underline{0} & 0 \\
Seed 2.1 Turbo & 0 & 20 & 0 & 0 & 0 & 10 & 0 & 0 & 0 & \underline{0} & \underline{90} & 0 & 0 & \underline{0} & 0 & \underline{0} & 10 & 0 & \underline{0} & 0 \\
GLM 5V Turbo & 0 & 30 & 10 & 0 & 0 & 10 & 0 & 0 & 0 & \underline{0} & 70 & 0 & 0 & \underline{0} & 0 & \underline{0} & 0 & 0 & \underline{0} & 0 \\
Seed-2.0-Mini & 0 & 10 & 10 & 0 & 0 & 0 & 0 & 0 & 0 & \underline{0} & 40 & 0 & 0 & \underline{0} & 10 & \underline{0} & 0 & 0 & \underline{0} & 0 \\
Gemini 3.5 Flash Lite & 0 & 20 & 0 & 0 & 0 & 0 & 0 & 0 & 0 & \underline{0} & 30 & 0 & 0 & \underline{0} & 10 & \underline{0} & 0 & 0 & \underline{0} & 0 \\
\midrule
\multicolumn{21}{@{}l}{\textbf{Open-Weight Models}} \\
MiniMax M3.1 Flash & 20 & 50 & 0 & 30 & \underline{30} & 0 & 0 & 0 & 0 & \underline{0} & \textbf{100} & 0 & 20 & \underline{0} & 10 & \underline{0} & 10 & 0 & \underline{0} & 0 \\
GLM 5.3 Flash & 0 & 30 & 0 & 40 & 0 & 10 & 0 & 0 & 10 & \underline{0} & \underline{90} & 0 & 0 & \underline{0} & 20 & \underline{0} & 40 & 0 & \underline{0} & 20 \\
Qwen3.8 27B & 0 & 60 & 0 & 50 & 0 & 10 & 0 & 0 & 0 & \underline{0} & \textbf{100} & 0 & 0 & \underline{0} & 0 & \underline{0} & 10 & 0 & \underline{0} & 10 \\
DeepSeek V4.1 Flash & 20 & 40 & 0 & 10 & 20 & 10 & 0 & 0 & 0 & \underline{0} & \underline{90} & \underline{10} & 10 & \underline{0} & 0 & \underline{0} & 10 & 0 & \textbf{10} & 10 \\
MiniMax M3 & 0 & 30 & 0 & 0 & 0 & 10 & 0 & 10 & 0 & \underline{0} & 50 & 0 & 10 & \underline{0} & 0 & \underline{0} & 0 & 0 & \underline{0} & 10 \\
Gemma 4 31B & 0 & 40 & 0 & 0 & 0 & 10 & 0 & 0 & 0 & \underline{0} & 80 & 0 & 0 & \underline{0} & 0 & \underline{0} & 0 & 0 & \underline{0} & 10 \\
Kimi K2.6 & 0 & 0 & 0 & 0 & 0 & 0 & 0 & 0 & 0 & \underline{0} & 70 & 0 & 0 & \underline{0} & 0 & \underline{0} & 10 & 0 & \underline{0} & 0 \\
Kimi K2.5 & 0 & 0 & 0 & 0 & 0 & 10 & 0 & 0 & 0 & \underline{0} & 30 & 0 & 10 & \underline{0} & 0 & \underline{0} & 10 & 0 & \underline{0} & 10 \\
Llama 4 Maverick & 0 & 10 & 10 & 0 & 0 & 20 & 0 & 0 & 0 & \underline{0} & 10 & 0 & 0 & \underline{0} & 10 & \underline{0} & 0 & 0 & \underline{0} & 0 \\
Qwen3.5-9B & 0 & 0 & 0 & 0 & 0 & 0 & 0 & 0 & 0 & \underline{0} & 20 & 0 & 0 & \underline{0} & 0 & \underline{0} & 0 & 0 & \underline{0} & 0 \\
DeepSeek V4 Flash & 0 & 0 & 0 & 0 & 0 & 20 & 0 & 0 & 0 & \underline{0} & 10 & 0 & 0 & \underline{0} & 10 & \underline{0} & 0 & 0 & \underline{0} & 0 \\
\bottomrule\end{tabular}}
\end{minipage}\par
\par\noindent\begin{minipage}{\linewidth}\centering
\pdfbookmark[2]{Apparatus Scores at k = 5}{matrix-k5}
\captionof{table}{Sequence Exact Match (\%) by Apparatus for $k=5$ ($N=200$).}\label{tab:matrix-k5}
{\fontsize{8.5}{9.3}\selectfont\setlength{\tabcolsep}{0pt}\renewcommand{\arraystretch}{1.0}
\begin{tabular}{@{}l@{\hspace{3pt}}*{20}{>{\centering\arraybackslash}p{\dimexpr(\linewidth-95pt)/20\relax}}@{}}\toprule
 & \multicolumn{4}{c}{Dynamics} & \multicolumn{4}{c}{Thermal} & \multicolumn{4}{c}{Electromagnetism} & \multicolumn{4}{c}{Acoustics} & \multicolumn{4}{c}{Optics} \\
\cmidrule(lr){2-5}\cmidrule(lr){6-9}\cmidrule(lr){10-13}\cmidrule(lr){14-17}\cmidrule(lr){18-21}
Model & \hyperref[promptpage:S01]{S01} & \hyperref[promptpage:S02]{S02} & \hyperref[promptpage:S03]{S03} & \hyperref[promptpage:S04]{S04} & \hyperref[promptpage:S05]{S05} & \hyperref[promptpage:S06]{S06} & \hyperref[promptpage:S07]{S07} & \hyperref[promptpage:S08]{S08} & \hyperref[promptpage:S09]{S09} & \hyperref[promptpage:S10]{S10} & \hyperref[promptpage:S11]{S11} & \hyperref[promptpage:S12]{S12} & \hyperref[promptpage:S13]{S13} & \hyperref[promptpage:S14]{S14} & \hyperref[promptpage:S15]{S15} & \hyperref[promptpage:S16]{S16} & \hyperref[promptpage:S17]{S17} & \hyperref[promptpage:S18]{S18} & \hyperref[promptpage:S19]{S19} & \hyperref[promptpage:S20]{S20} \\\midrule
\multicolumn{21}{@{}l}{\textbf{Proprietary Models}} \\
GPT-6.1 Sol & \textbf{60} & \textbf{100} & \textbf{80} & \underline{80} & \textbf{100} & \textbf{70} & \textbf{100} & \textbf{90} & \textbf{40} & \textbf{20} & \textbf{100} & \textbf{50} & \textbf{100} & \textbf{10} & \textbf{100} & \underline{90} & \textbf{80} & \textbf{50} & \textbf{10} & \underline{80} \\
GPT-6.1 Terra & \textbf{60} & \textbf{100} & 40 & \textbf{100} & \textbf{100} & \textbf{70} & \textbf{100} & \underline{70} & \underline{30} & \underline{10} & \textbf{100} & \underline{20} & \textbf{100} & \underline{0} & \textbf{100} & \textbf{100} & \textbf{80} & \textbf{50} & \underline{0} & \textbf{100} \\
GPT-6 Luna & \underline{30} & \textbf{100} & 10 & \underline{80} & 10 & \underline{30} & 60 & 0 & 10 & 0 & \underline{90} & \underline{20} & \underline{10} & \underline{0} & 60 & 0 & 20 & \underline{0} & \underline{0} & 10 \\
Gemini 3.8 Flash & 0 & \underline{90} & 40 & 60 & \underline{20} & 20 & 0 & 0 & 20 & 0 & \textbf{100} & 0 & \underline{10} & \underline{0} & \underline{90} & 10 & \underline{30} & \underline{0} & \underline{0} & 20 \\
Claude Sonnet 5.5 & 20 & 70 & 0 & \textbf{100} & 0 & 10 & 10 & 0 & 20 & 0 & \textbf{100} & 0 & 0 & \underline{0} & 70 & 10 & 20 & \underline{0} & \underline{0} & 60 \\
GPT-5.6 Luna & 10 & 80 & 20 & 70 & \underline{20} & 10 & 10 & 0 & 10 & 0 & \underline{90} & 0 & 0 & \underline{0} & 40 & 0 & 20 & \underline{0} & \underline{0} & 10 \\
Gemini 3.6 Flash & 0 & 60 & \underline{60} & 10 & 10 & 20 & \underline{70} & 0 & 0 & 0 & \textbf{100} & 0 & 0 & \underline{0} & 40 & 0 & 0 & \underline{0} & \underline{0} & 40 \\
Claude Haiku 5.5 & 0 & 50 & 0 & 0 & 0 & 0 & \textbf{100} & 0 & 0 & 0 & \underline{90} & 0 & 0 & \underline{0} & 0 & 0 & 0 & \underline{0} & \underline{0} & 30 \\
Qwen3.8 Flash & 10 & 10 & 10 & 60 & 10 & 0 & 0 & 0 & 0 & 0 & \textbf{100} & 0 & 0 & \underline{0} & 60 & 0 & 20 & \underline{0} & \underline{0} & 30 \\
Qwen3.7 Plus & 10 & 30 & 0 & 10 & 0 & 10 & 10 & 0 & 0 & 0 & \textbf{100} & 0 & 0 & \underline{0} & 20 & 0 & 0 & \underline{0} & \underline{0} & 30 \\
Seed-2.0-Lite & 0 & 0 & 10 & 0 & 0 & 10 & 0 & 0 & 0 & 0 & \textbf{100} & 0 & 0 & \underline{0} & 10 & 0 & 10 & \underline{0} & \underline{0} & 0 \\
Claude Sonnet 5 & 0 & 50 & 10 & 0 & 0 & 0 & 0 & 0 & 0 & 0 & 20 & 0 & 0 & \underline{0} & 0 & 0 & 0 & \underline{0} & \underline{0} & 0 \\
Seed 2.1 Turbo & 0 & 0 & 0 & 0 & 0 & 0 & 0 & 0 & 0 & 0 & 80 & 0 & 0 & \underline{0} & 0 & 0 & 0 & \underline{0} & \underline{0} & 0 \\
GLM 5V Turbo & 0 & 0 & 20 & 0 & 0 & 0 & 0 & 0 & 10 & 0 & 60 & 0 & 0 & \underline{0} & 0 & 0 & 10 & \underline{0} & \underline{0} & 0 \\
Seed-2.0-Mini & 0 & 0 & 10 & 0 & 0 & 0 & 0 & 0 & 0 & 0 & 60 & 0 & 0 & \underline{0} & 0 & 0 & 20 & \underline{0} & \underline{0} & 0 \\
Gemini 3.5 Flash Lite & 0 & 20 & 0 & 0 & 0 & 0 & 0 & 0 & 0 & 0 & 20 & 0 & 0 & \underline{0} & 10 & 0 & 0 & \underline{0} & \underline{0} & 0 \\
\midrule
\multicolumn{21}{@{}l}{\textbf{Open-Weight Models}} \\
MiniMax M3.1 Flash & 0 & 70 & 0 & 20 & 10 & 10 & 0 & 0 & 0 & 0 & \textbf{100} & 0 & \underline{10} & \underline{0} & 40 & 0 & 10 & \underline{0} & \underline{0} & 0 \\
GLM 5.3 Flash & 0 & 50 & 0 & 20 & 0 & \underline{30} & 0 & 0 & 0 & 0 & 60 & 0 & 0 & \underline{0} & 10 & 0 & 20 & \underline{0} & \underline{0} & 0 \\
Qwen3.8 27B & 0 & 60 & 0 & 40 & 0 & \underline{30} & 0 & 0 & 0 & 0 & \underline{90} & 0 & 0 & \underline{0} & 20 & 0 & 0 & \underline{0} & \underline{0} & 0 \\
DeepSeek V4.1 Flash & 0 & 60 & 0 & 10 & 10 & 0 & 0 & 0 & 0 & 0 & \textbf{100} & 0 & 0 & \underline{0} & 20 & 0 & 10 & \underline{0} & \underline{0} & 20 \\
MiniMax M3 & 0 & 0 & 0 & 0 & 10 & 0 & 0 & 0 & 0 & 0 & 30 & 0 & 0 & \underline{0} & 0 & 0 & 0 & \underline{0} & \underline{0} & 0 \\
Gemma 4 31B & 0 & 40 & 0 & 0 & 0 & 0 & 0 & 0 & 0 & 0 & 70 & 0 & 0 & \underline{0} & 20 & 0 & 0 & \underline{0} & \underline{0} & 0 \\
Kimi K2.6 & 0 & 0 & 0 & 0 & 0 & 0 & 0 & 0 & 0 & 0 & 70 & 0 & 0 & \underline{0} & 20 & 0 & 0 & \underline{0} & \underline{0} & 0 \\
Kimi K2.5 & 0 & 10 & 10 & 0 & 0 & 0 & 0 & 0 & 0 & 0 & 40 & 0 & 0 & \underline{0} & 0 & 0 & 0 & \underline{0} & \underline{0} & 0 \\
Llama 4 Maverick & 0 & 0 & 10 & 0 & 0 & 0 & 0 & 0 & 0 & 0 & 0 & 0 & 0 & \underline{0} & 0 & 0 & 0 & \underline{0} & \underline{0} & 0 \\
Qwen3.5-9B & 0 & 0 & 0 & 0 & 0 & 0 & 0 & 0 & 0 & 0 & 20 & 0 & 0 & \underline{0} & 0 & 0 & 0 & \underline{0} & \underline{0} & 0 \\
DeepSeek V4 Flash & 0 & 0 & 10 & 0 & 0 & 10 & 0 & 0 & 0 & 0 & 10 & 0 & 0 & \underline{0} & 0 & 0 & 0 & \underline{0} & \underline{0} & 0 \\
\bottomrule\end{tabular}}
\end{minipage}\par
\par\noindent\begin{minipage}{\linewidth}\centering
\pdfbookmark[2]{Apparatus Scores at k = 6}{matrix-k6}
\captionof{table}{Sequence Exact Match (\%) by Apparatus for $k=6$ ($N=200$).}\label{tab:matrix-k6}
{\fontsize{8.5}{9.3}\selectfont\setlength{\tabcolsep}{0pt}\renewcommand{\arraystretch}{1.0}
\begin{tabular}{@{}l@{\hspace{3pt}}*{20}{>{\centering\arraybackslash}p{\dimexpr(\linewidth-95pt)/20\relax}}@{}}\toprule
 & \multicolumn{4}{c}{Dynamics} & \multicolumn{4}{c}{Thermal} & \multicolumn{4}{c}{Electromagnetism} & \multicolumn{4}{c}{Acoustics} & \multicolumn{4}{c}{Optics} \\
\cmidrule(lr){2-5}\cmidrule(lr){6-9}\cmidrule(lr){10-13}\cmidrule(lr){14-17}\cmidrule(lr){18-21}
Model & \hyperref[promptpage:S01]{S01} & \hyperref[promptpage:S02]{S02} & \hyperref[promptpage:S03]{S03} & \hyperref[promptpage:S04]{S04} & \hyperref[promptpage:S05]{S05} & \hyperref[promptpage:S06]{S06} & \hyperref[promptpage:S07]{S07} & \hyperref[promptpage:S08]{S08} & \hyperref[promptpage:S09]{S09} & \hyperref[promptpage:S10]{S10} & \hyperref[promptpage:S11]{S11} & \hyperref[promptpage:S12]{S12} & \hyperref[promptpage:S13]{S13} & \hyperref[promptpage:S14]{S14} & \hyperref[promptpage:S15]{S15} & \hyperref[promptpage:S16]{S16} & \hyperref[promptpage:S17]{S17} & \hyperref[promptpage:S18]{S18} & \hyperref[promptpage:S19]{S19} & \hyperref[promptpage:S20]{S20} \\\midrule
\multicolumn{21}{@{}l}{\textbf{Proprietary Models}} \\
GPT-6.1 Sol & \underline{30} & \textbf{100} & \textbf{100} & 50 & \textbf{80} & \underline{60} & \textbf{100} & \underline{80} & \textbf{40} & \textbf{30} & \textbf{100} & \textbf{60} & \textbf{90} & \underline{0} & \underline{90} & \textbf{100} & \textbf{80} & \textbf{80} & \underline{0} & \underline{90} \\
GPT-6.1 Terra & \textbf{50} & \textbf{100} & \underline{50} & \textbf{100} & \textbf{80} & \textbf{70} & \textbf{100} & \textbf{90} & \underline{30} & \underline{0} & \textbf{100} & \underline{20} & \textbf{90} & \textbf{10} & \textbf{100} & \textbf{100} & \textbf{80} & \underline{30} & \underline{0} & \textbf{100} \\
GPT-6 Luna & \underline{30} & \underline{90} & 20 & \underline{70} & 0 & 30 & \underline{90} & 0 & 0 & \underline{0} & \underline{90} & 0 & \underline{20} & \underline{0} & \underline{90} & \underline{0} & 10 & 0 & \textbf{10} & 20 \\
Gemini 3.8 Flash & 10 & \underline{90} & 0 & 60 & \underline{20} & 0 & 10 & 0 & \underline{30} & \underline{0} & \textbf{100} & 0 & 0 & \underline{0} & 60 & \underline{0} & 10 & 0 & \underline{0} & 40 \\
Claude Sonnet 5.5 & 0 & 40 & 0 & 30 & 0 & 20 & 0 & 0 & 10 & \underline{0} & \textbf{100} & 10 & 10 & \underline{0} & 60 & \underline{0} & \underline{30} & 0 & \underline{0} & 20 \\
GPT-5.6 Luna & 10 & 70 & 40 & 50 & 10 & 10 & 20 & 0 & 0 & \underline{0} & 80 & 10 & 10 & \underline{0} & 60 & \underline{0} & 10 & 0 & \underline{0} & 10 \\
Gemini 3.6 Flash & 10 & 50 & 20 & 20 & 0 & 10 & 20 & 0 & 0 & \underline{0} & \textbf{100} & 0 & 0 & \underline{0} & 70 & \underline{0} & 10 & 0 & \underline{0} & 30 \\
Claude Haiku 5.5 & 0 & 80 & 0 & 0 & 0 & 10 & 60 & 0 & 0 & \underline{0} & \underline{90} & 0 & 0 & \underline{0} & 0 & \underline{0} & 0 & 0 & \underline{0} & 30 \\
Qwen3.8 Flash & 10 & 10 & 0 & \underline{70} & 0 & 10 & 20 & 0 & 0 & \underline{0} & \textbf{100} & 0 & \underline{20} & \underline{0} & 20 & \underline{0} & 0 & 0 & \underline{0} & 20 \\
Qwen3.7 Plus & 0 & 30 & 10 & 0 & 0 & 20 & 0 & 0 & 10 & \underline{0} & \textbf{100} & 0 & 0 & \underline{0} & 20 & \underline{0} & 0 & 0 & \underline{0} & 40 \\
Seed-2.0-Lite & 0 & 0 & 0 & 0 & 0 & 0 & 0 & 0 & 10 & \underline{0} & \textbf{100} & 0 & 0 & \underline{0} & 0 & \underline{0} & 0 & 0 & \underline{0} & 0 \\
Claude Sonnet 5 & 0 & 30 & 10 & 0 & 0 & 0 & 0 & 0 & 0 & \underline{0} & 20 & 0 & 0 & \underline{0} & 0 & \underline{0} & 0 & 0 & \underline{0} & 0 \\
Seed 2.1 Turbo & 0 & 0 & 0 & 0 & 0 & 0 & 0 & 0 & 0 & \underline{0} & 70 & 0 & 0 & \underline{0} & 0 & \underline{0} & 0 & 0 & \underline{0} & 10 \\
GLM 5V Turbo & 0 & 10 & 10 & 0 & 0 & 0 & 0 & 0 & 0 & \underline{0} & 30 & 0 & 0 & \underline{0} & 0 & \underline{0} & 0 & 0 & \underline{0} & 0 \\
Seed-2.0-Mini & 0 & 0 & 0 & 0 & 0 & 0 & 0 & 0 & 0 & \underline{0} & 60 & 0 & 0 & \underline{0} & 0 & \underline{0} & 0 & 0 & \underline{0} & 0 \\
Gemini 3.5 Flash Lite & 0 & 10 & 0 & 0 & 0 & 0 & 10 & 0 & 0 & \underline{0} & 0 & 0 & 0 & \underline{0} & 0 & \underline{0} & 0 & 0 & \underline{0} & 0 \\
\midrule
\multicolumn{21}{@{}l}{\textbf{Open-Weight Models}} \\
MiniMax M3.1 Flash & 0 & 60 & 0 & 0 & 0 & 0 & 0 & 0 & 10 & \underline{0} & \textbf{100} & 0 & \underline{20} & \underline{0} & 30 & \underline{0} & 10 & 0 & \underline{0} & 20 \\
GLM 5.3 Flash & 0 & 40 & 0 & 20 & 0 & 0 & 0 & 0 & 0 & \underline{0} & 70 & 0 & 0 & \underline{0} & 20 & \underline{0} & 0 & 0 & \underline{0} & 30 \\
Qwen3.8 27B & 0 & 30 & 0 & 40 & 0 & 0 & 10 & 0 & 0 & \underline{0} & \textbf{100} & 0 & 0 & \underline{0} & 0 & \underline{0} & 0 & 0 & \underline{0} & 30 \\
DeepSeek V4.1 Flash & 0 & 20 & 0 & 0 & 10 & 10 & 0 & 0 & 0 & \underline{0} & 80 & 0 & 0 & \underline{0} & 0 & \underline{0} & 0 & 0 & \underline{0} & 10 \\
MiniMax M3 & 10 & 0 & 0 & 0 & 0 & 0 & 0 & 0 & 10 & \underline{0} & 70 & 0 & 0 & \underline{0} & 10 & \underline{0} & 0 & 0 & \underline{0} & 0 \\
Gemma 4 31B & 0 & 20 & 0 & 0 & 0 & 0 & 0 & 0 & 0 & \underline{0} & 60 & 0 & 0 & \underline{0} & 10 & \underline{0} & 0 & 0 & \underline{0} & 0 \\
Kimi K2.6 & 0 & 0 & 0 & 0 & 0 & 0 & 0 & 0 & 0 & \underline{0} & 60 & 0 & 0 & \underline{0} & 0 & \underline{0} & 10 & 0 & \underline{0} & 0 \\
Kimi K2.5 & 0 & 0 & 0 & 0 & 0 & 0 & 0 & 0 & 0 & \underline{0} & 40 & 0 & 0 & \underline{0} & 0 & \underline{0} & 0 & 0 & \underline{0} & 0 \\
Llama 4 Maverick & 0 & 0 & 10 & 0 & 0 & 0 & 0 & 0 & 0 & \underline{0} & 0 & 0 & 0 & \underline{0} & 0 & \underline{0} & 0 & 0 & \underline{0} & 0 \\
Qwen3.5-9B & 0 & 0 & 0 & 0 & 0 & 0 & 0 & 0 & 0 & \underline{0} & 10 & 0 & 0 & \underline{0} & 0 & \underline{0} & 0 & 0 & \underline{0} & 0 \\
DeepSeek V4 Flash & 0 & 0 & 0 & 0 & 0 & 0 & 0 & 0 & 0 & \underline{0} & 10 & 0 & 0 & \underline{0} & 0 & \underline{0} & 0 & 0 & \underline{0} & 0 \\
\bottomrule\end{tabular}}
\end{minipage}\par

\subsection{Preference Training across Seeds}\label{sec:preference-seeds}
Table~\ref{tab:preference-seeds} separates the runs that contribute to the two-seed means in the main paper. Within each method, the runs share the supervised starting point, preference data and test questions. The two SimPO runs each score 57.2\% Overall after rounding, although their four-event scores are 57.0\% and 55.0\%. KTO shows a larger spread at that length, with scores of 35.0\% and 41.5\%, a difference of 13 complete answers across the same 200 questions. Its six-event scores agree at 14.5\%. The variation between seeds is thus concentrated at particular lengths and can disappear in a pooled score.
\par\noindent\begin{minipage}{\linewidth}\centering
\captionof{table}{Qwen3.5-9B Sequence Exact Match (\%) across Seeds for SimPO~\cite{simpo} and KTO~\cite{kto}.}\label{tab:preference-seeds}
{\small\setlength{\tabcolsep}{3pt}
\begin{tabular*}{\linewidth}{@{}l@{\extracolsep{\fill}}rrrrrrrr@{}}\toprule
Method & Training seed & $k=1$ & $k=2$ & $k=3$ & $k=4$ & $k=5$ & $k=6$ & Overall \\\midrule
SFT + SimPO & 20261108 & 87.4 & 72.6 & 65.0 & 57.0 & 32.0 & 31.5 & 57.2 \\
SFT + SimPO & 20261109 & 86.3 & 72.6 & 66.5 & 55.0 & 33.0 & 32.0 & 57.2 \\
\midrule
SFT + KTO & 20261107 & 85.3 & 75.8 & 67.0 & 35.0 & 12.5 & 14.5 & 47.8 \\
SFT + KTO & 20261108 & 87.4 & 73.7 & 69.0 & 41.5 & 19.0 & 14.5 & 50.3 \\
\bottomrule\end{tabular*}}
\end{minipage}\par

\subsection{Complete Results by Reasoning Effort}\label{sec:reasoning-effort}
Table~\ref{tab:reasoning-effort} follows the same 200 six-event questions from the initial generation through the final answer at each effort setting. Exact match counts complete sequences and event accuracy counts the 1,200 individual control sets. Mean output and reasoning tokens refer to the initial generation. Output tokens include reasoning tokens.

For Gemini, moving from Medium to High raises mean reasoning use from 5,279.5 to 5,773.1 tokens, while correctly recovered final events fall from 608 to 467 of 1,200. The High setting yields 12 correct initial sequences and 16 correct final sequences. Of the 467 event sets correctly recovered in the final answers, 371 belong to the 184 sequences that still contain an error.
\par\noindent\begin{minipage}{\linewidth}\centering
\captionof{table}{Accuracy and Mean Token Use by Reasoning Effort.}\label{tab:reasoning-effort}
{\small\setlength{\tabcolsep}{3pt}
\begin{tabular*}{\linewidth}{@{}l@{\extracolsep{\fill}}rrrrrr@{}}\toprule
Model & Effort & \shortstack{Initial\\exact (\%)} & \shortstack{Final\\exact (\%)} & \shortstack{Final event\\accuracy (\%)} & \shortstack{Output\\tokens} & \shortstack{Reasoning\\tokens} \\\midrule
GPT-6.1 Sol & Low & 59.5 & 59.5 & 82.33 & 520.4 & 471.9 \\
GPT-6.1 Sol & Medium & 67.0 & 67.0 & 85.92 & 1,266.1 & 1,217.6 \\
GPT-6.1 Sol & High & 73.0 & 74.0 & 88.42 & 3,330.1 & 3,287.4 \\
\midrule
Gemini 3.8 Flash & Low & 14.0 & 15.5 & 45.58 & 1,841.2 & 1,785.9 \\
Gemini 3.8 Flash & Medium & 20.5 & 21.5 & 50.67 & 5,345.2 & 5,279.5 \\
Gemini 3.8 Flash & High & 6.0 & 8.0 & 38.92 & 5,869.8 & 5,773.1 \\
\bottomrule\end{tabular*}}
\end{minipage}\par

%% file: error_analysis.tex
\subsection{Error Analysis}\label{app:error-analysis}
We examined 100 incorrect six-event responses from GPT-6.1 Sol, Gemini 3.8 Flash, and GLM 5.3 Flash. For each response, we compared the model's account of the affected events with the images and recorded controls, then assigned the response one primary error category. Visual identification accounted for 38 responses, physical inference for 24, evidence integration for 31, and action composition for seven.

Across these responses, the model recovered the complete control set at 238 of the 600 event positions. In 120 of the remaining 362 positions, two controls acted together and the model named exactly one of them. Sixteen of these omissions occurred in responses classified under action composition. The other 104 occurred in responses classified under visual identification, physical inference, or evidence integration. A missing control is visible in the submitted sequence. To determine where the model lost it, we can trace a particular event from the change it described in the images to the controls it ultimately chose.

\begin{figure}[H]
\centering
\includegraphics[width=\linewidth]{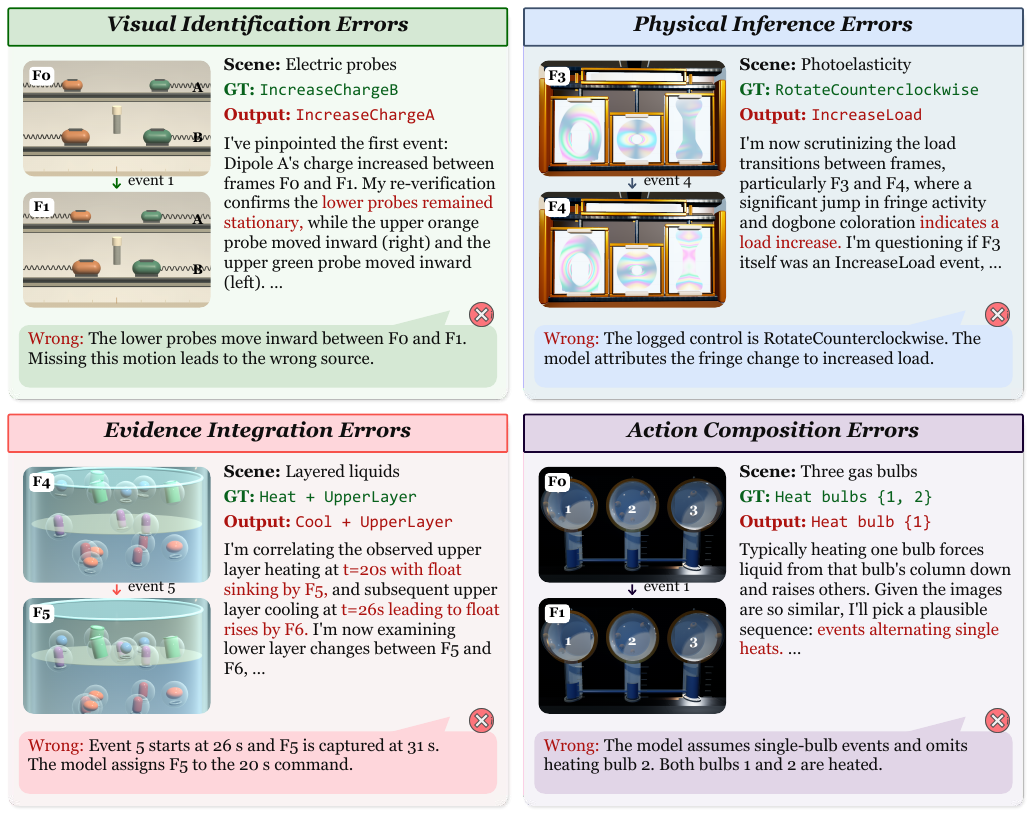}
\caption{\textbf{Four Annotated Errors.} Each panel pairs consecutive observations with the recorded controls, the model's prediction, and an excerpt from its response.}\label{fig:error-analysis}
\end{figure}

Figure~\ref{fig:error-analysis} illustrates how the four errors arise. In the electric-probe apparatus (S10), Gemini describes the lower probes as stationary. Both move inward, and it assigns the increase in charge to dipole A instead of dipole B. In the photoelastic apparatus (S17), Gemini sees the change in the fringes but attributes it to an increase in load. The recorded intervention was a counterclockwise turn of the polarizer. In the layered vessel (S06), Gemini places upper-layer heating and cooling at events four and five. The two actions did occur in that order, at events five and six. In the three-bulb apparatus (S07), GLM explicitly considers heating one bulb at a time. Its first answer identifies the left bulb and omits the middle bulb, which was heated with it. It also submits a single control at both later events involving paired heaters.

%% file: case-prompts.tex
\begingroup
\setlength{\intextsep}{2.5pt}
\setlength{\parskip}{0pt}
\captionsetup[figure]{skip=2pt}
\begin{figure}[H]
\centering
\pdfbookmark[2]{Prompt S01 Cabin Inertia}{prompt-S01}
\includegraphics[width=0.9084\linewidth]{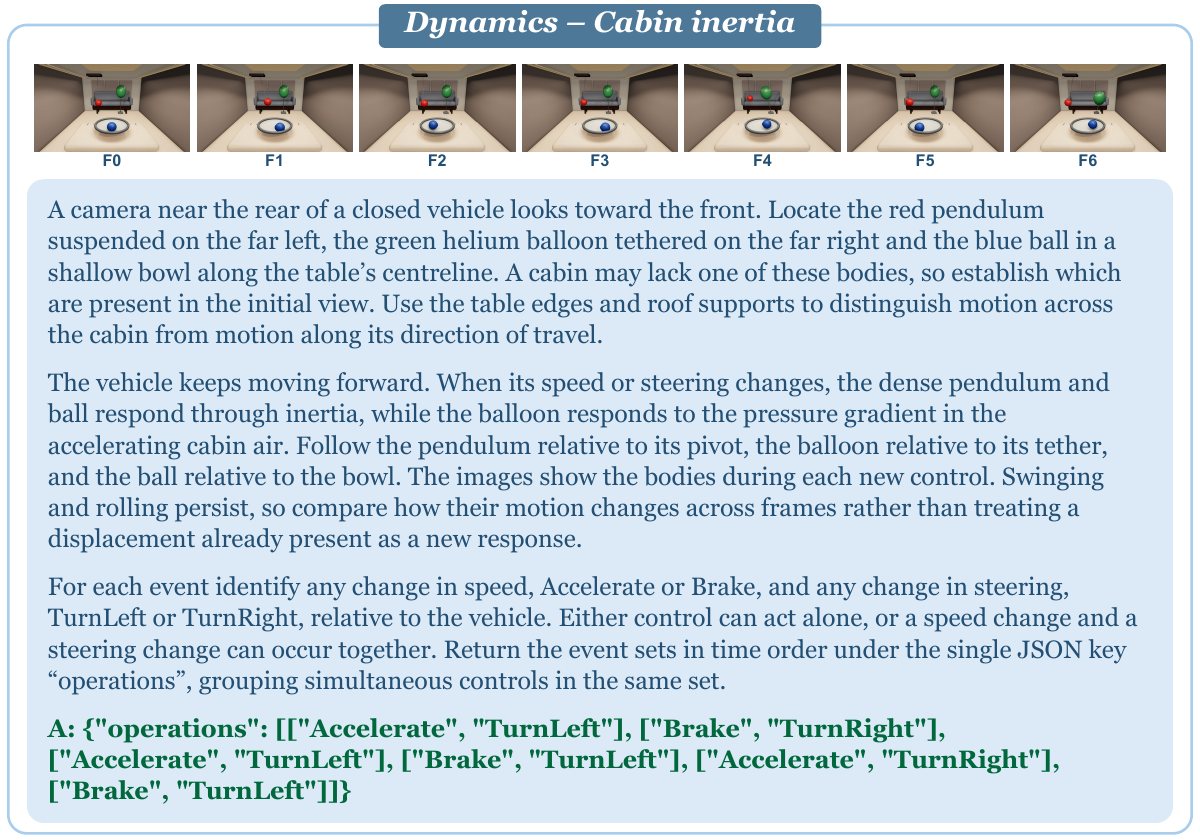}
\caption{\textbf{Example prompt and reference answer for inferring vehicle motion from inside the cabin.}}\label{promptpage:S01}
\end{figure}
\begin{figure}[H]
\centering
\pdfbookmark[2]{Prompt S02 Rotation \& Gas}{prompt-S02}
\includegraphics[width=0.9084\linewidth]{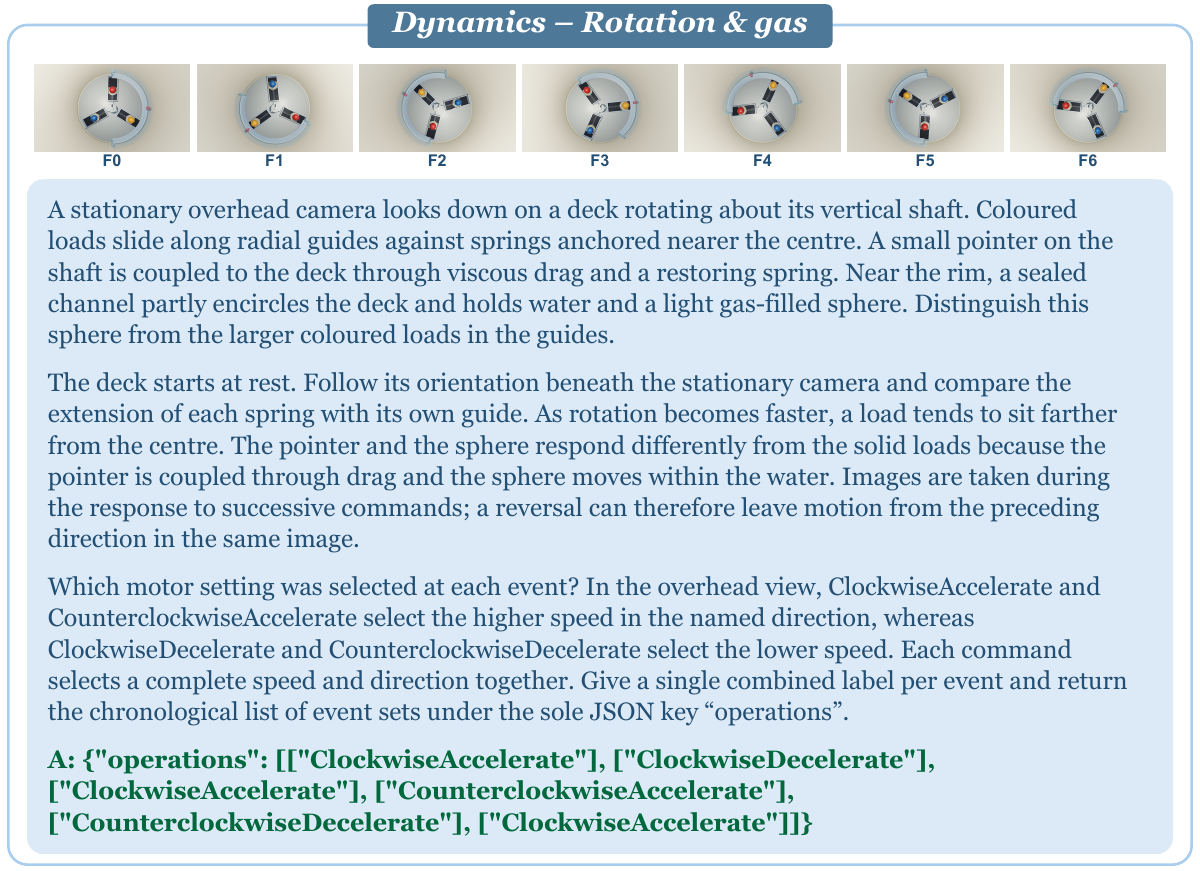}
\caption{\textbf{Example prompt and reference answer for identifying changes in motor speed and direction.}}\label{promptpage:S02}
\end{figure}
\begin{figure}[H]
\centering
\pdfbookmark[2]{Prompt S03 Driven Pendulums}{prompt-S03}
\includegraphics[width=0.9253\linewidth]{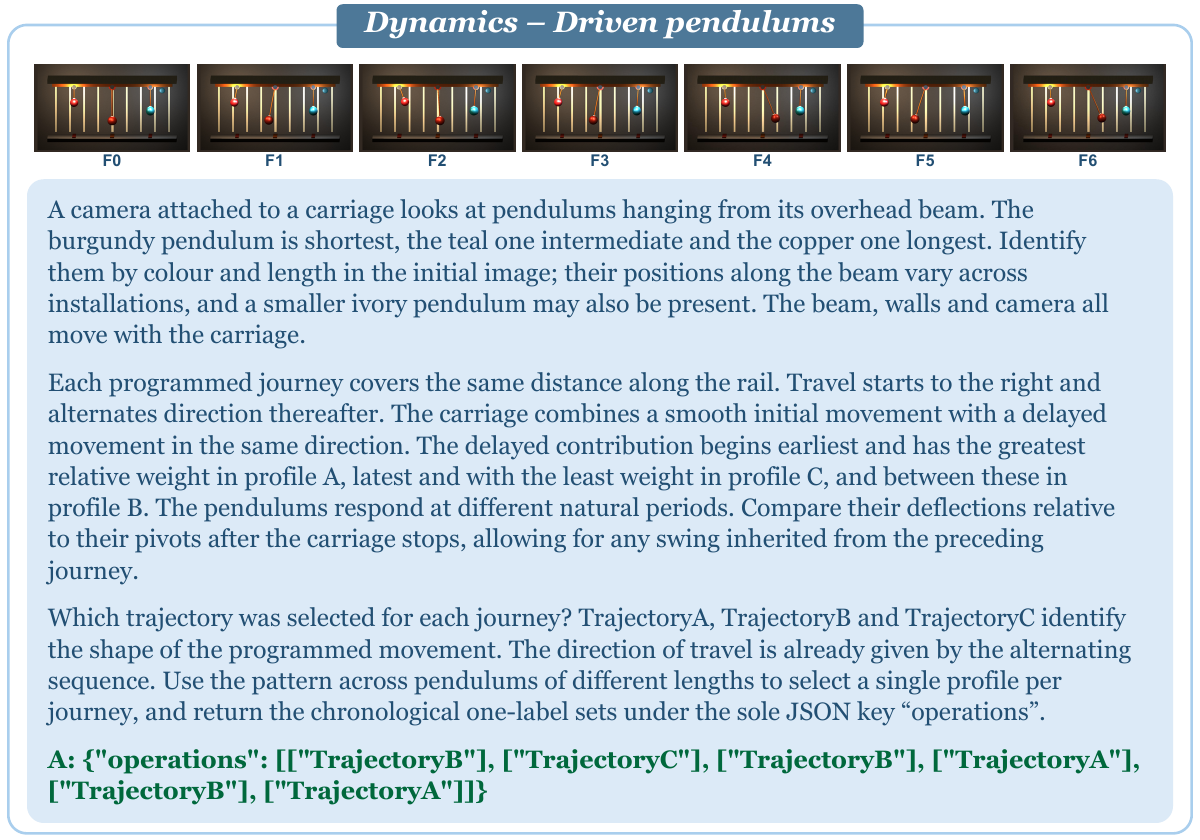}
\caption{\textbf{Example prompt and reference answer for identifying carriage paths from pendulum motion.}}\label{promptpage:S03}
\end{figure}
\begin{figure}[H]
\centering
\pdfbookmark[2]{Prompt S04 Tray Impact}{prompt-S04}
\includegraphics[width=0.9253\linewidth]{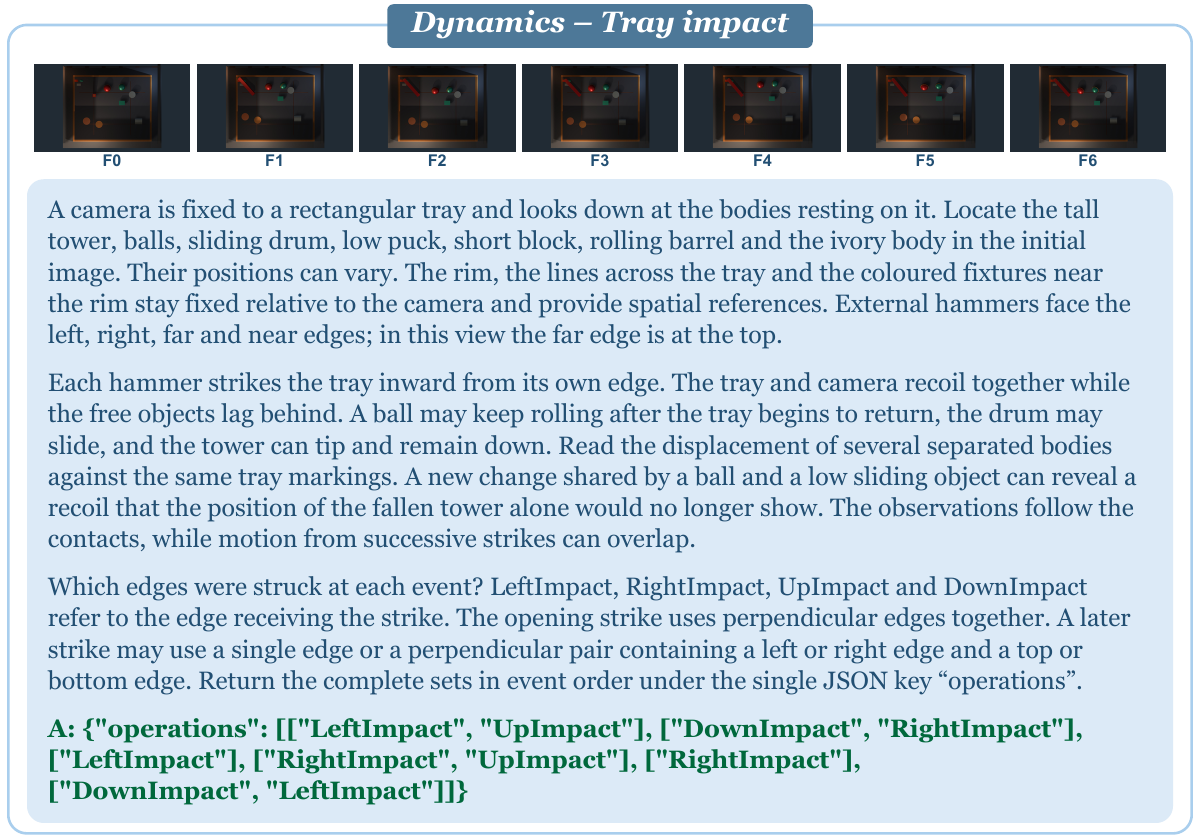}
\caption{\textbf{Example prompt and reference answer for locating impacts on the edges of a tray.}}\label{promptpage:S04}
\end{figure}
\begin{figure}[H]
\centering
\pdfbookmark[2]{Prompt S05 Moving Heat Source}{prompt-S05}
\includegraphics[width=0.9253\linewidth]{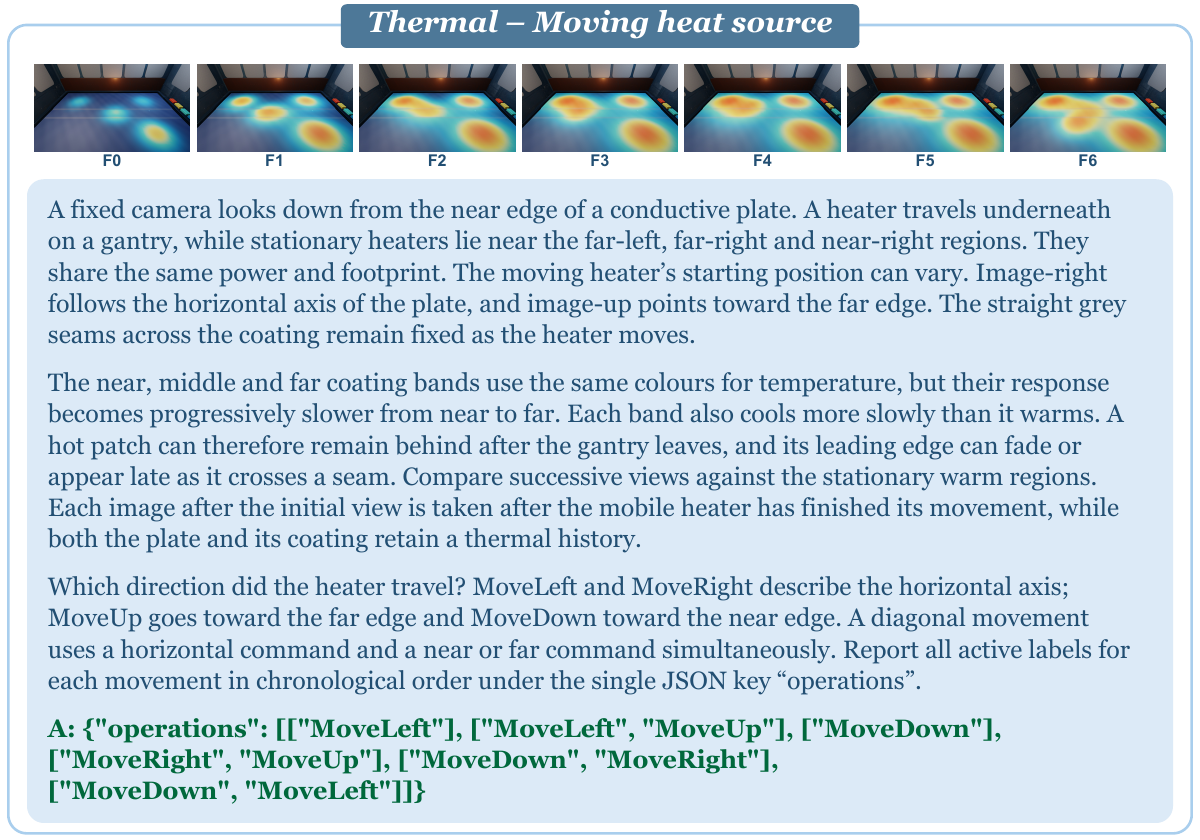}
\caption{\textbf{Example prompt and reference answer for tracking a moving heat source across a plate.}}\label{promptpage:S05}
\end{figure}
\begin{figure}[H]
\centering
\pdfbookmark[2]{Prompt S06 Layered Liquids}{prompt-S06}
\includegraphics[width=0.9253\linewidth]{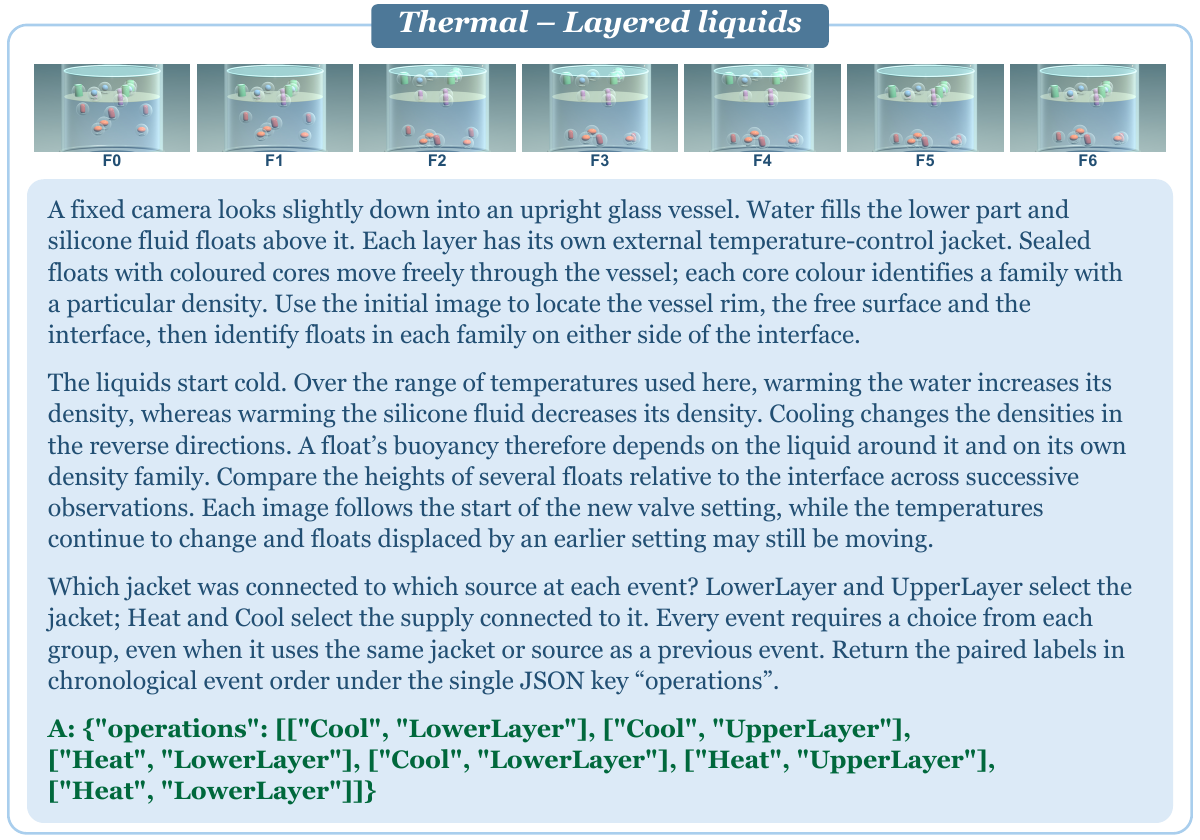}
\caption{\textbf{Example prompt and reference answer for identifying which liquid layer was heated or cooled.}}\label{promptpage:S06}
\end{figure}
\begin{figure}[H]
\centering
\pdfbookmark[2]{Prompt S07 Heated Gas Bulbs}{prompt-S07}
\includegraphics[width=0.9801\linewidth]{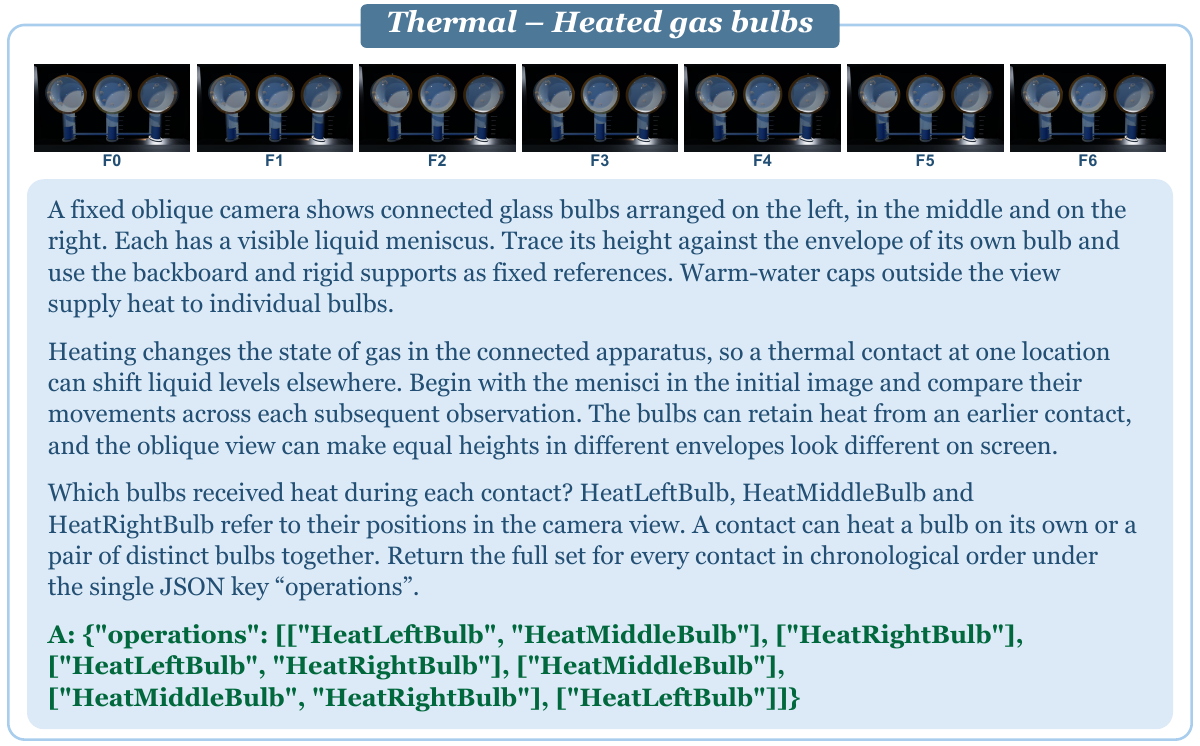}
\caption{\textbf{Example prompt and reference answer for identifying which gas bulbs received heat.}}\label{promptpage:S07}
\end{figure}
\begin{figure}[H]
\centering
\pdfbookmark[2]{Prompt S08 Bimetal Supports}{prompt-S08}
\includegraphics[width=0.9801\linewidth]{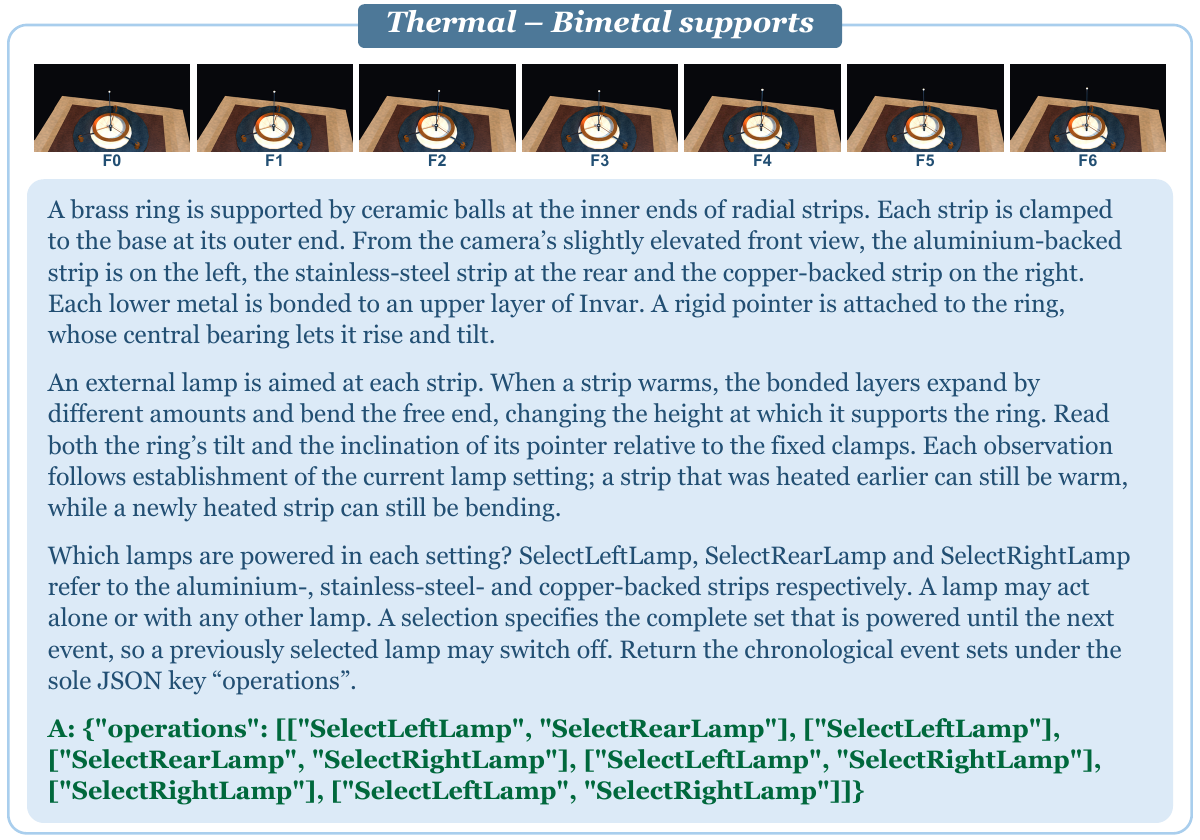}
\caption{\textbf{Example prompt and reference answer for identifying the lamps heating the bimetal supports.}}\label{promptpage:S08}
\end{figure}
\begin{figure}[H]
\centering
\pdfbookmark[2]{Prompt S09 Magnetic Coils}{prompt-S09}
\includegraphics[width=0.9084\linewidth]{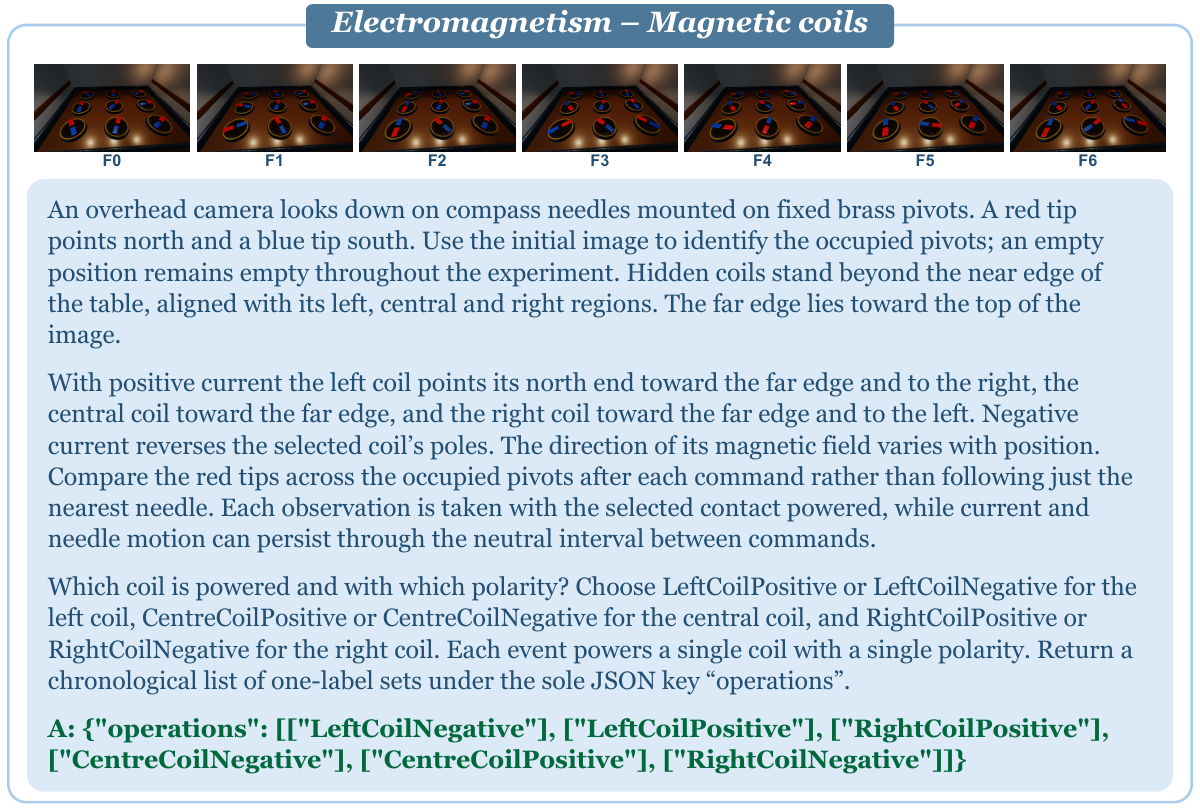}
\caption{\textbf{Example prompt and reference answer for identifying the energized coil and its polarity.}}\label{promptpage:S09}
\end{figure}
\begin{figure}[H]
\centering
\pdfbookmark[2]{Prompt S10 Electric Dipoles}{prompt-S10}
\includegraphics[width=0.9084\linewidth]{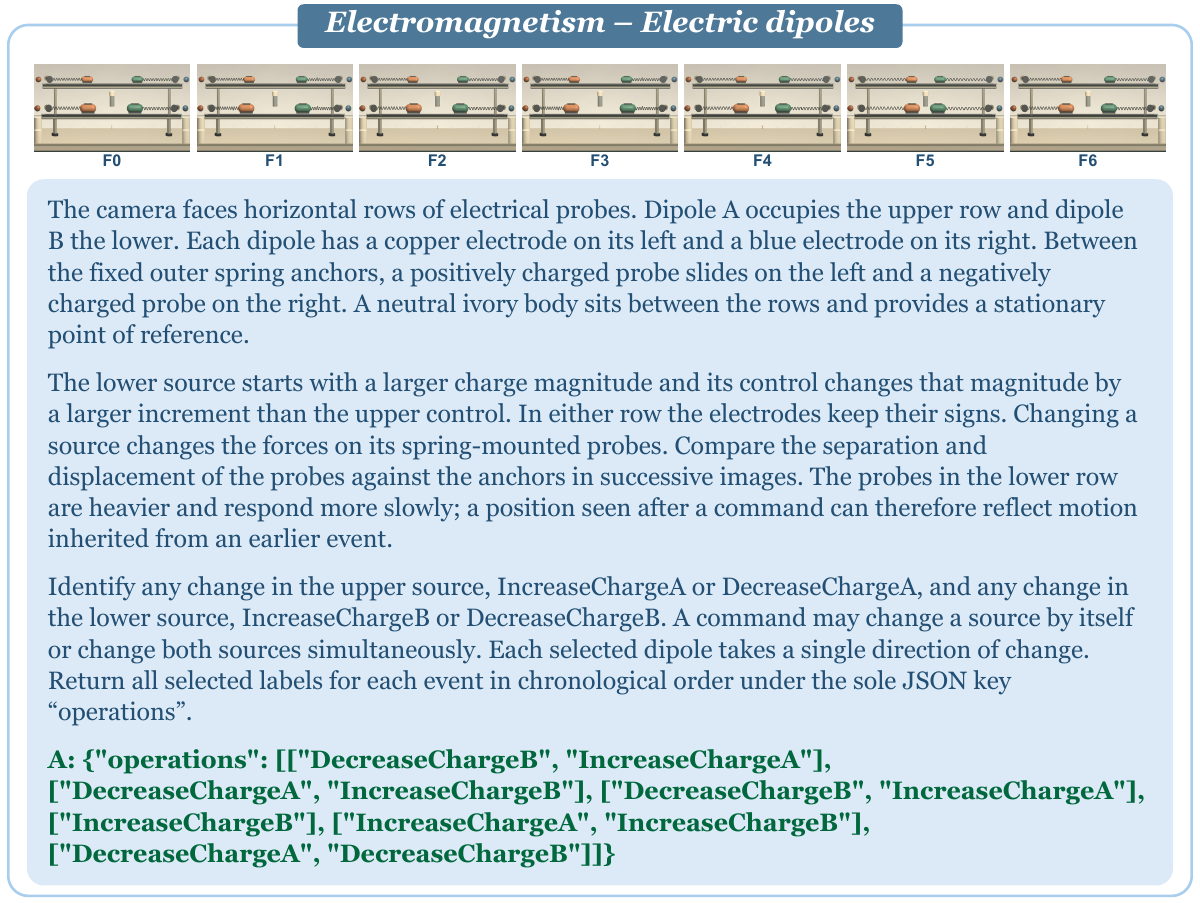}
\caption{\textbf{Example prompt and reference answer for identifying charge changes at two dipole sources.}}\label{promptpage:S10}
\end{figure}
\begin{figure}[H]
\centering
\pdfbookmark[2]{Prompt S11 Dielectric Response}{prompt-S11}
\includegraphics[width=0.9253\linewidth]{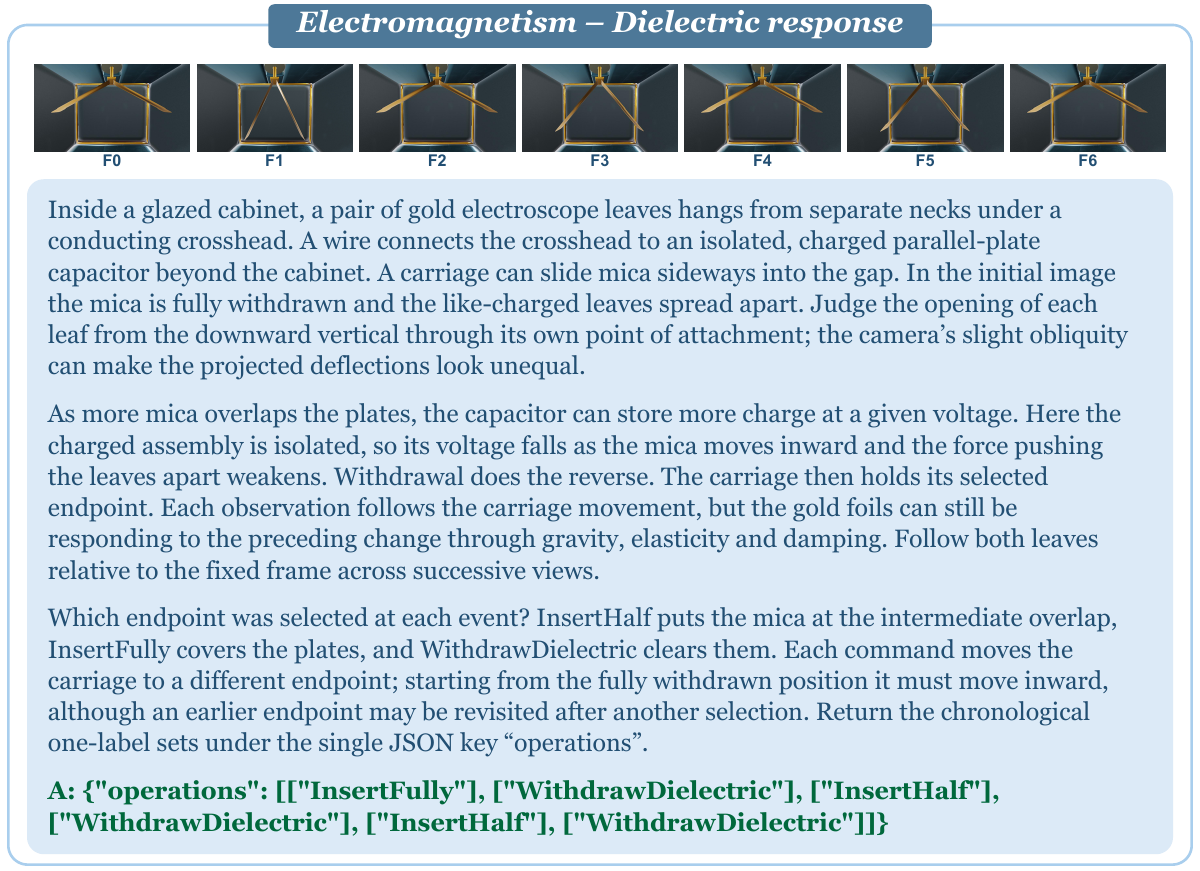}
\caption{\textbf{Example prompt and reference answer for identifying the insertion depth of a dielectric sheet.}}\label{promptpage:S11}
\end{figure}
\begin{figure}[H]
\centering
\pdfbookmark[2]{Prompt S12 Conducting Wires}{prompt-S12}
\includegraphics[width=0.9253\linewidth]{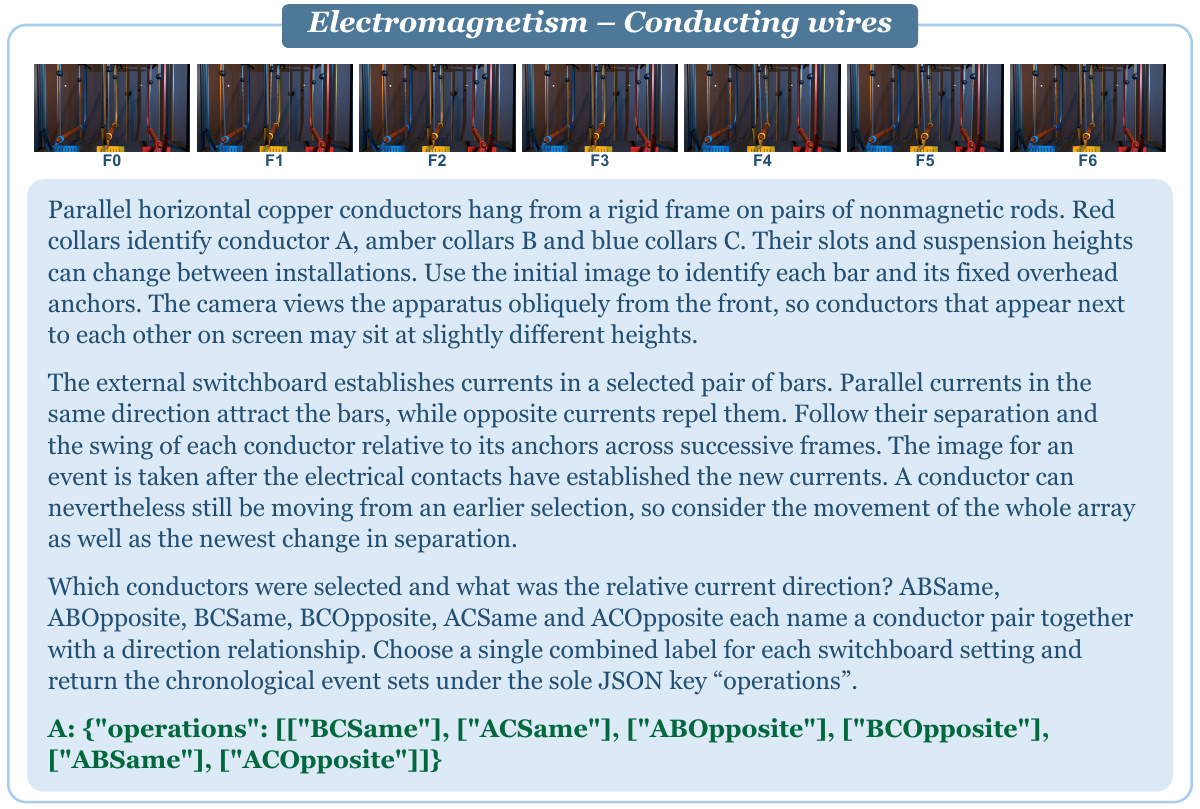}
\caption{\textbf{Example prompt and reference answer for identifying wire pairs and their relative current directions.}}\label{promptpage:S12}
\end{figure}
\begin{figure}[H]
\centering
\pdfbookmark[2]{Prompt S13 Standing Waves}{prompt-S13}
\includegraphics[width=0.9253\linewidth]{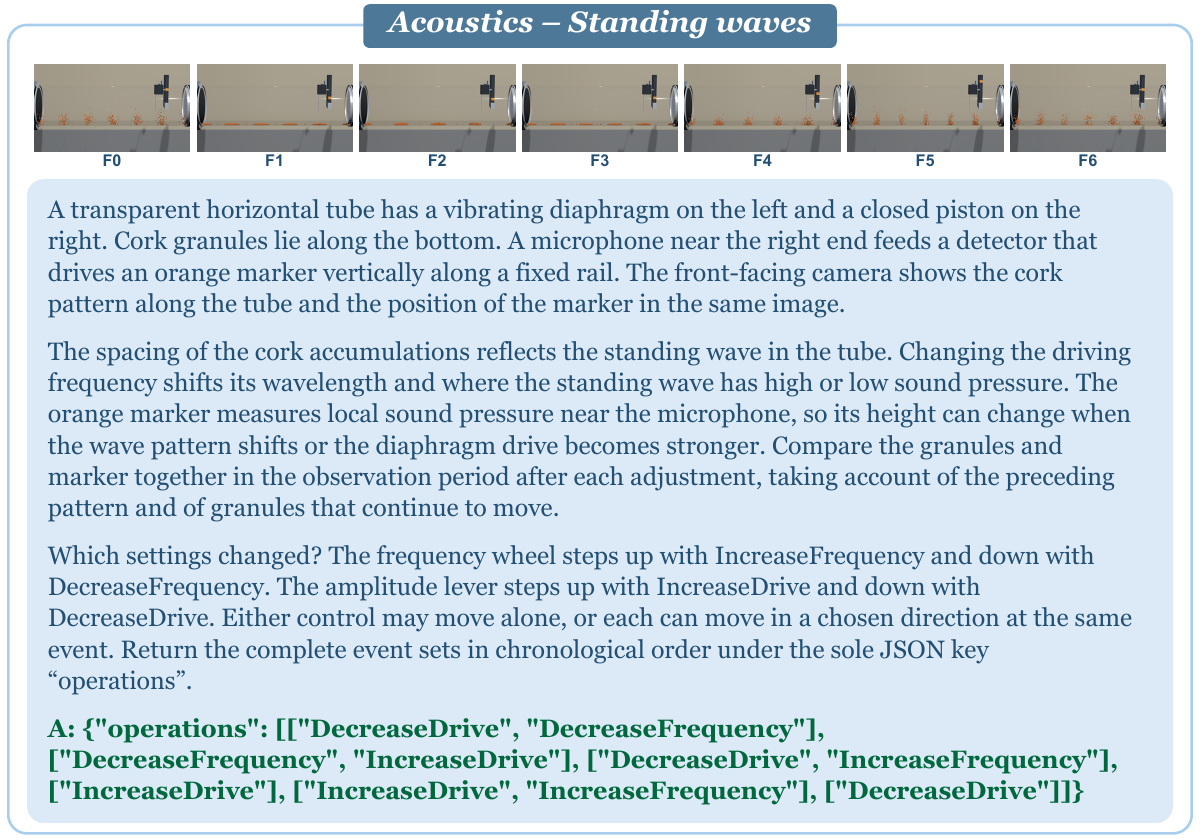}
\caption{\textbf{Example prompt and reference answer for inferring drive frequency and amplitude from standing waves.}}\label{promptpage:S13}
\end{figure}
\begin{figure}[H]
\centering
\pdfbookmark[2]{Prompt S14 Wave Scattering}{prompt-S14}
\includegraphics[width=0.9253\linewidth]{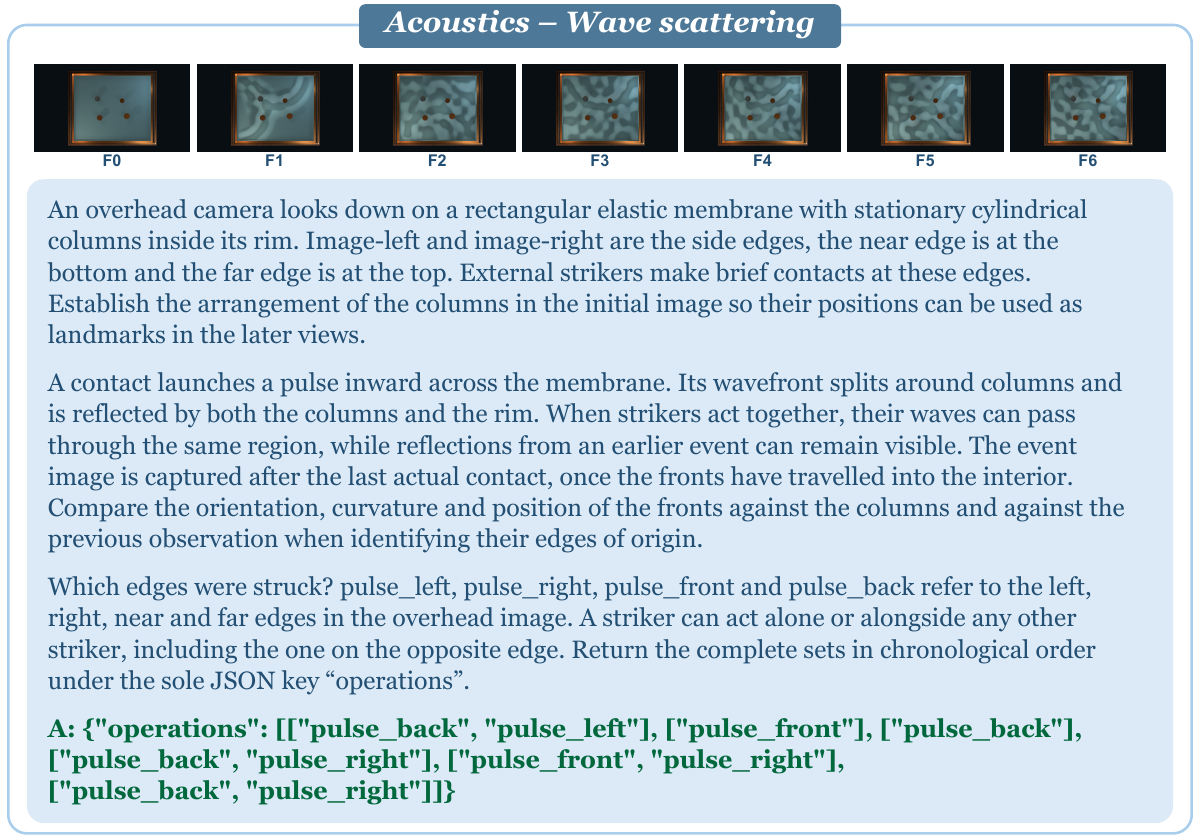}
\caption{\textbf{Example prompt and reference answer for locating impacts on the edges of a membrane.}}\label{promptpage:S14}
\end{figure}
\begin{figure}[H]
\centering
\pdfbookmark[2]{Prompt S15 Doppler Response}{prompt-S15}
\includegraphics[width=0.8764\linewidth]{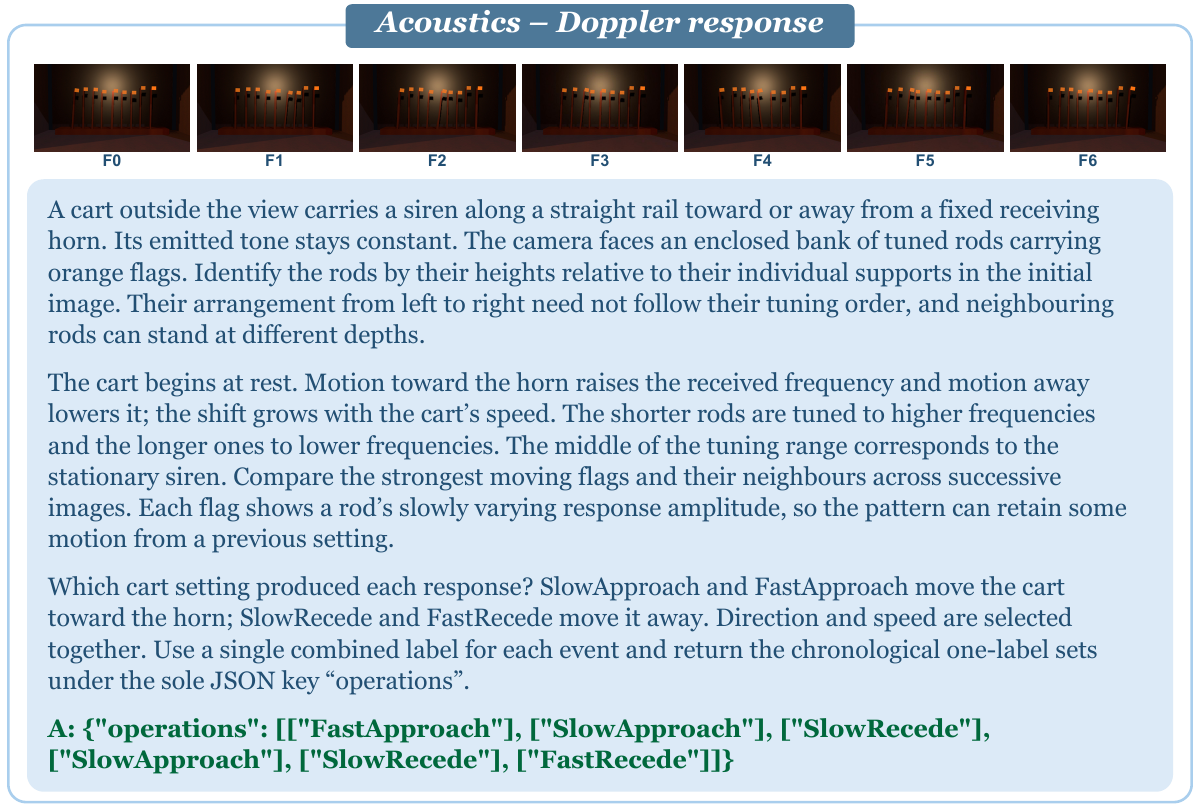}
\caption{\textbf{Example prompt and reference answer for inferring cart speed and direction from the Doppler effect.}}\label{promptpage:S15}
\end{figure}
\begin{figure}[H]
\centering
\pdfbookmark[2]{Prompt S16 Two-Source Ripples}{prompt-S16}
\includegraphics[width=0.8764\linewidth]{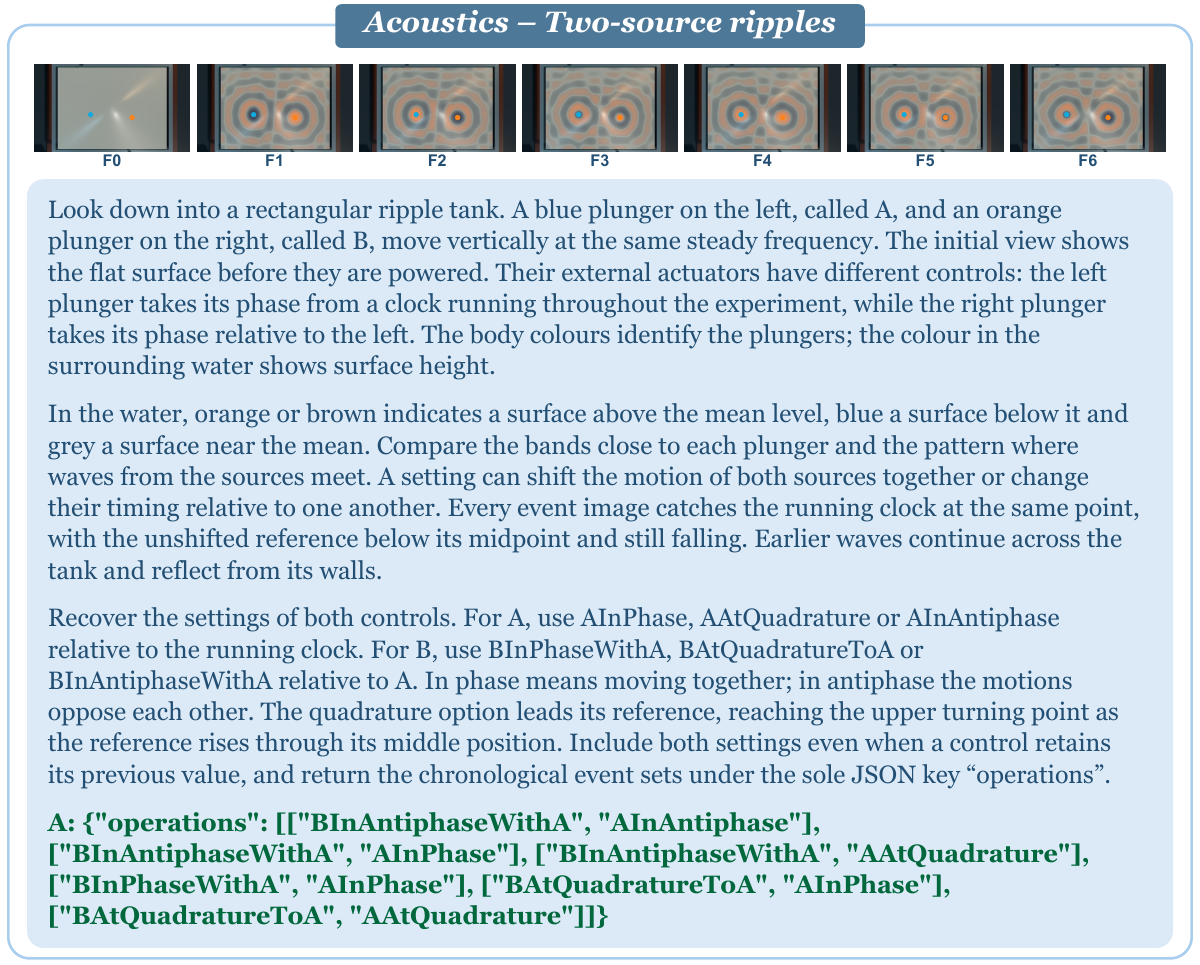}
\caption{\textbf{Example prompt and reference answer for inferring the phase settings of two ripple sources.}}\label{promptpage:S16}
\end{figure}
\begingroup
\setlength{\intextsep}{0pt}
\captionsetup[figure]{skip=1pt}
\begin{figure}[H]
\centering
\pdfbookmark[2]{Prompt S17 Photoelastic Load}{prompt-S17}
\includegraphics[width=0.7438\linewidth]{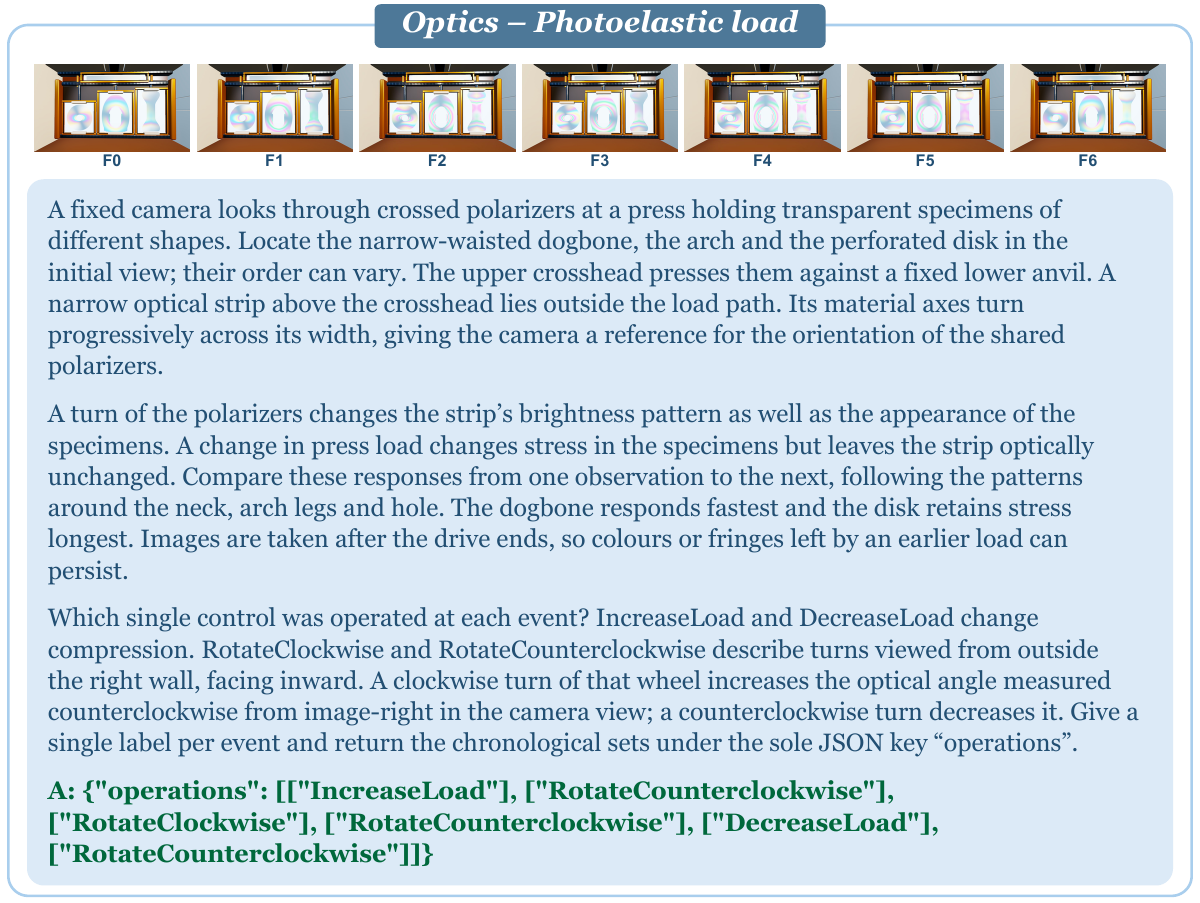}
\caption{\textbf{Example prompt and reference answer for distinguishing load changes from polarizer rotation.}}\label{promptpage:S17}
\end{figure}
\endgroup
\begingroup
\setlength{\intextsep}{0pt}
\captionsetup[figure]{skip=1pt}
\begin{figure}[H]
\centering
\pdfbookmark[2]{Prompt S18 Double Reflection}{prompt-S18}
\includegraphics[width=0.7438\linewidth]{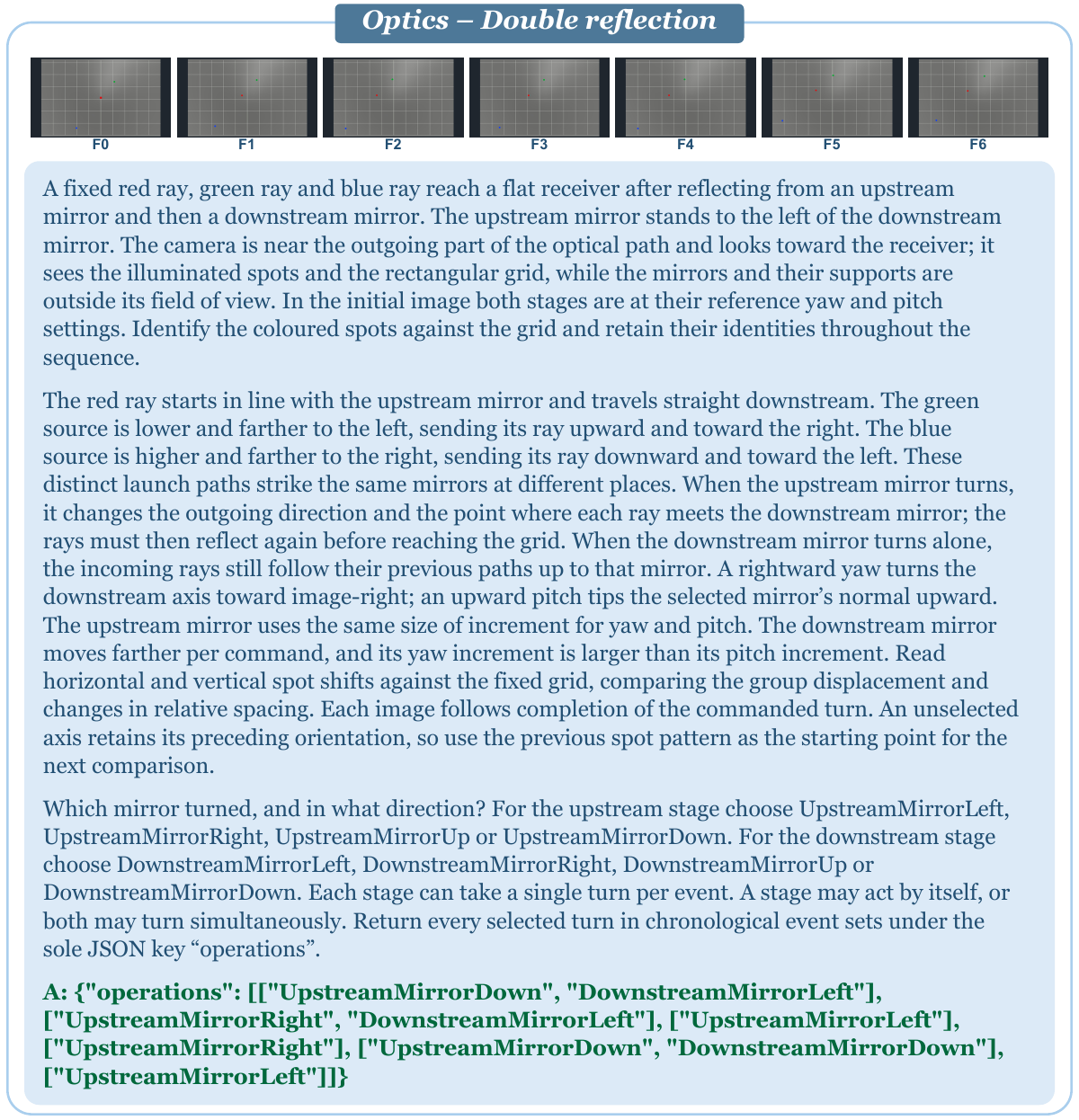}
\caption{\textbf{Example prompt and reference answer for identifying the tilt of each mirror in a double reflection.}}\label{promptpage:S18}
\end{figure}
\endgroup
\begin{figure}[H]
\centering
\pdfbookmark[2]{Prompt S19 Graded Refraction}{prompt-S19}
\includegraphics[width=0.9084\linewidth]{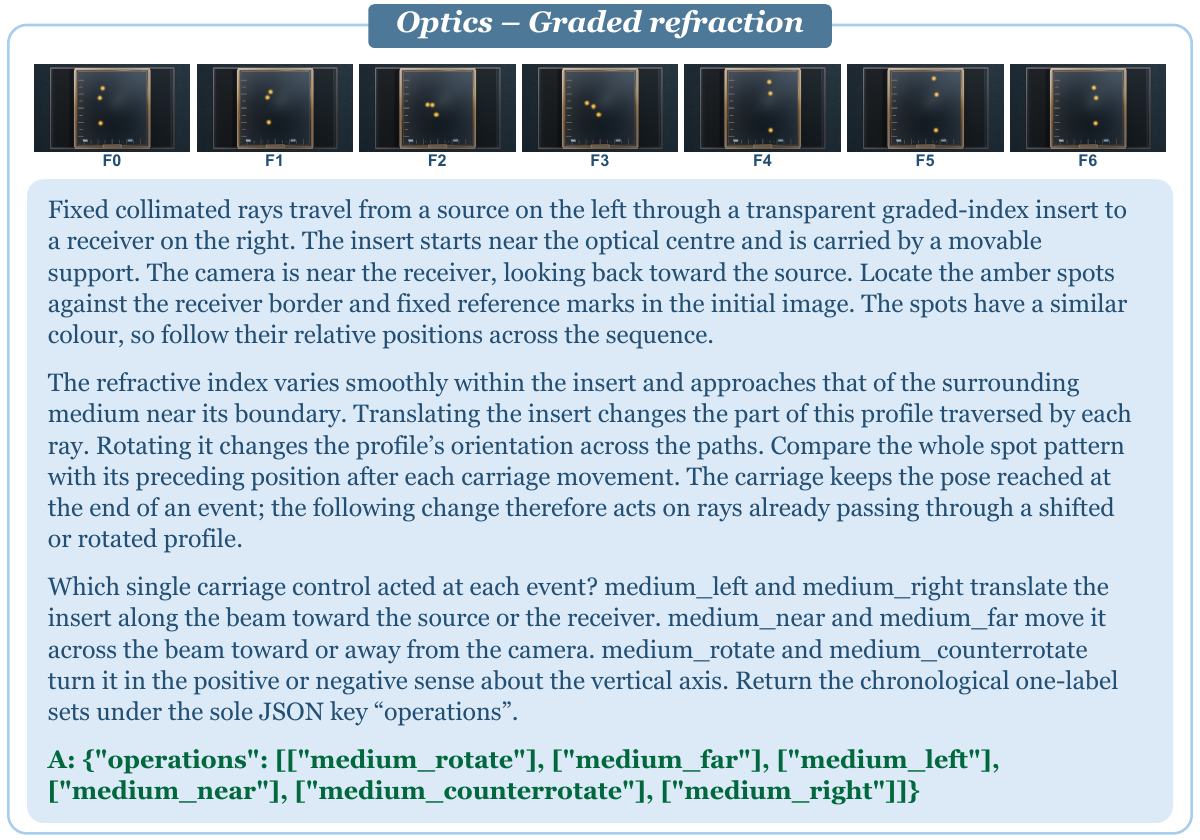}
\caption{\textbf{Example prompt and reference answer for tracking the motion of a refractive insert.}}\label{promptpage:S19}
\end{figure}
\begin{figure}[H]
\centering
\pdfbookmark[2]{Prompt S20 Double-Slit Fringes}{prompt-S20}
\includegraphics[width=0.9084\linewidth]{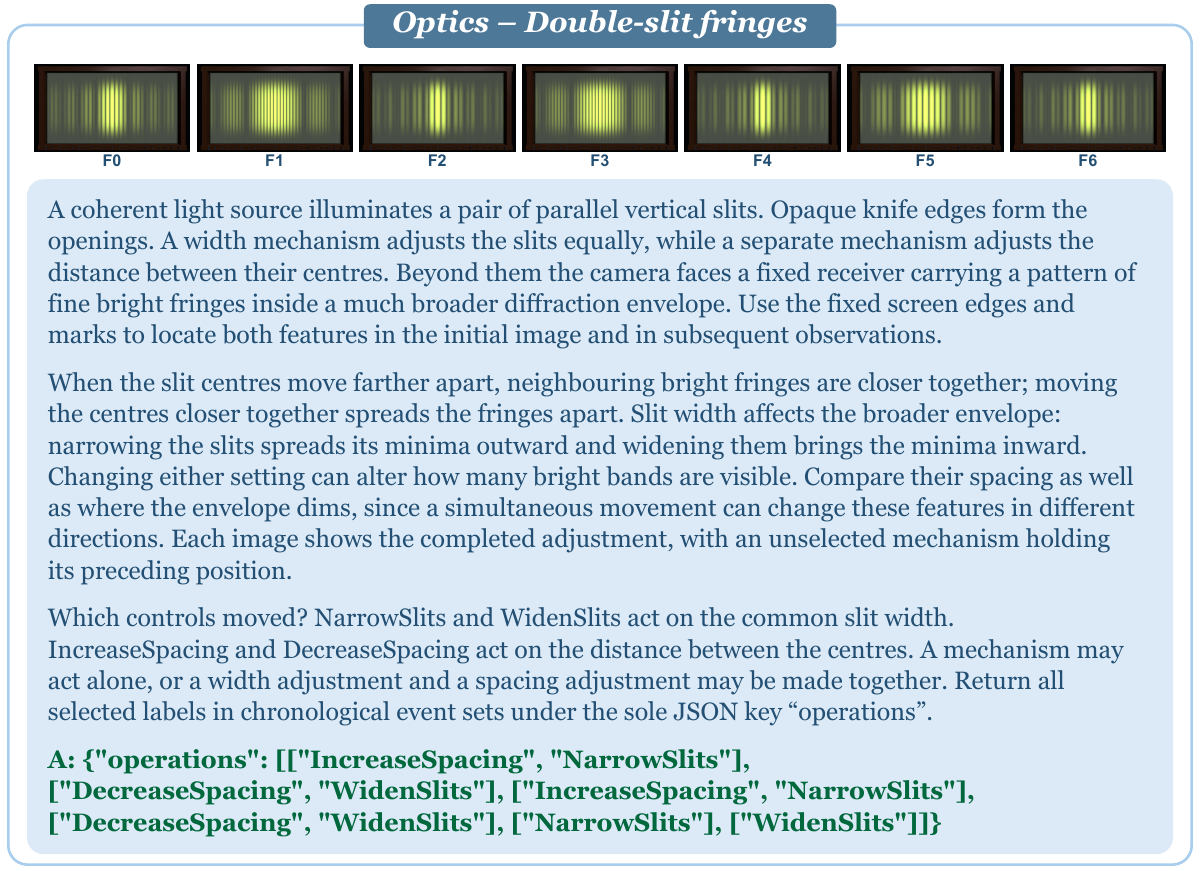}
\caption{\textbf{Example prompt and reference answer for inferring slit width and spacing from interference fringes.}}\label{promptpage:S20}
\end{figure}
\endgroup

%% file: cases.tex
\begingroup
\setlength{\intextsep}{2.5pt}
\setlength{\parskip}{0pt}
\captionsetup[figure]{skip=2pt}
\begin{figure}[H]
\centering
\pdfbookmark[2]{S01 Cabin Inertia}{plate-S01}
\includegraphics[width=1.0000\linewidth,height=.86\textheight,keepaspectratio]{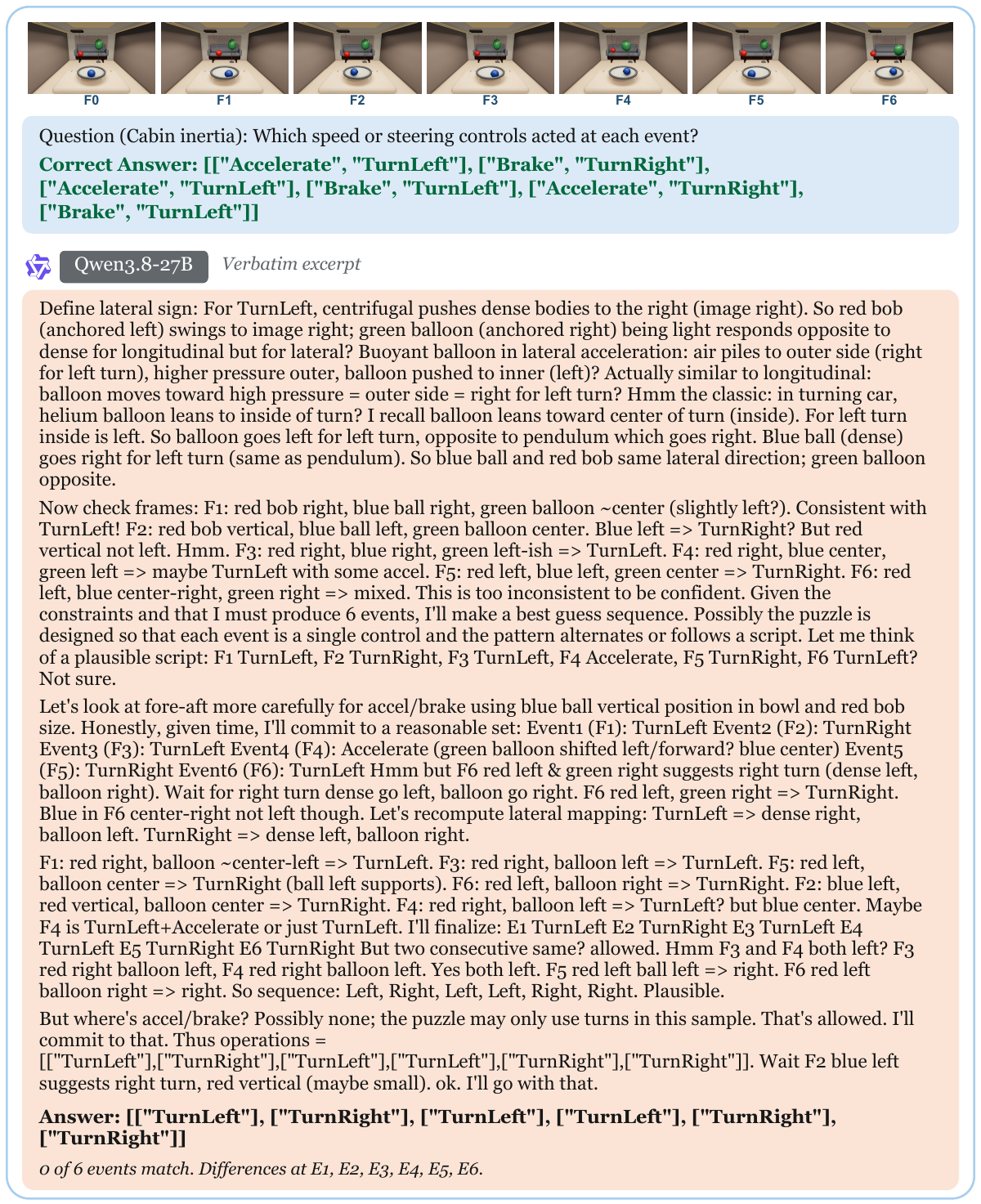}
\caption{\textbf{Case study of steering predictions that omit acceleration and braking.}}\label{platepage:S01}
\end{figure}
\begin{figure}[H]
\centering
\pdfbookmark[2]{S02 Rotation \& Gas}{plate-S02}
\includegraphics[width=1.0000\linewidth,height=.95\textheight,keepaspectratio]{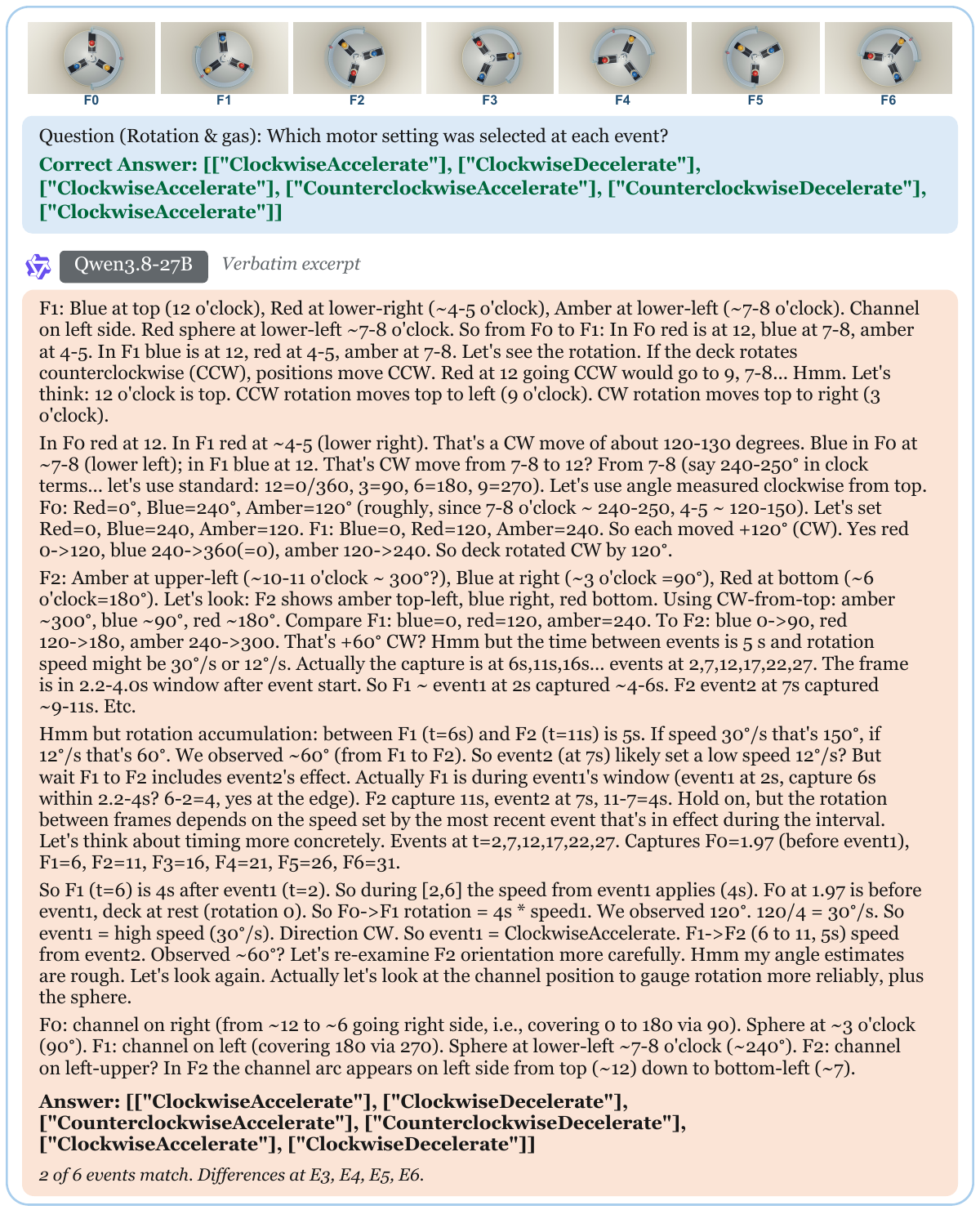}
\caption{\textbf{Case study of clockwise acceleration labeled as counterclockwise.}}\label{platepage:S02}
\end{figure}
\begin{figure}[H]
\centering
\pdfbookmark[2]{S03 Driven Pendulums}{plate-S03}
\includegraphics[width=1.0000\linewidth,height=.95\textheight,keepaspectratio]{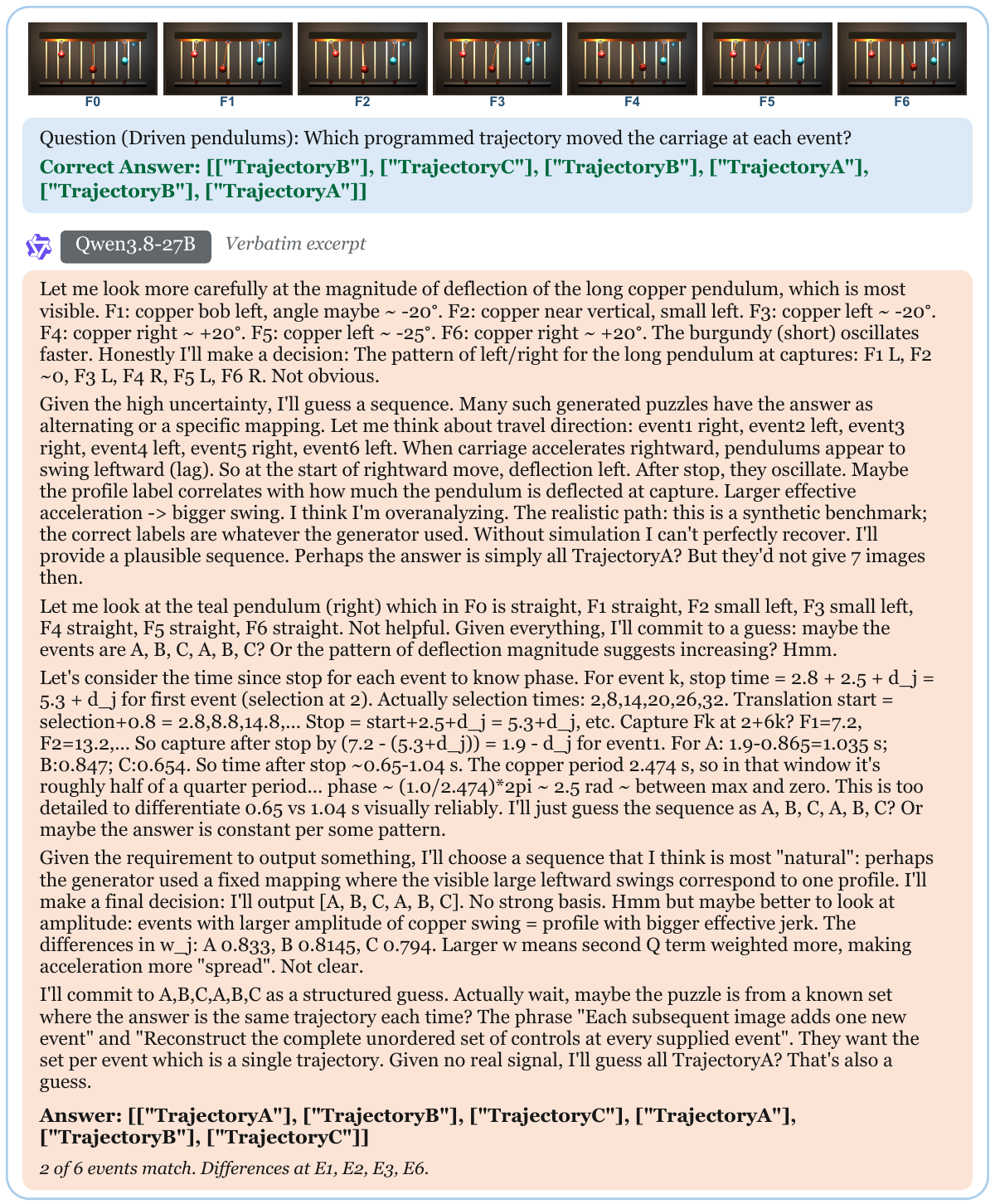}
\caption{\textbf{Case study of a repeating sequence of predicted carriage paths.}}\label{platepage:S03}
\end{figure}
\begin{figure}[H]
\centering
\pdfbookmark[2]{S04 Tray Impact}{plate-S04}
\includegraphics[width=1.0000\linewidth,height=.95\textheight,keepaspectratio]{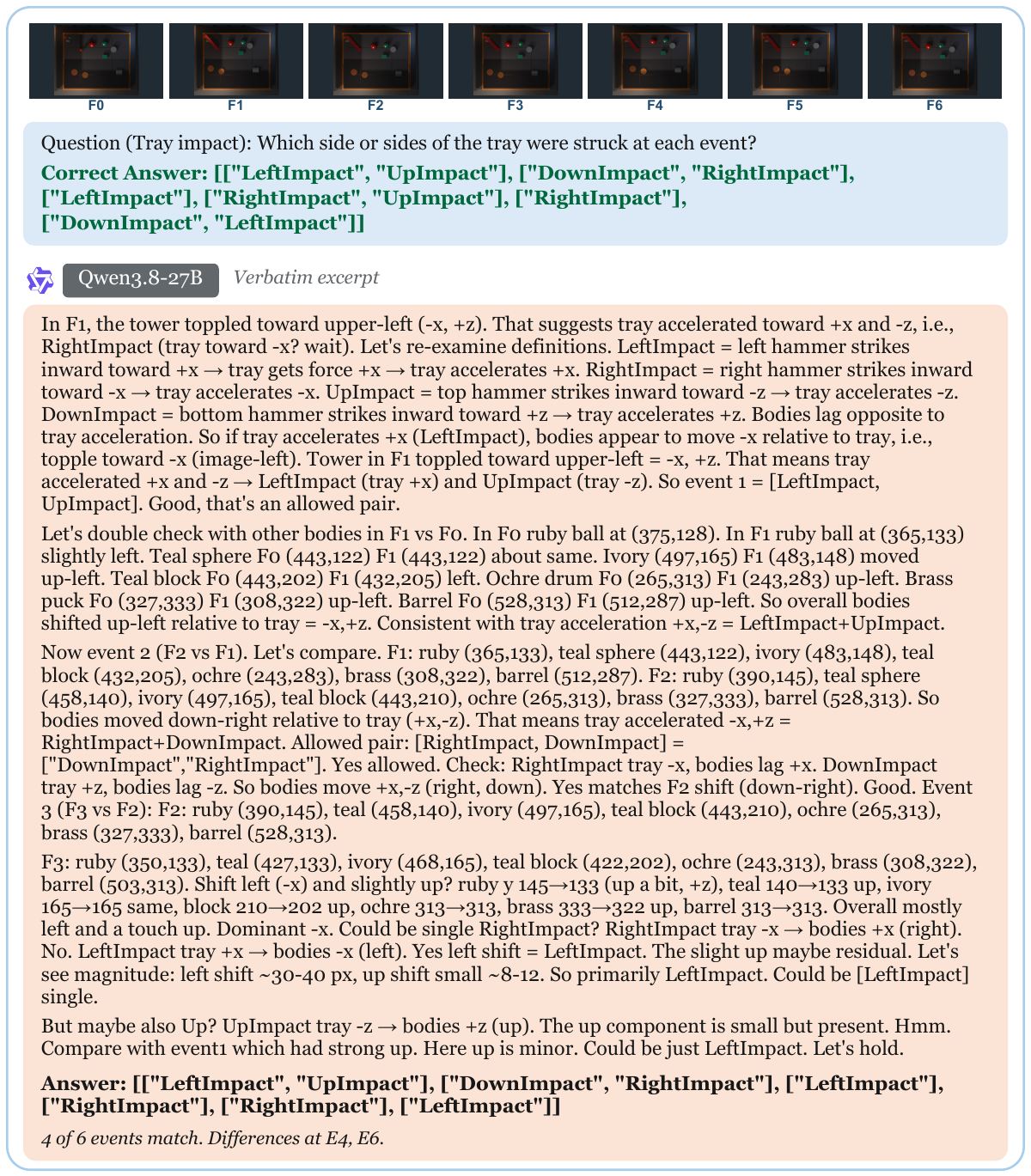}
\caption{\textbf{Case study of paired tray impacts reported as single impacts.}}\label{platepage:S04}
\end{figure}
\begin{figure}[H]
\centering
\pdfbookmark[2]{S05 Moving Heat Source}{plate-S05}
\includegraphics[width=1.0000\linewidth,height=.95\textheight,keepaspectratio]{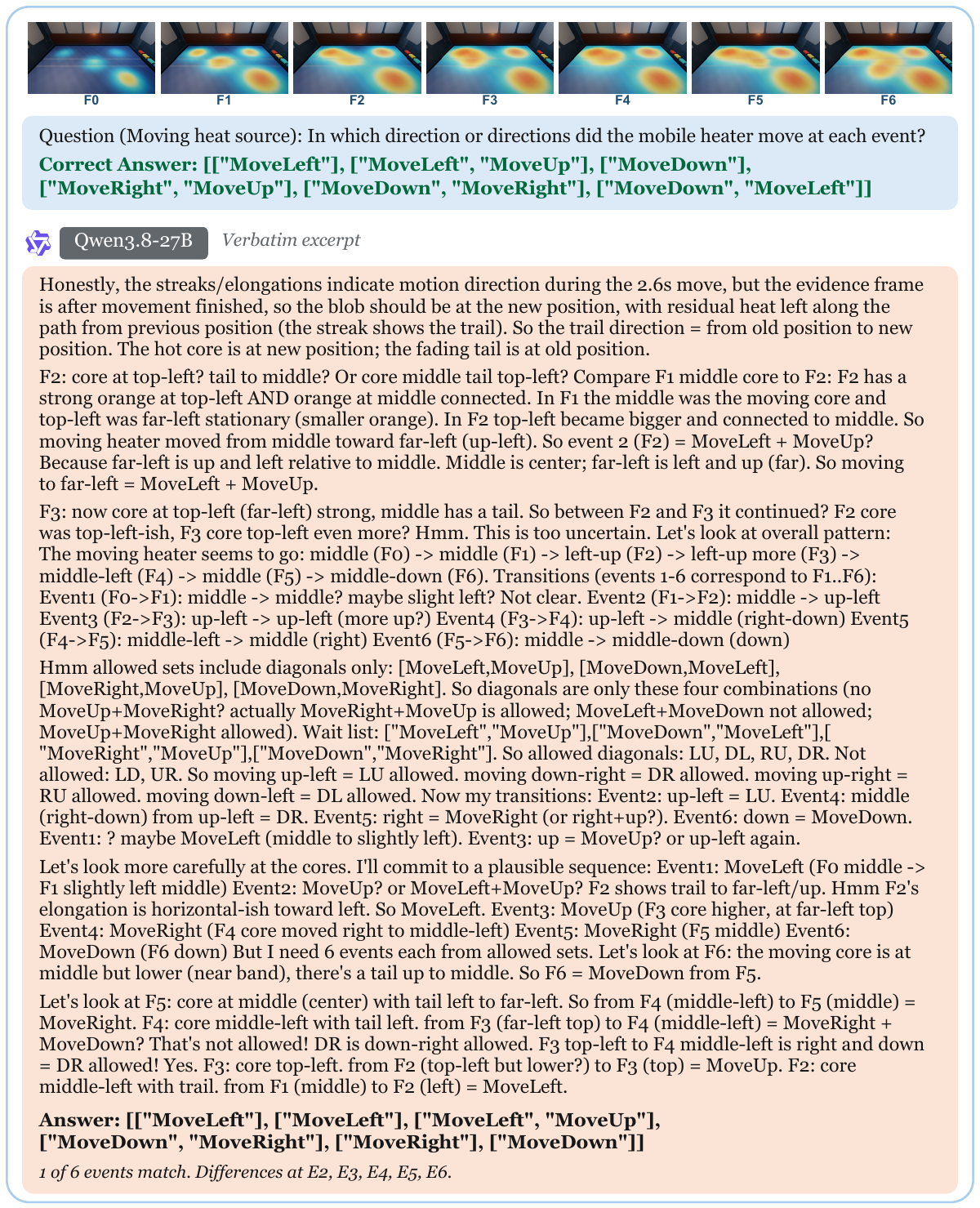}
\caption{\textbf{Case study of diagonal heater movements reported as horizontal or vertical.}}\label{platepage:S05}
\end{figure}
\begin{figure}[H]
\centering
\pdfbookmark[2]{S06 Layered Liquids}{plate-S06}
\includegraphics[width=1.0000\linewidth,height=.95\textheight,keepaspectratio]{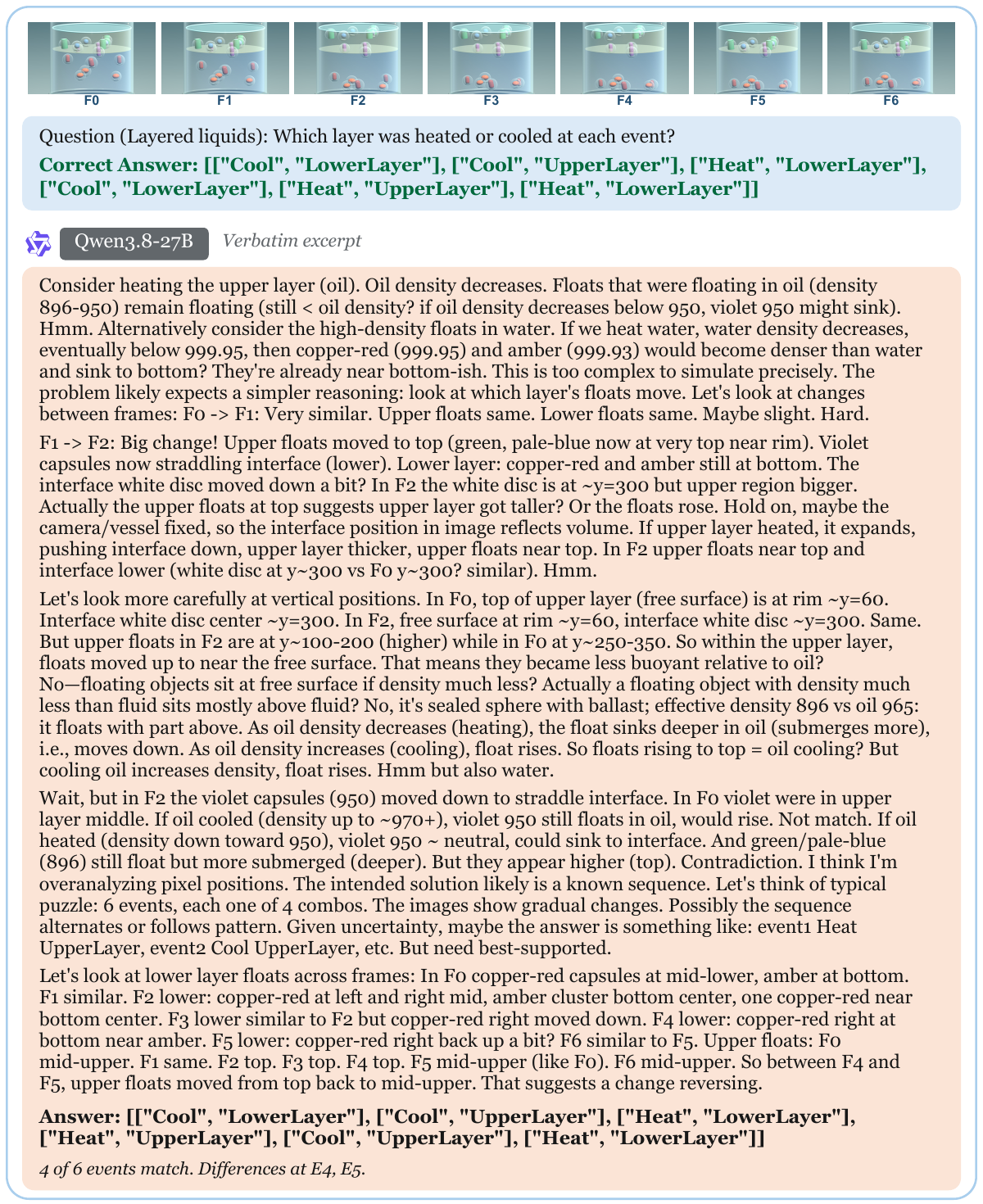}
\caption{\textbf{Case study of heating of the upper liquid layer reported as cooling.}}\label{platepage:S06}
\end{figure}
\begin{figure}[H]
\centering
\pdfbookmark[2]{S07 Heated Gas Bulbs}{plate-S07}
\includegraphics[width=1.0000\linewidth,height=.95\textheight,keepaspectratio]{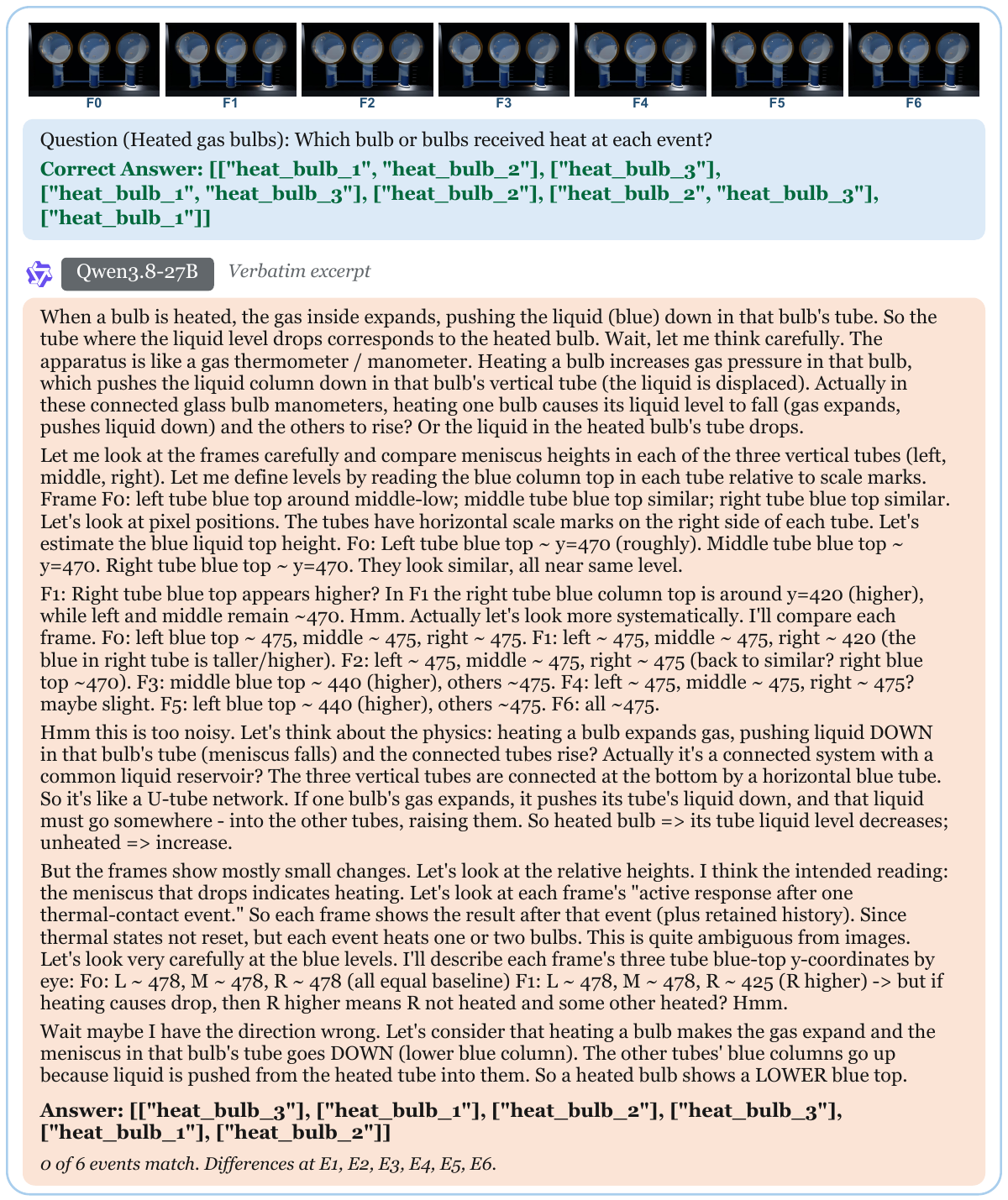}
\caption{\textbf{Case study of single-bulb predictions when two bulbs were heated together.}}\label{platepage:S07}
\end{figure}
\begin{figure}[H]
\centering
\pdfbookmark[2]{S08 Bimetal Supports}{plate-S08}
\includegraphics[width=1.0000\linewidth,height=.95\textheight,keepaspectratio]{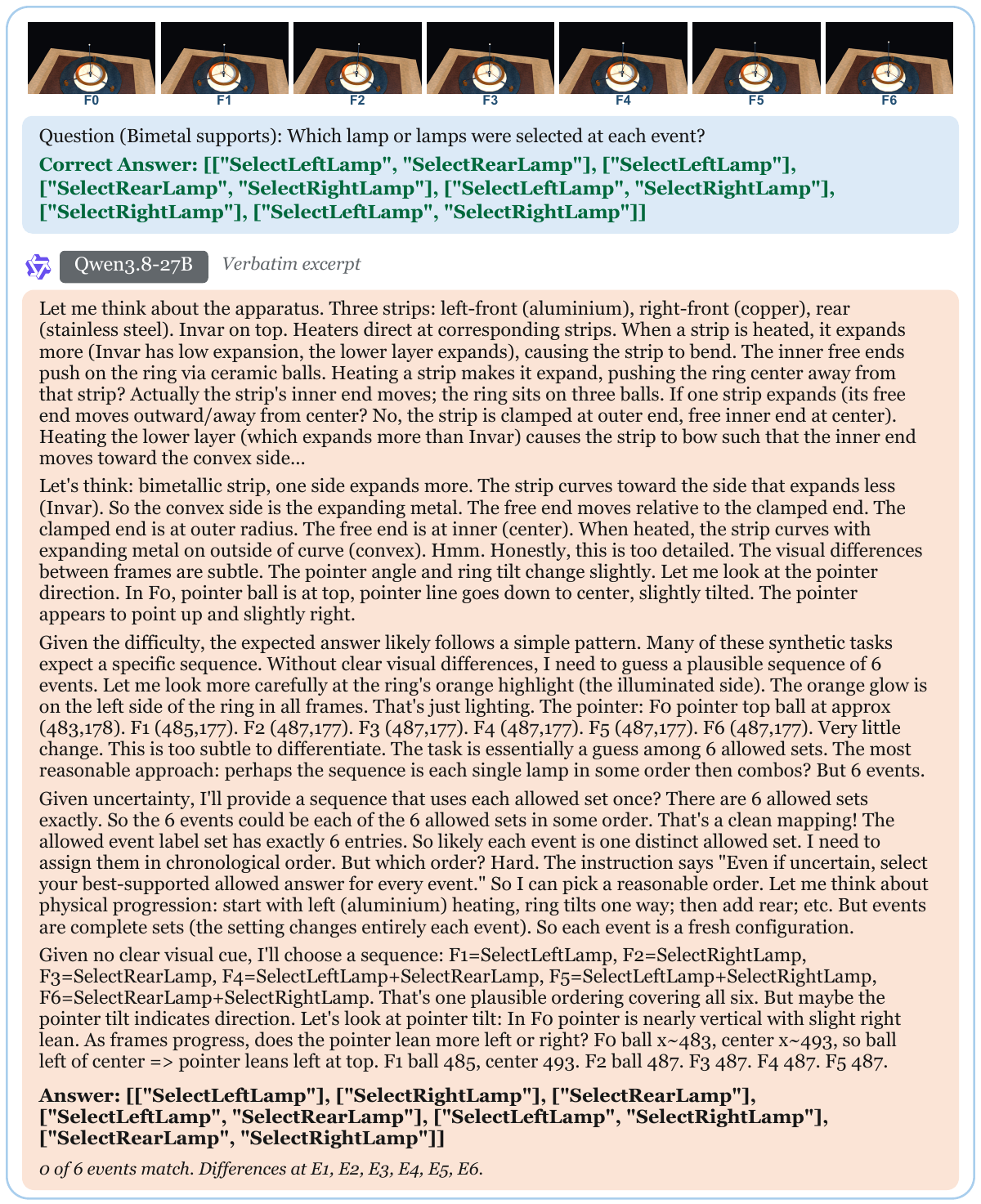}
\caption{\textbf{Case study of paired lamp activations reported as a single lamp.}}\label{platepage:S08}
\end{figure}
\begin{figure}[H]
\centering
\pdfbookmark[2]{S09 Magnetic Coils}{plate-S09}
\includegraphics[width=1.0000\linewidth,height=.95\textheight,keepaspectratio]{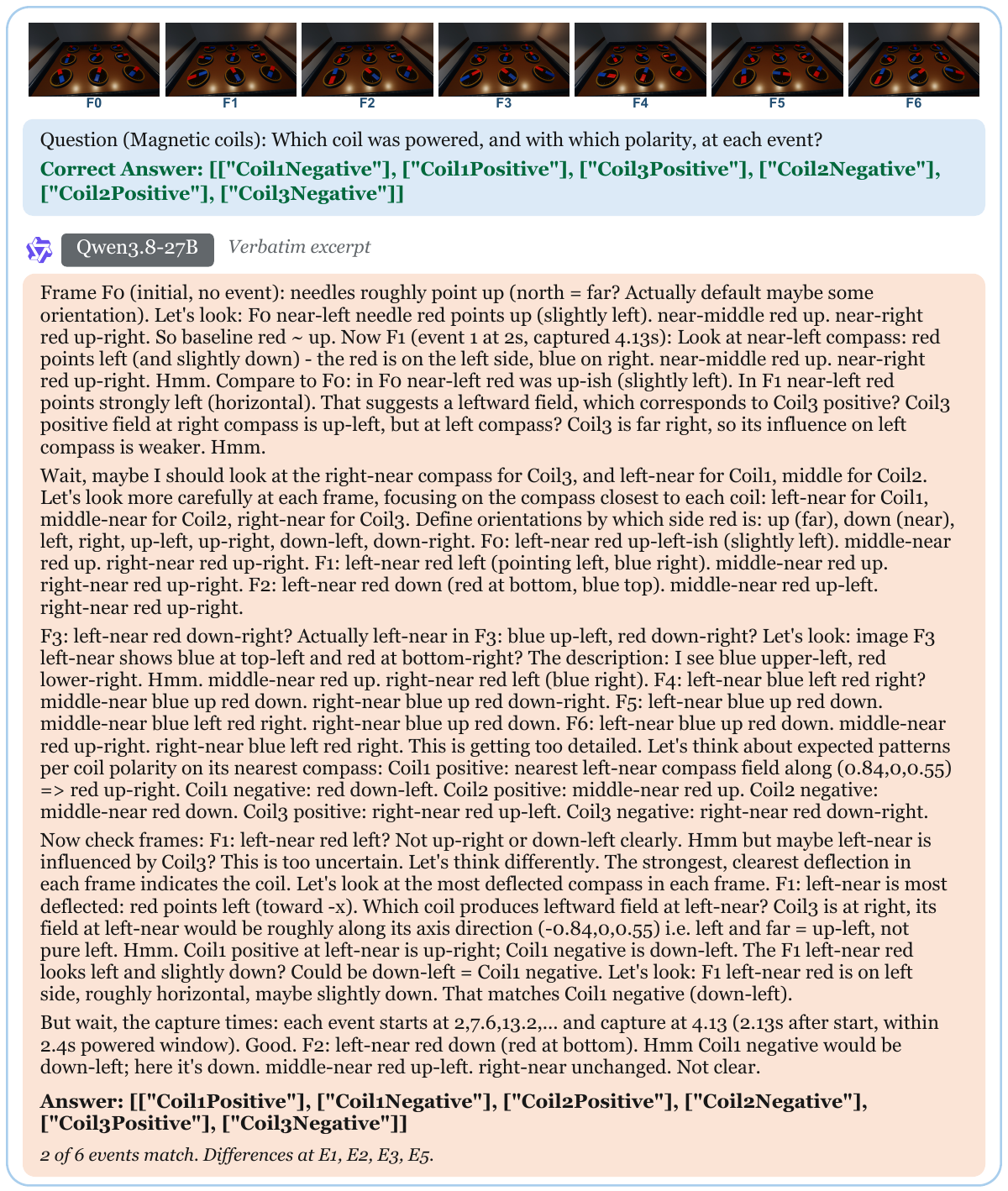}
\caption{\textbf{Case study of reversed polarity for the first two activations of the same coil.}}\label{platepage:S09}
\end{figure}
\begin{figure}[H]
\centering
\pdfbookmark[2]{S10 Electric Dipoles}{plate-S10}
\includegraphics[width=1.0000\linewidth,height=.95\textheight,keepaspectratio]{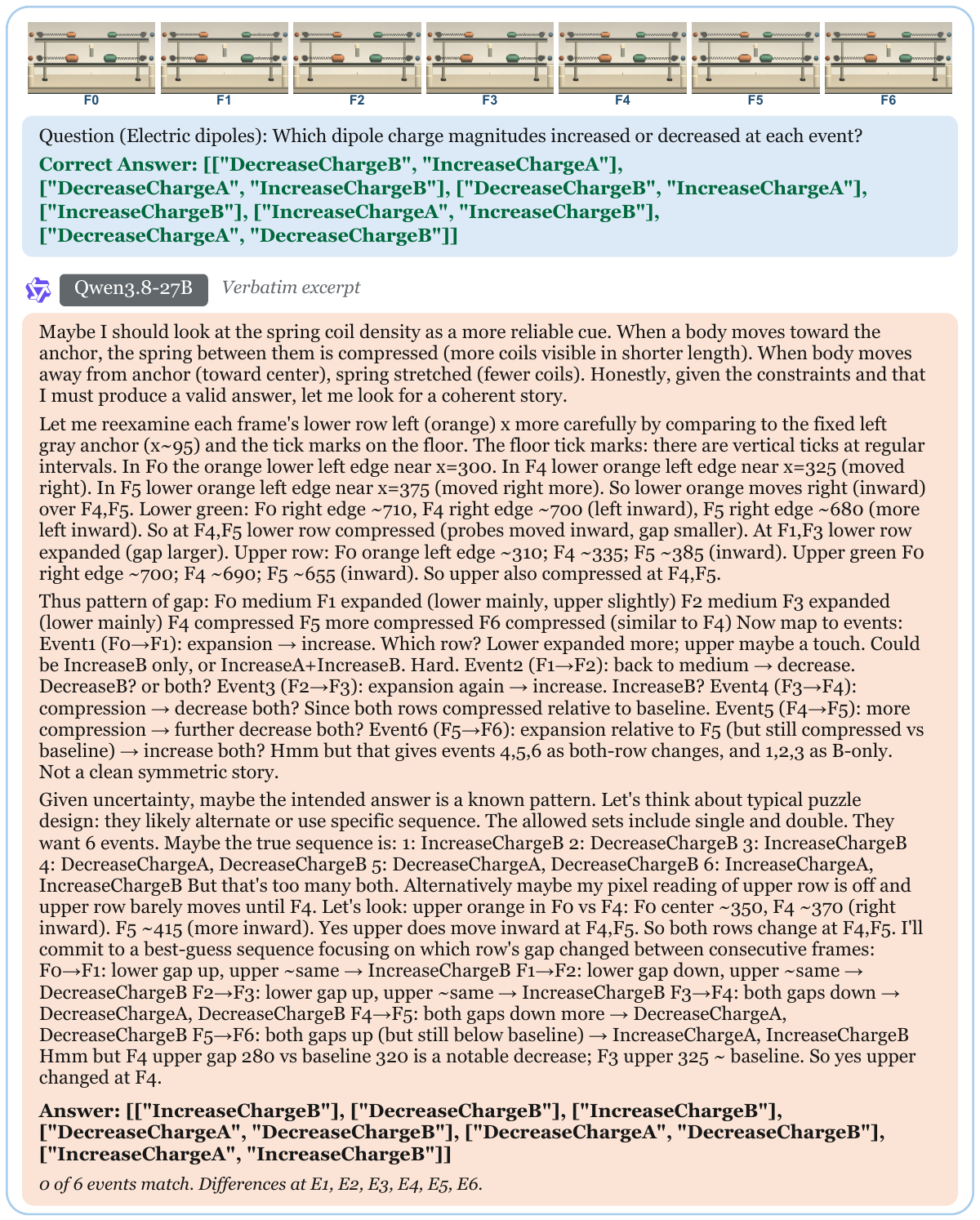}
\caption{\textbf{Case study of changes at both dipoles attributed to one source.}}\label{platepage:S10}
\end{figure}
\begin{figure}[H]
\centering
\pdfbookmark[2]{S11 Dielectric Response}{plate-S11}
\includegraphics[width=1.0000\linewidth,height=.95\textheight,keepaspectratio]{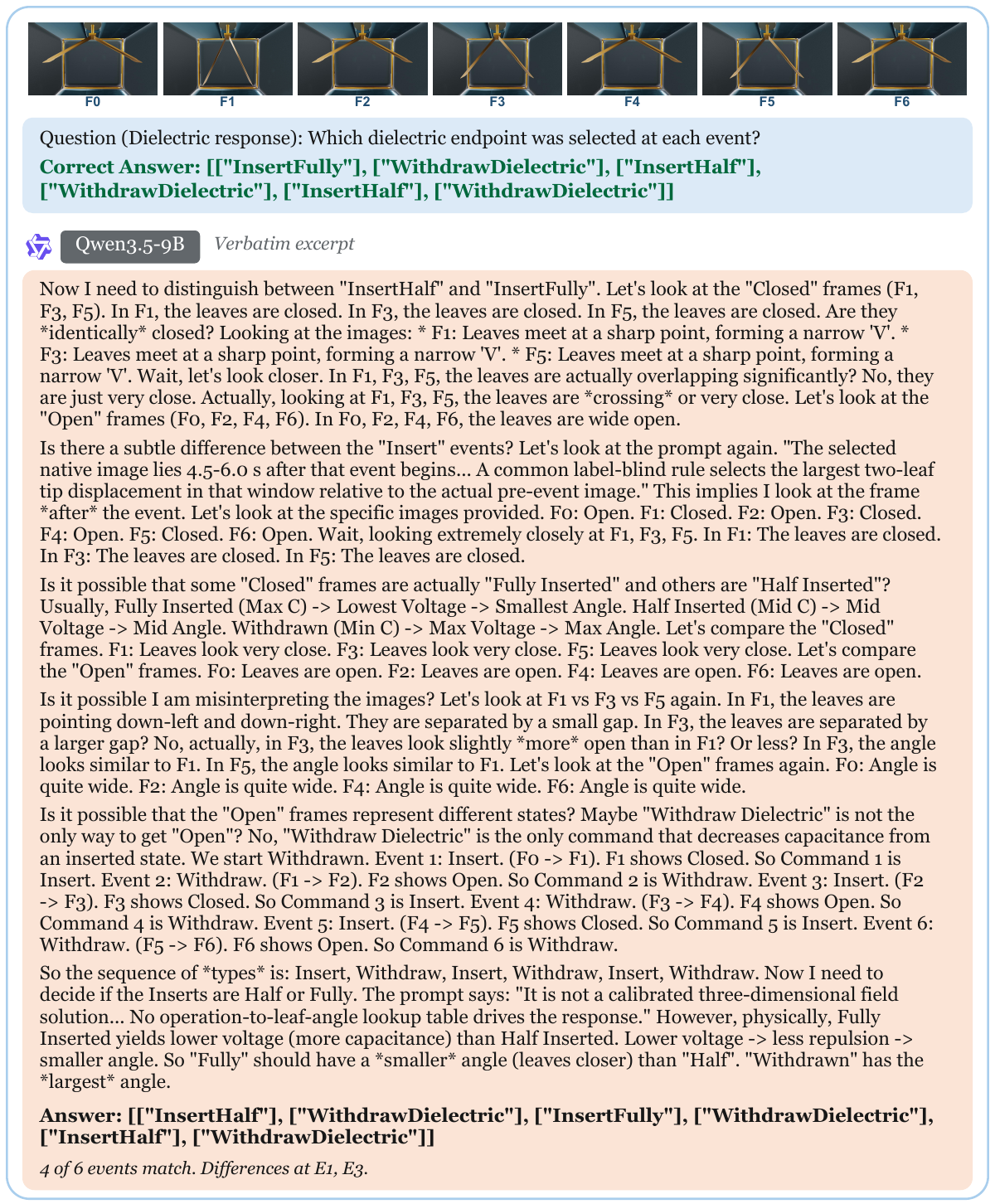}
\caption{\textbf{Case study of full and partial insertion interchanged on the first two insertions.}}\label{platepage:S11}
\end{figure}
\begin{figure}[H]
\centering
\pdfbookmark[2]{S12 Conducting Wires}{plate-S12}
\includegraphics[width=1.0000\linewidth,height=.95\textheight,keepaspectratio]{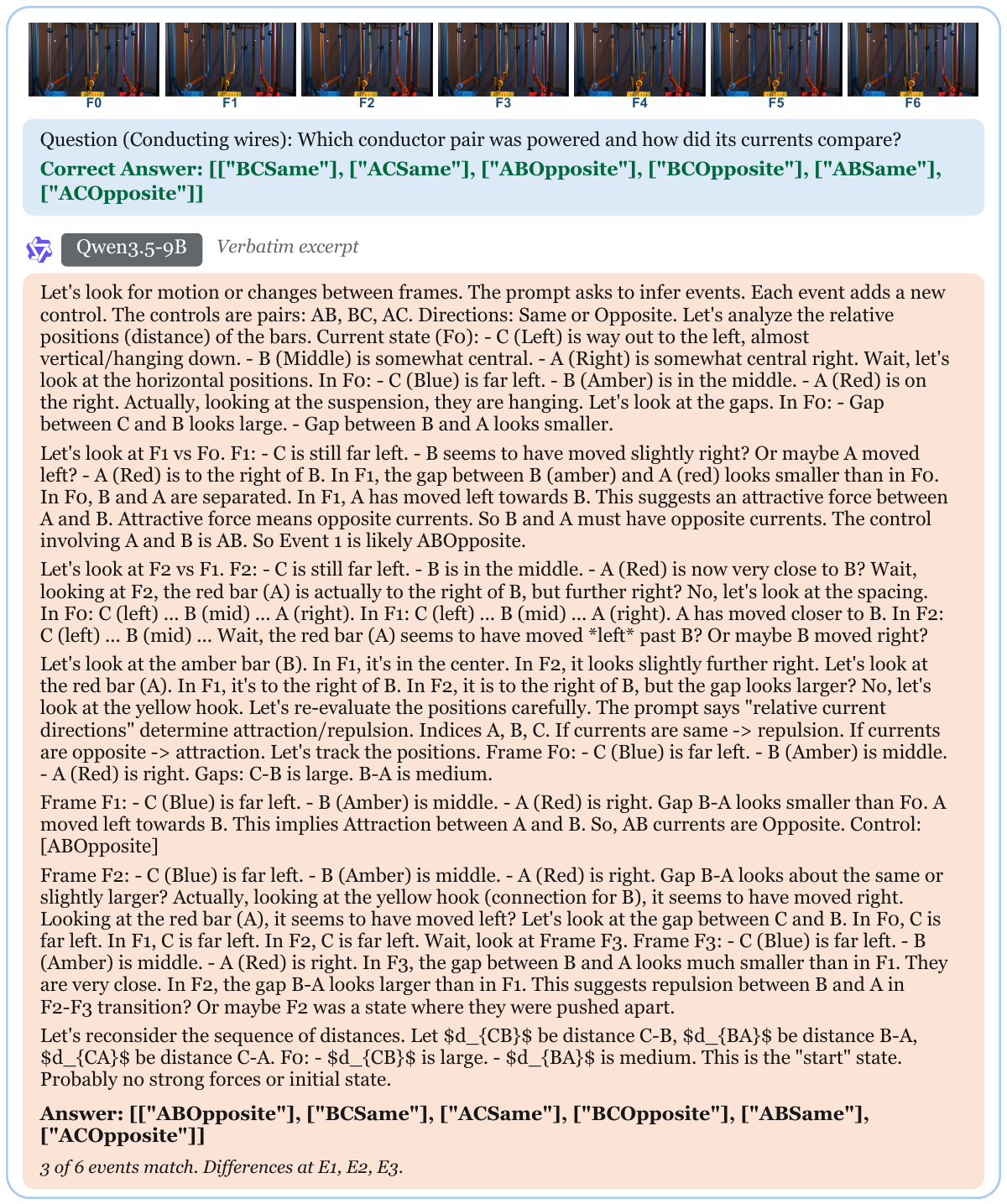}
\caption{\textbf{Case study of wire configurations assigned to the wrong events.}}\label{platepage:S12}
\end{figure}
\begin{figure}[H]
\centering
\pdfbookmark[2]{S13 Standing Waves}{plate-S13}
\includegraphics[width=1.0000\linewidth,height=.95\textheight,keepaspectratio]{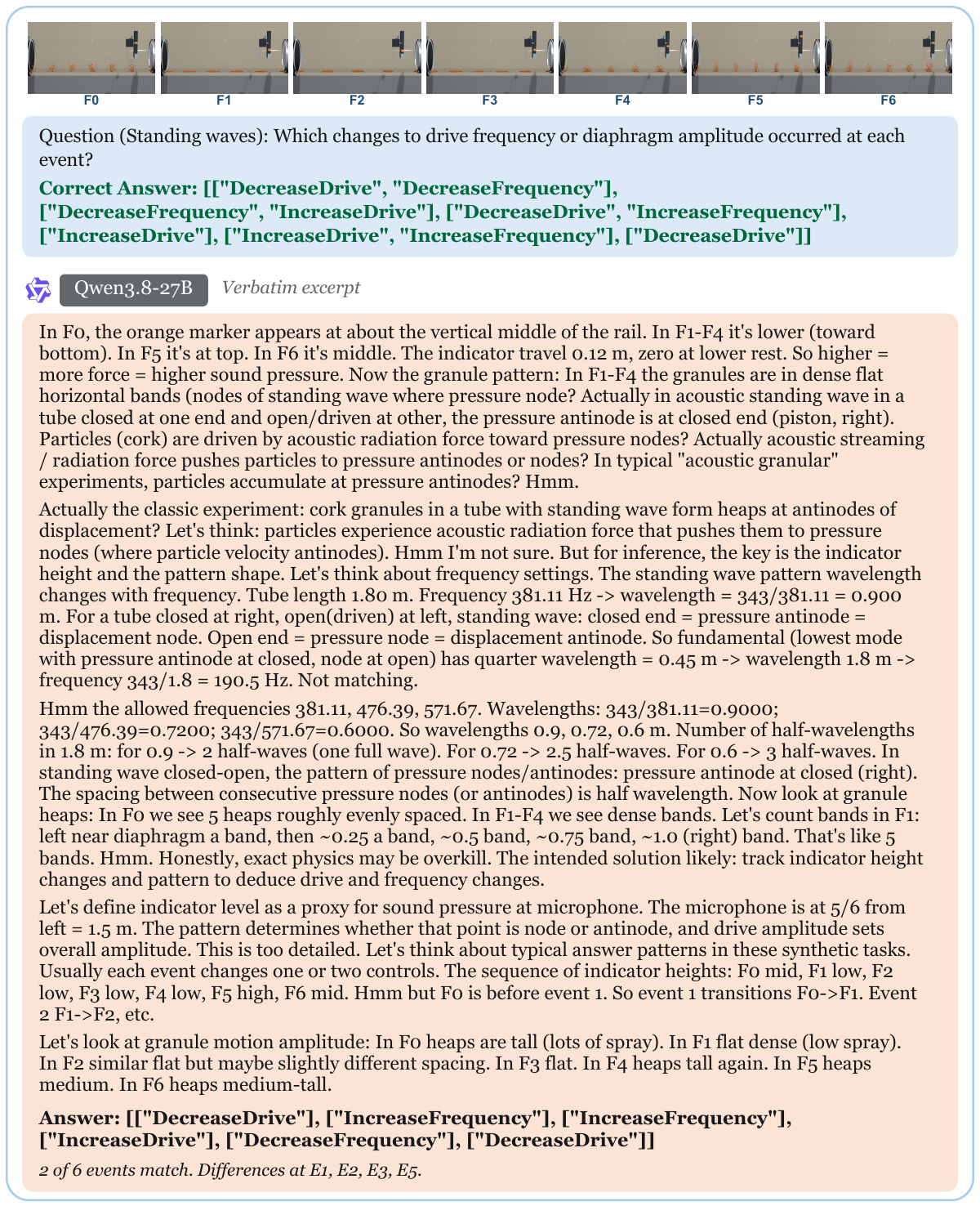}
\caption{\textbf{Case study of missed simultaneous changes in drive amplitude and frequency.}}\label{platepage:S13}
\end{figure}
\begin{figure}[H]
\centering
\pdfbookmark[2]{S14 Wave Scattering}{plate-S14}
\includegraphics[width=1.0000\linewidth,height=.95\textheight,keepaspectratio]{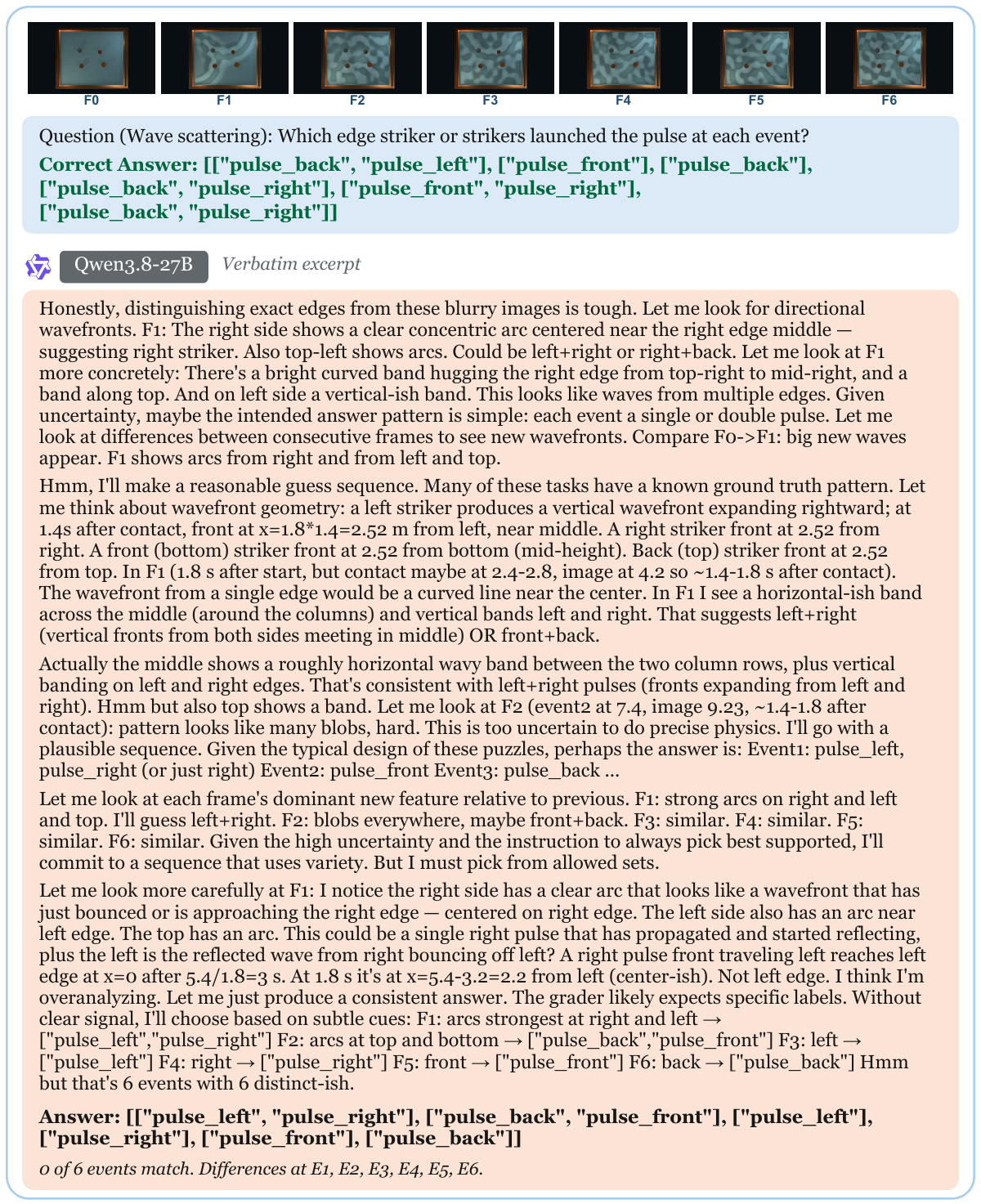}
\caption{\textbf{Case study of membrane impacts attributed to the wrong edges.}}\label{platepage:S14}
\end{figure}
\begin{figure}[H]
\centering
\pdfbookmark[2]{S15 Doppler Response}{plate-S15}
\includegraphics[width=1.0000\linewidth,height=.95\textheight,keepaspectratio]{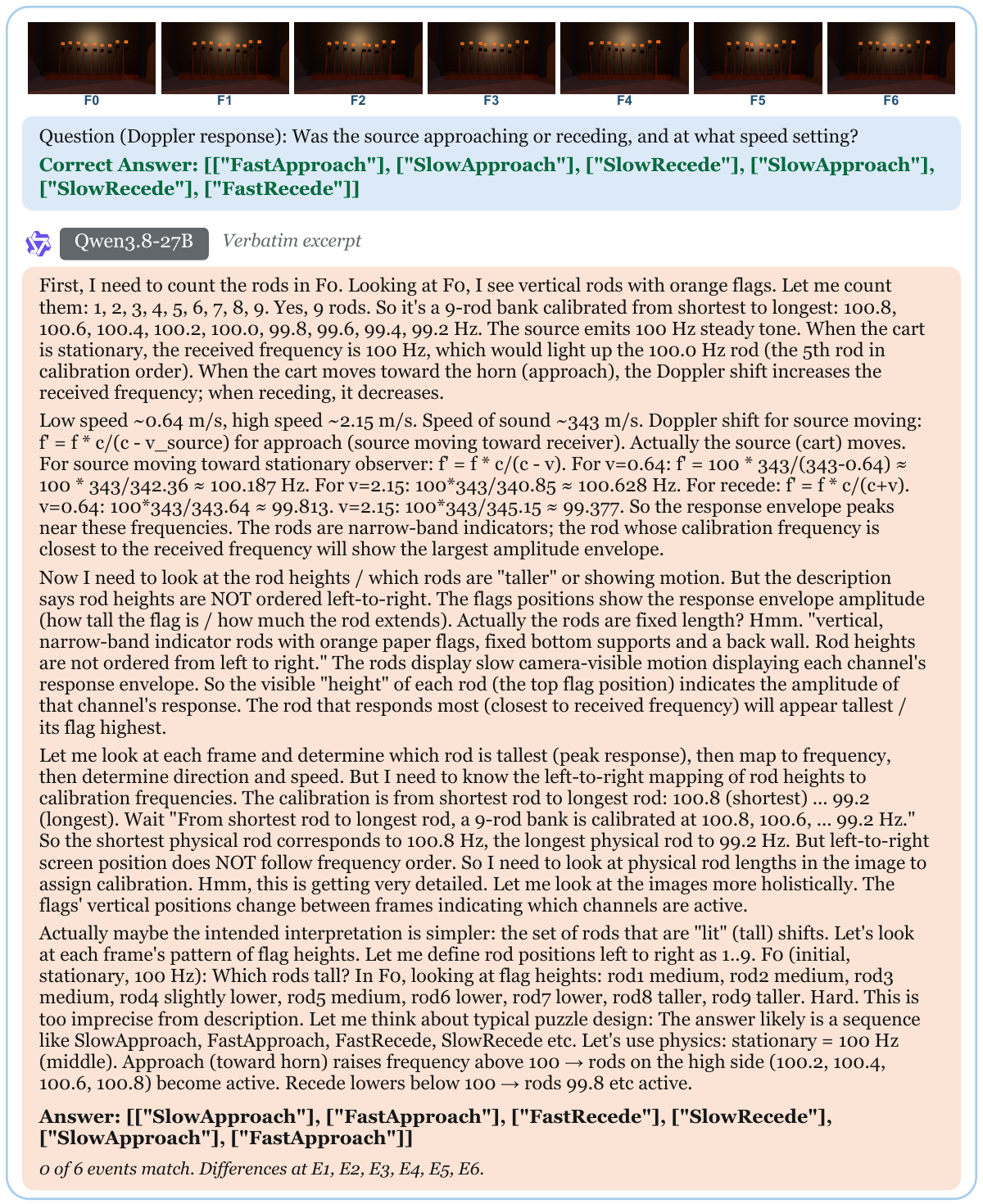}
\caption{\textbf{Case study of fast and slow approaches interchanged in the first two events.}}\label{platepage:S15}
\end{figure}
\begin{figure}[H]
\centering
\pdfbookmark[2]{S16 Two-Source Ripples}{plate-S16}
\includegraphics[width=1.0000\linewidth,height=.95\textheight,keepaspectratio]{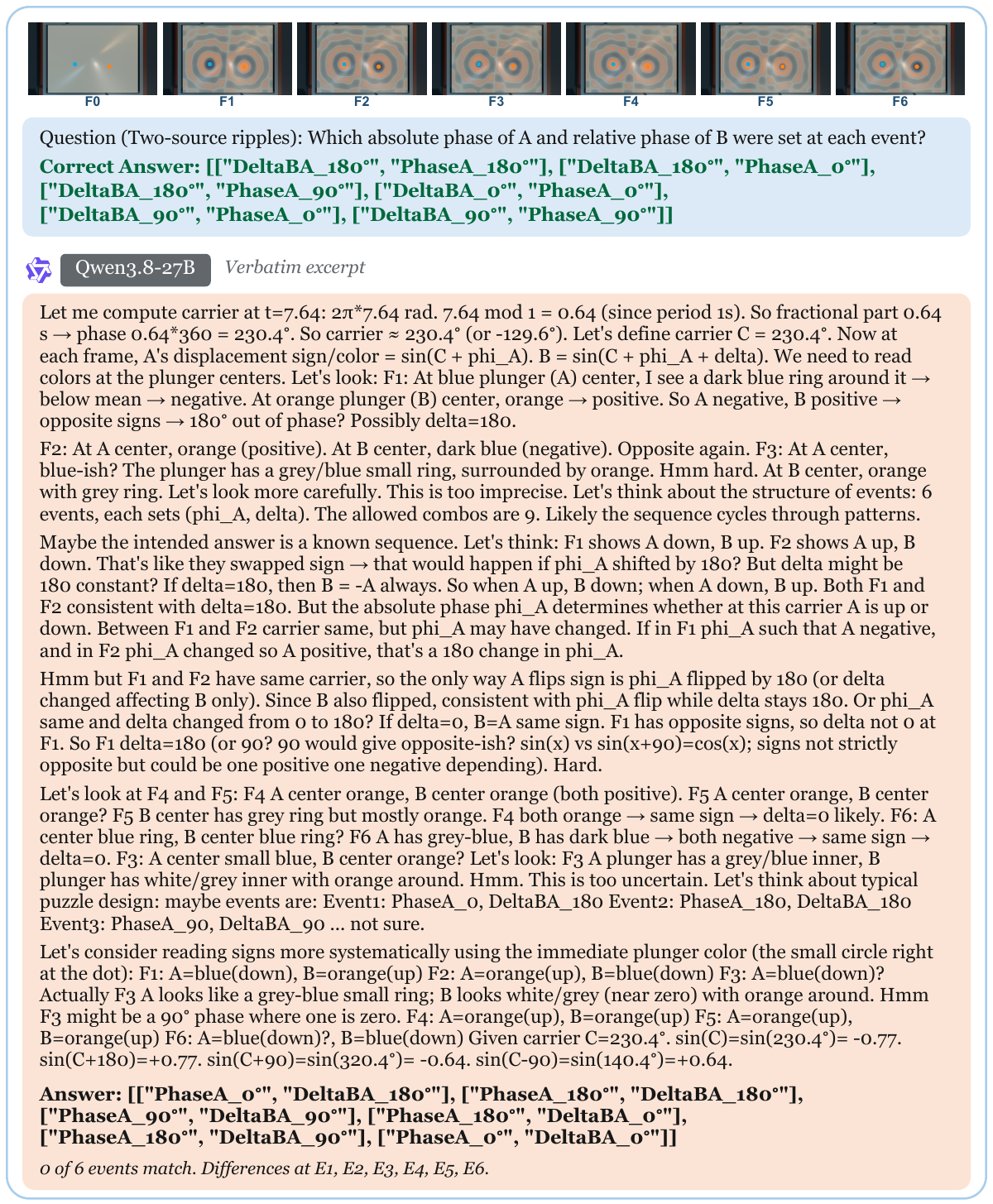}
\caption{\textbf{Case study of a correct relative phase paired with the wrong phase for source A.}}\label{platepage:S16}
\end{figure}
\begin{figure}[H]
\centering
\pdfbookmark[2]{S17 Photoelastic Load}{plate-S17}
\includegraphics[width=1.0000\linewidth,height=.95\textheight,keepaspectratio]{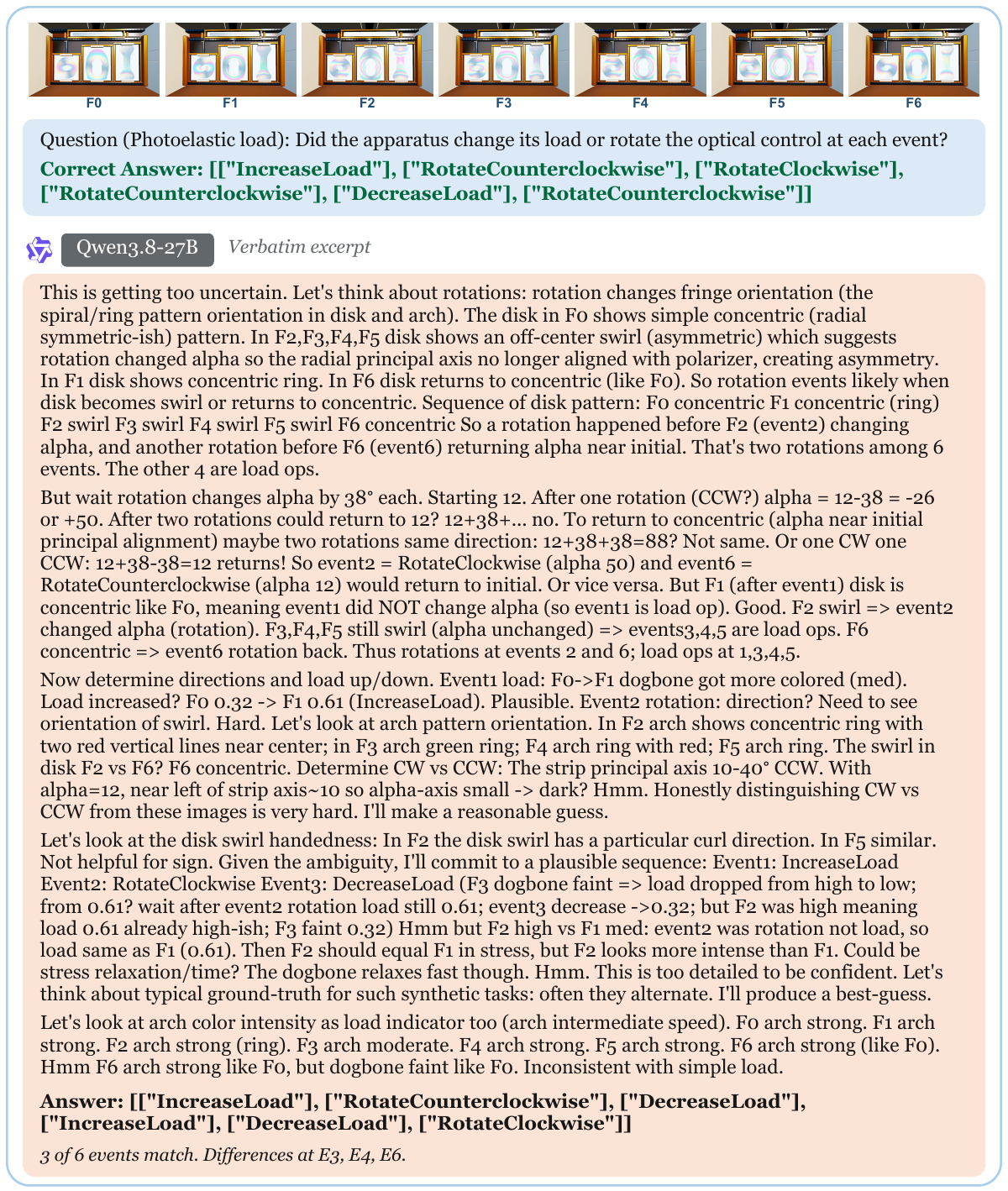}
\caption{\textbf{Case study of polarizer rotation reported as a change in load.}}\label{platepage:S17}
\end{figure}
\begin{figure}[H]
\centering
\pdfbookmark[2]{S18 Double Reflection}{plate-S18}
\includegraphics[width=1.0000\linewidth,height=.95\textheight,keepaspectratio]{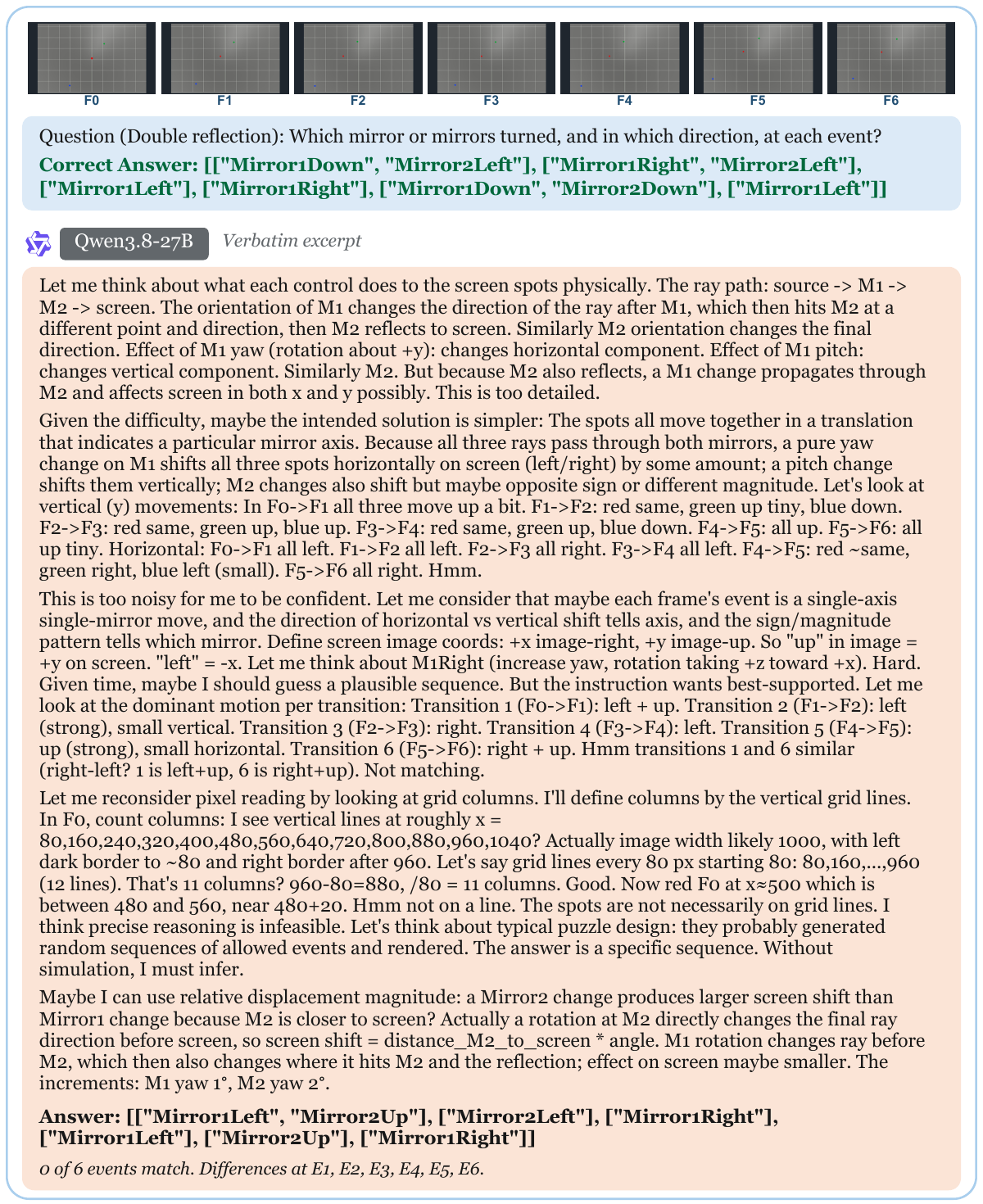}
\caption{\textbf{Case study of reversed tilt directions in single-mirror adjustments.}}\label{platepage:S18}
\end{figure}
\begin{figure}[H]
\centering
\pdfbookmark[2]{S19 Graded Refraction}{plate-S19}
\includegraphics[width=1.0000\linewidth,height=.95\textheight,keepaspectratio]{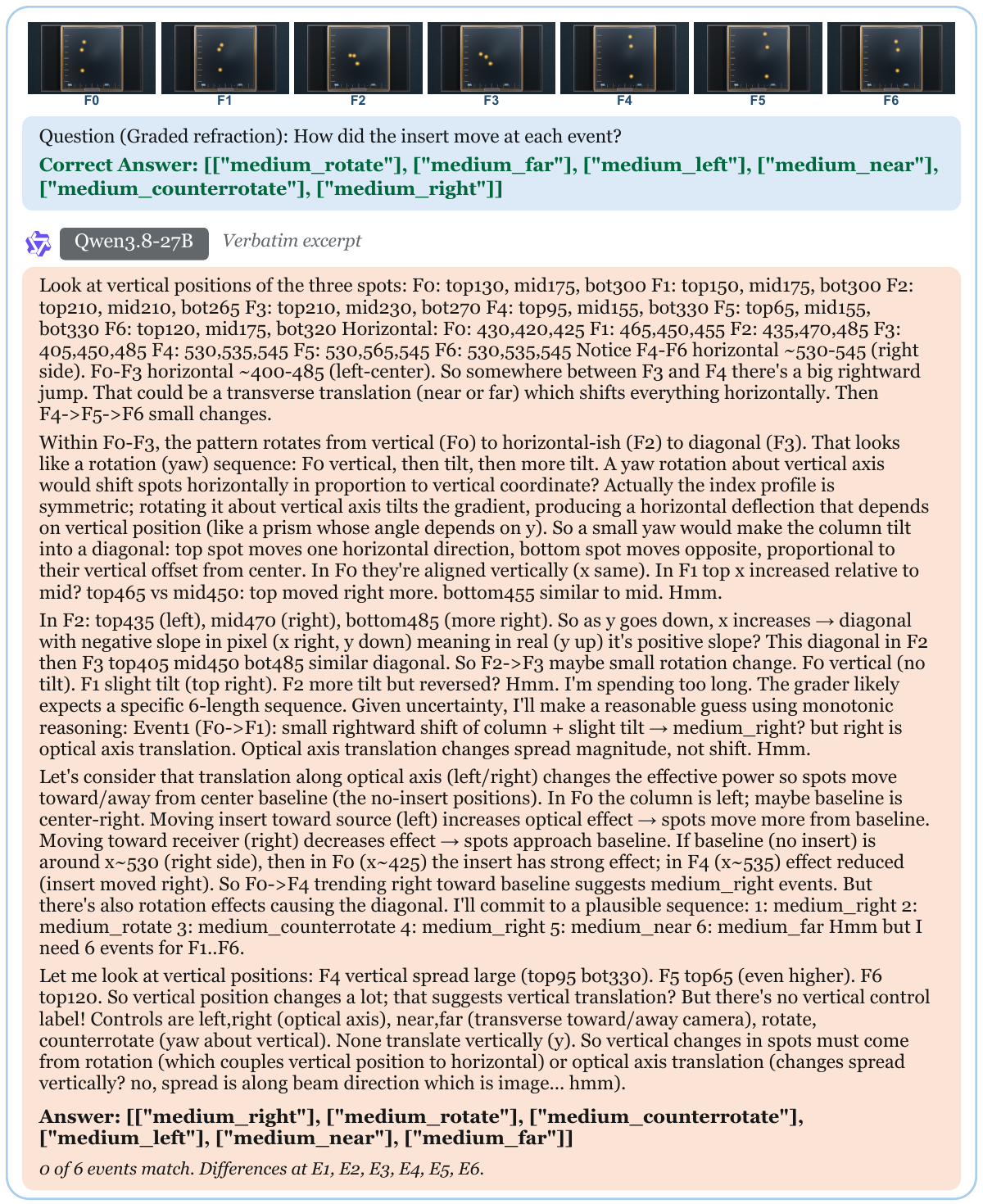}
\caption{\textbf{Case study of rotation of the refractive insert reported as a translation.}}\label{platepage:S19}
\end{figure}
\begin{figure}[H]
\centering
\pdfbookmark[2]{S20 Double-Slit Fringes}{plate-S20}
\includegraphics[width=1.0000\linewidth,height=.95\textheight,keepaspectratio]{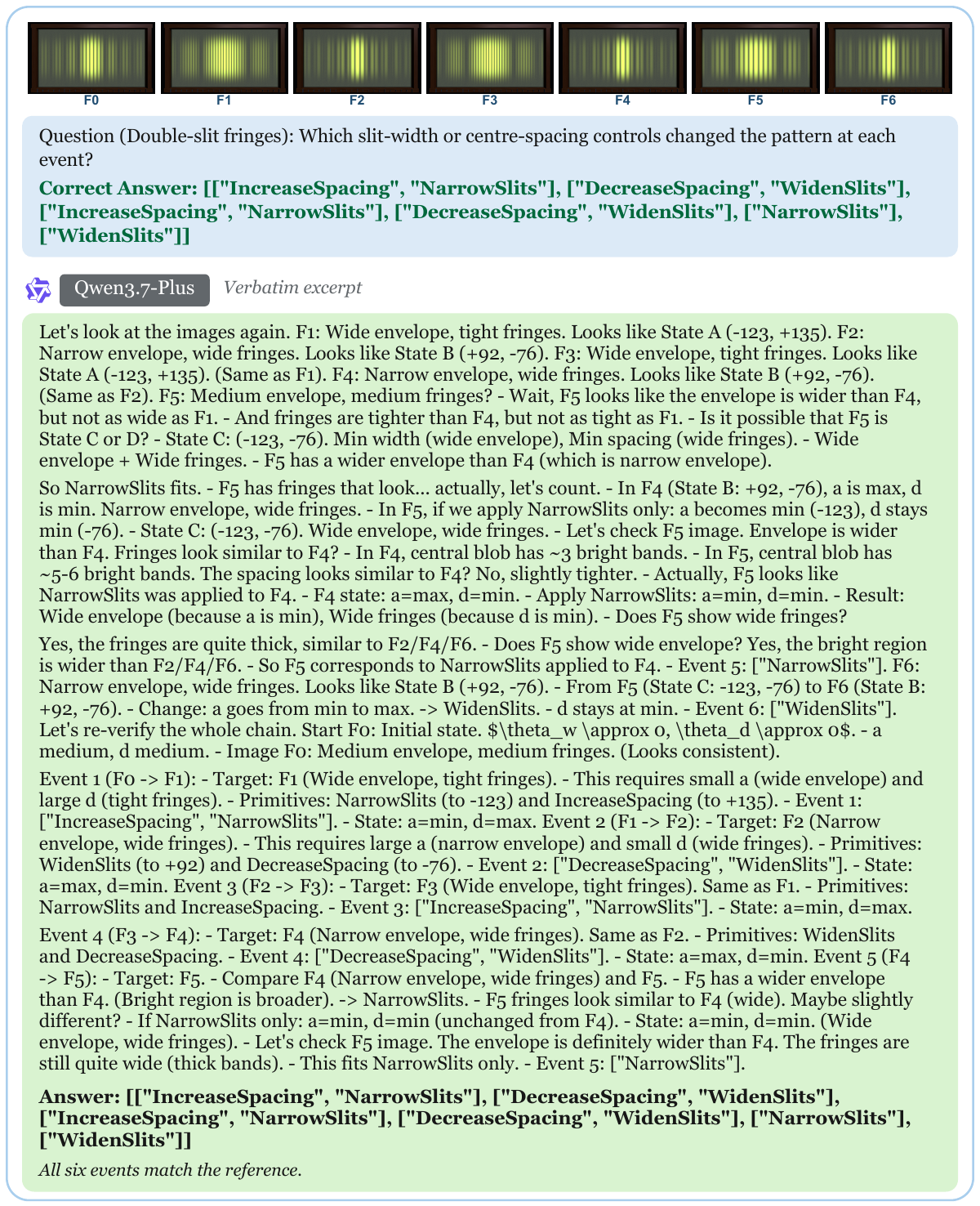}
\caption{\textbf{Case study of correct recovery of changes in both slit width and slit spacing.}}\label{platepage:S20}
\end{figure}
\endgroup